\pdfoutput=1
\documentclass{article}

\PassOptionsToPackage{numbers, compress}{natbib}

\usepackage[main, final]{neurips_2026}

\usepackage[utf8]{inputenc} %
\usepackage[T1]{fontenc}    %
\usepackage{xcolor}         %
\definecolor{mydarkred}{rgb}{0.6,0,0}
\definecolor{mydarkgreen}{rgb}{0,0.6,0}
\usepackage{hyperref}       %
\hypersetup{colorlinks,
  linkcolor=mydarkred,
  citecolor=mydarkgreen}
\usepackage{url}            %
\usepackage{booktabs}       %
\usepackage{amsfonts}       %
\usepackage{nicefrac}       %
\usepackage{microtype}      %
\usepackage{xcolor}         %

\usepackage{amsmath}        %
\usepackage{amssymb}        %
\usepackage{mathtools}      %
\usepackage{amsthm}         %
\usepackage{amsfonts}       %
\usepackage{nicefrac}       %
\usepackage{xcolor}  
\usepackage{array}
\usepackage{algorithm}
\usepackage{algorithmic}
\usepackage{multirow}
\usepackage[capitalize,noabbrev]{cleveref}
\usepackage{enumitem}
\usepackage{adjustbox}
\usepackage{wrapfig}
\usepackage{threeparttable}
\usepackage{etoc}
\usepackage[table]{xcolor}
\theoremstyle{plain}
\newtheorem{theorem}{Theorem}[section]
\newtheorem{proposition}[theorem]{Proposition}
\newtheorem{lemma}[theorem]{Lemma}

\theoremstyle{definition}

\newtheorem{assumption}[theorem]{Assumption}
\theoremstyle{remark}

\title{\textsl{DELTA}: Robustly Training Label-Conditional \underline{D}iffusion Mod\underline{EL}s with Weak Anno\underline{TA}tions}

\author{%
  Dong-Dong Wu$^{1,2}$, \quad
  Jiacheng Cui$^{3}$, \quad
  Wei Wang$^{2}$, \\[4pt]
  {\bfseries Zhiqiang Shen$^{3}$, \quad
  Masashi Sugiyama$^{2,1}$} \\[4pt]
  $^{1}$The University of Tokyo \\
  $^{2}$RIKEN Center for Advanced Intelligence Project \\
  $^{3}$Mohamed bin Zayed University of Artificial Intelligence
}

\begin{document}

\etocdepthtag.toc{mtmain}%

\maketitle

\begin{abstract}
  While label-conditional diffusion models have recently exhibited remarkable generative capabilities, their success heavily relies on massive, cleanly labeled datasets.
  In practice, categorical supervision is rarely perfect: it is often corrupted by noise, clouded by ambiguity, or partially missing.
  Training directly on such weak annotations severely degrades generation quality, and existing robust methods offer only fragmented solutions that depend on scarce auxiliary priors.
  To overcome these limitations, we propose $\textsl{DELTA}$, a unified framework that robustly trains \underline{D}iffusion mod\underline{EL}s across all three weak anno\underline{TA}tion types.
  By treating the unknown true label as a latent variable, $\textsl{DELTA}$ optimizes a principled variational objective that jointly recovers the clean data distribution and infers the true label posteriors.
  To make this joint training computationally tractable, we further introduce a median-centered timestep sampling strategy that efficiently concentrates estimation where the diffusion process is most informative.
  Extensive experiments on synthetic and real-world benchmarks demonstrate that $\textsl{DELTA}$ produces high-fidelity, class-consistent samples across diverse weak supervision scenarios, outperforming specialized baselines while demanding strictly less prior knowledge.
\end{abstract}

\newsavebox{\dmscmpbox}
\newlength{\dmscmprowh}
\newcommand{\dmscmpcell}[1]{%
    \vbox to \dmscmprowh{\vfil\hbox to 1.6em{\hss\scriptsize\bfseries#1\hss}\vfil}%
}

\section{Introduction}
\label{sec:Intro}
Diffusion models (DMs)~\citep{ho2020denoising,songscore,fuest2026diffusion,rombach2022high} have become the de facto framework for high-fidelity image generation across diverse visual domains~\citep{ho2022video,wu2024improvedvideovaelatent,fei2025diffusion}.
In practice, such generation is usually paired with a target concept, and the choice of conditioning signal shapes what the model can produce.
Two of the most prominent paradigms are \textit{text-conditional} diffusion~\citep{esser2024scaling,cao2025controllable}, which targets open-ended generation from natural-language prompts, and \textit{label-conditional} diffusion models (LCDMs)~\citep{ho2022classifier,shenoy2026gradient,na2024label}, which target class-conditional distributions over discrete labels.
Which paradigm applies depends on the form of supervision available in the target domain.

It is often assumed that natural language can express richer and more fine-grained descriptions than discrete class labels, leading the community's attention toward text-conditional generation~\citep{zhu2023conditional, zhanconditional, cao2025controllable, noguchi2024scene}. Yet \textit{outside natural-language interaction settings}, the supervision routinely comes in categorical form: web-scale media is overwhelmingly labeled by tags rather than captions~\cite{mahajan2018exploring}; medical imaging is annotated with diagnostic codes such as lesion types, rather than free-text descriptions~\cite{kazerouni2023diffusion, yellapragada2025pixcell}; and remote-sensing imagery is supervised by land-cover or defect-category metadata~\cite{khannadiffusionsat, tang2024crs}.
\textit{Label-conditional diffusion remains uniquely suited to these settings}: it conditions on categorical supervision and can 
synthesize samples for any target class, helping to mitigate practical data challenges such as costly real-world acquisition, limited 
distribution coverage, and scarce per-class samples.

\begin{figure}[!t]
    \centering
    \includegraphics[width=0.9\linewidth]{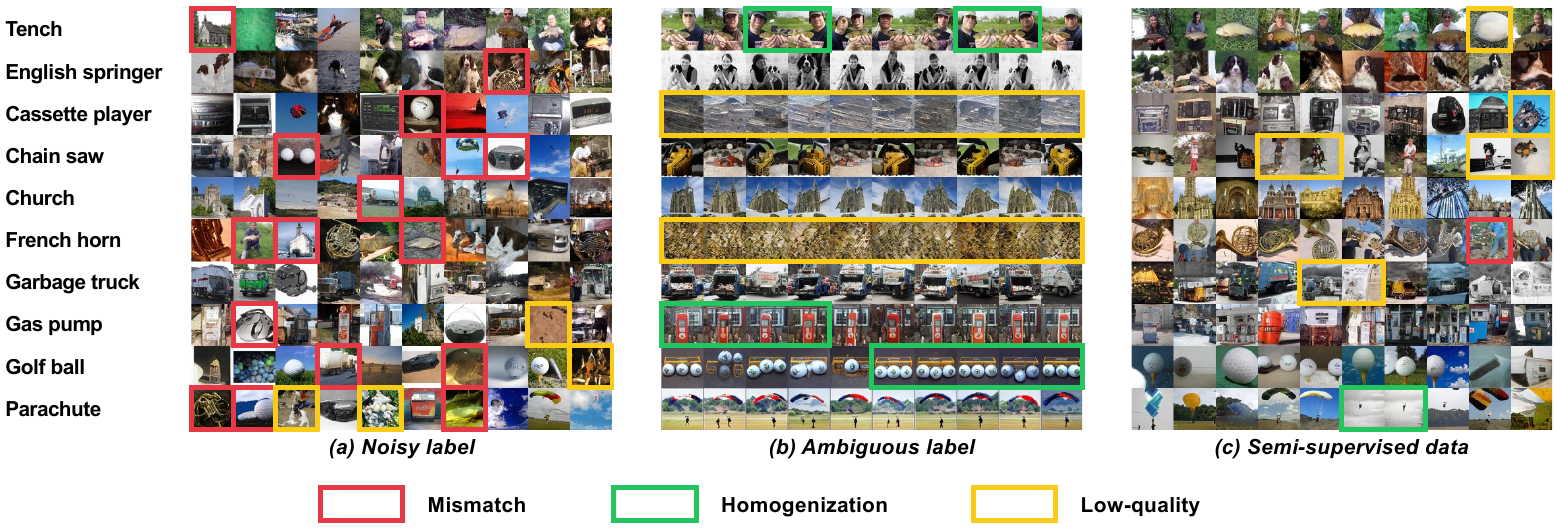}
    \vspace{-6pt}
    \caption{Randomly generated ImageNette samples from naive LCDMs trained under three forms of weak annotations: (a) noisy labels, (b) ambiguous labels, and (c) semi-supervised data. Colored boxes highlight three distinct failure modes that arise across these settings: {\setlength{\fboxsep}{0pt}\setlength{\fboxrule}{0.8pt}\fcolorbox{red}{white}{\rule{0pt}{6pt}\rule{12pt}{0pt}}}~class--image \textit{mismatch}; {\setlength{\fboxsep}{0pt}\setlength{\fboxrule}{0.8pt}\fcolorbox{green}{white}{\rule{0pt}{6pt}\rule{12pt}{0pt}}}~per-class \textit{homogenization}; and {\setlength{\fboxsep}{0pt}\setlength{\fboxrule}{0.8pt}\fcolorbox{orange}{white}{\rule{0pt}{6pt}\rule{12pt}{0pt}}}~\textit{low-quality} generations.}
    \label{fig:vanilla_imagenette}
\end{figure}

Yet in such LCDM scenarios, the real-world data used for training rarely comes with clean labels.
In reality, such label supervision commonly manifests in three forms: labels are \textit{noisy} when accumulated from heterogeneous sources such as user tags or automated taggers~\citep{li2017webvision}; \textit{ambiguous} when the true class is hard to single out, such as among similar lesion types in medical imaging~\citep{wang2025realistic,wang2023climage}; or simply \textit{missing} when much of the collected data carries no associated label at all~\citep{he2023species196}.
We refer to these three forms of imperfect supervision collectively as \textit{weak annotations}~\citep{sugiyama2022machine}.
Training an LCDM directly on such imperfect supervision is fundamentally hazardous: as Figure~\ref{fig:vanilla_imagenette} illustrates, the generated samples consistently suffer from class--image \textit{mismatch}, per-class \textit{homogenization}, and degraded overall \textit{quality}.
To address these challenges, recent studies have proposed adaptations of diffusion models, such as noise-robust~\citep{na2024label,li2024risk,dufour2024don,chen2025robust} and semi-supervised~\citep{you2023diffusion} variants. 
However, these approaches each target a single weak form and further rely on task-specific external supervision, such as pre-estimated noise transition matrices~\citep{na2024label} or risk confidence scores~\citep{li2024risk}, which are seldom available in practice and cannot be straightforwardly combined to cover all three weak forms jointly.
What is missing is a single LCDM training framework that handles all forms of weak annotations without invoking such auxiliary priors or committing to any single weak-label assumption.

In this paper, we propose \textsl{DELTA}, a unified framework for robustly training label-conditional \underline{D}iffusion mod\underline{EL}s under all three forms of weak anno\underline{TA}tions.
Our key idea is that the joint likelihood of data and weak annotations admits a variational lower bound (Section~\ref{subsec:learning}) whose maximization naturally yields an Expectation-Maximization (EM)-style training procedure. The \textbf{E-step} estimates per-sample true-label posteriors with the model fixed, and the \textbf{M-step} updates the model using two synergistic loss terms: a \textit{Generative Term} for posterior-weighted score matching (Section~\ref{subsec:generation}) and a \textit{Classification Term} for refining the model's class-posterior estimates (Section~\ref{subsec:classification}).
Because both the E-step and the M-step involve expensive per-sample posterior computations, we further propose an efficient timestep sampling strategy (Section~\ref{subsec:timereduction}) that substantially reduces this overhead without sacrificing accuracy.
Extensive experiments on image generation and classification validate that \textsl{DELTA} is robust across all three weak forms, outperforms specialized baselines that rely on auxiliary priors, and yields posteriors that rival dedicated classifiers.

Our contributions can be summarized as follows:
\begin{itemize}
    \item We propose \textsl{DELTA}, a unified LCDM training framework that handles noisy, ambiguous, and missing labels within a single latent-variable objective, requiring less prior knowledge.
    \item We derive a median-centered timestep sampling strategy, supported by both theoretical and empirical analyses, that reduces posterior estimation cost by $27.5\times$ with no accuracy loss.
    \item Extensive experiments validate that \textsl{DELTA} outperforms specialized baselines that rely on auxiliary supervision, with analyses confirming posterior quality and sampling efficiency.
\end{itemize}

\section{Related Work}
\paragraph{Robust Training of LCDMs}
Standard LCDMs~\citep{ho2020denoising,ho2022classifier} are trained with clean labels, yet real-world supervision is often noisy, ambiguous, or missing, prompting a recent line of work on robust LCDM training under such weak annotations~\citep{na2024label,li2024risk,dufour2024don,chen2025robust,cong2025guiding,you2023diffusion,takahashi2025positive}.
For noisy labels, prior work either explicitly models the data distribution via noise transition matrices~\citep{na2024label} or risk-parameterized Stochastic Differential Equations (SDEs)~\citep{li2024risk}, or repairs the conditioning signal through trainable pseudo-conditions~\citep{chen2025robust}, confidence-based reweighting~\citep{dufour2024don,yu2024your}, and post-hoc discriminator correction~\citep{cong2025guiding}.
A separate line addresses missing labels via consistency regularization~\citep{you2023diffusion} or positive-unlabeled formulations~\citep{takahashi2025positive}.
Despite these efforts, each handles only a \emph{single} weak form and relies on task-specific external supervision (e.g., transition matrices and confidence scores) rarely available in practice.
\textsl{DELTA} closes this gap by treating clean labels as {latent variables}, yielding a single EM-style LCDM training procedure that handles all three weak forms without such priors or auxiliary networks.

\paragraph{Diffusion Classifiers}
A pretrained diffusion model can itself act as a zero-shot classifier by inferring class likelihoods via Bayes' rule on its denoising-based evidence lower bound (ELBO)~\citep{clark2023text,li2023your}.
This paradigm has since been refined for adversarial robustness via likelihood maximization~\citep{chen2023robust}, certified robustness via randomized smoothing~\citep{chen2024diffusion}, matched-noise selection for sharper logits~\citep{wang2025noise}, and computational efficiency~\citep{qi2024simple}; recent work has also probed its compositional understanding~\citep{jeong2025diffusion}.
However, all of these methods deploy the diffusion classifier exclusively at \emph{inference time} on a frozen, pretrained backbone, where the expensive per-class likelihood evaluation across many timesteps is amortized over a single classification call.
\textsl{DELTA} is, to our knowledge, the first to embed the diffusion classifier \emph{into training itself}, using its posterior to guide updates of the same model. This is made practical by our proposed median-centered timestep sampling strategy.

\section{Background}
\label{sec:background}

\textbf{Diffusion Models.}~~~Let $\mathcal{X} \subseteq \mathbb{R}^d$ be the $d$-dimensional input space. Given a clean sample $\mathbf{x}$ from the real data distribution with density $q(\mathbf{x})$, the forward diffusion process injects Gaussian noise to generate a sequence of latent variables $\{\mathbf{x}_t\}_{t=1}^T$. This transition is defined by the conditional Gaussian density $\mathcal{N}$:
\begin{equation}
\label{eq:forward}
q(\mathbf{x}_t\!\mid\!\mathbf{x}) = \mathcal{N}(\mathbf{x}_t; \alpha_t\mathbf{x}, \sigma_t^2 \mathbf{I}),
\end{equation}

where $\alpha_t>0$ and $\sigma_t>0$ are the signal scaling and noise standard deviation schedules, respectively, and $\mathbf{I}$ represents the identity matrix.
The reverse generative process aims to invert this corruption by learning a denoising distribution $p_\theta(\mathbf{x}_{t-1}\!\mid\!\mathbf{x}_t)$ to recover $\mathbf{x}$ from a pure noise prior $\mathbf{x}_T \sim \mathcal{N}(\mathbf{0}, \mathbf{I})$, where $\mathbf{0}$ denotes the zero vector.
This is typically achieved by maximizing the ELBO on the log-likelihood $\log p_\theta(\mathbf{x})$.
By parameterizing the reverse step with a score network $\mathbf{s}_\theta(\mathbf{x}_t, t)$, the ELBO simplifies to a weighted denoising score matching (DSM) objective:

\begin{equation}
\label{eq:LL}
\mathcal{L}(\theta) = \mathbb{E}_{t, \mathbf{x}, \mathbf{x}_t} \bigg[
   w_t \Big\|
   \mathbf{s}_\theta(\mathbf{x}_t, t) 
   - \nabla_{\mathbf{x}_t}\log q(\mathbf{x}_t \!\mid\! \mathbf{x})
   \Big\|_2^2
\bigg],
\end{equation}

where the expectation is taken over timestep $t$, clean samples $\mathbf{x} \!\sim\! q(\mathbf{x})$, and noisy samples $\mathbf{x}_t \!\sim\! q(\mathbf{x}_t \!\mid\! \mathbf{x})$; $w_t$ is a weighting function, and the analytic conditional score $\nabla_{\mathbf{x}_t} \log q(\mathbf{x}_t \!\mid\! \mathbf{x}) = (\alpha_t\mathbf{x} - \mathbf{x}_t)/\sigma_t^2$ serves as the tractable training target~\citep{vincent2011connection}.
By varying schedules $\alpha_t, \sigma_t, w_t$, this framework unifies Denoising Diffusion Probabilistic Models (DDPM)~\citep{ho2020denoising}, Score-based SDEs~\citep{songscore}, and Elucidated Diffusion Models (EDM)~\citep{karras2022elucidating} (see Appendix~\ref{proof:convert}).
In this work, we adopt the EDM formulation as our default backbone for its well-optimized parameterization.

\textbf{Label-Conditional Diffusion Models (LCDMs).}~~~For label-conditional generation, the score network additionally takes a class label $y$ as input and is trained to approximate the clean class-conditional score $\nabla_{\mathbf{x}_t}\log q_t(\mathbf{x}_t \!\mid\! y)$ via the conditional analog of Eq.~(\ref{eq:LL}):
\begin{equation}
\label{eq:LL_cond}
\mathcal{L}_\text{LCDM}(\theta) = \mathbb{E}_{t, \mathbf{x}, y, \mathbf{x}_t} \bigg[ w_t \Big\| \mathbf{s}_\theta(\mathbf{x}_t, y, t) - \nabla_{\mathbf{x}_t}\log q(\mathbf{x}_t \!\mid\! \mathbf{x}, y) \Big\|_2^2 \bigg].
\end{equation}
Eq.~(\ref{eq:LL_cond}) is the standard LCDM objective and presupposes access to clean labels $y$.

\textbf{Weak Annotations.}~~~In real-world settings, we observe pairs $(X, Z)$ where the input $X \in \mathcal{X}$ is paired with a weak annotation $Z$ that conveys imperfect information about the unobserved true label $Y \in \mathcal{Y} = \{1, \dots, c\}$. We denote the dataset of $N$ such observations as $\mathcal{D} = \{(\mathbf{x}^{(i)}, z^{(i)})\}_{i=1}^N$, and focus on three representative weak annotation scenarios:
\begin{itemize}[left=10pt, itemsep=0pt, topsep=2pt]
    \item \textit{Noisy Labels~\citep{han2020survey}:} The annotation $Z = \hat{Y}$ is a corrupted version of $Y$ drawn from a noise transition $p(\hat{Y} \!\mid\! Y)$, common when labels come from user tags or automated taggers.
    \item \textit{Ambiguous Labels~\citep{tian2023partial}:} The annotation $Z \subseteq \mathcal{Y}$ is a candidate set containing the true label ($Y \in Z$), arising when the true class is hard to single out among look-alike categories.
    \item \textit{Semi-supervised Data~\citep{yang2022survey}:} The annotation $Z \in \mathcal{Y} \cup \{\emptyset\}$ is either the true label $Y$ (labeled instances) or $\emptyset$ (unlabeled ones), since much of the collected data carries no label at all.
\end{itemize}

\section{Methodology}
\label{sec:method}
In this section, we develop a unified probabilistic framework for training LCDMs under weak annotations, in which the learning objective is derived from a maximum-likelihood perspective and decomposed into generative and classification components.

\subsection{Unified Probabilistic Framework}
\label{subsec:learning}
Given a dataset with inputs $X$ and weak annotations $Z$, we propose to treat the unobserved true label $Y$ as a latent variable, and to learn the optimal model parameters $\theta^*$ by maximizing the log-likelihood of the observed joint observation $(X, Z)$:
\begin{equation}
\label{eq:MLP}
\begin{aligned}
\theta^* &= {\arg\max}_{\theta} \log p_{\theta}(X,Z) = {\arg\max}_{\theta}\log\sum\nolimits_{Y\in\mathcal{Y}}p_{\theta}(X,Y,Z).
\end{aligned}
\end{equation}

Since Eq.~(\ref{eq:MLP}) involves the log of the marginalization over latent variables and cannot generally be solved in closed form, we instead maximize a variational lower bound on the log-likelihood:
\begin{equation}
\label{eq:VLB_LL}
\begin{aligned}
\theta^*
&= {\arg\max}_{\theta}\ \mathbb{E}_{p_{\phi}(Y \mid X,Z)} \big[ \log p_{\theta}(X,Y,Z) \big] \\
&= {\arg\max}_{\theta}\ \Big\{ \underbrace{\mathbb{E}_{p_{\phi}(Y \mid X,Z)} \big[\log p_{\theta}(X \!\mid\! Z)\big]}_{\text{Generative Term}} + \underbrace{\mathbb{E}_{p_{\phi}(Y \mid X,Z)} \big[ \log p_{\theta}(Y \!\mid\! X, Z) \big]}_{\text{Classification Term}} \Big\},
\end{aligned}
\end{equation}

where $p_{\phi}(Y \!\mid\! X,Z)$ is a parameterized proxy for the true class-posterior, with a detailed derivation in Appendix~\ref{proof:VLB}. This decomposition naturally gives rise to an EM-like training procedure: in the \textbf{E-step}, $p_\phi$ is evaluated with $\phi$ held fixed to supply per-sample soft posteriors; in the \textbf{M-step}, $\theta$ is updated via the two loss terms using these posteriors. In practice, we set $\phi$ to an exponential moving average (EMA) of $\theta$, so the posteriors are stably refined alongside the model, and a better $\theta$ yields a more accurate $p_\phi$ that improves posterior estimates in the next E-step.
The two terms reflect the dual role of the diffusion model: the \textit{Generative Term} recovers clean class-conditional distributions, while the \textit{Classification Term} aligns the model's own posterior $p_\theta(Y \!\mid\! X, Z)$ with the E-step soft targets, whose updates propagate back to $p_\phi$ via EMA.
In this paper, we adopt the standard class-conditional assumption~\citep{liu2023identifiability}, where the generation of the weak annotation $Z$ only depends on the true label $Y$, i.e., $p(Z \!\mid\! Y, X) = p(Z \!\mid\! Y)$.
Each term in Eq.~(\ref{eq:VLB_LL}) is maximized by minimizing a corresponding loss, derived in Sections~\ref{subsec:generation} and~\ref{subsec:classification} respectively.

\subsection{Generative Learning via Posterior-Weighted Score Matching}
\label{subsec:generation}
In this subsection, we focus on optimizing the \textit{Generative Term} in Eq.~(\ref{eq:VLB_LL}) using only weak pairs $(\mathbf{x}, z)$ to learn a class-conditional score network that approximates the clean-conditional score.

Mathematically, since $\log p_\theta(X \!\mid\! Z)$ does not depend on $Y$, the expectation $\mathbb{E}_{p_\phi(Y \mid X, Z)}$ in Eq.~(\ref{eq:VLB_LL}) drops out, and the \textit{Generative Term} reduces to maximizing $\log p_\theta(X \!\mid\! Z)$ directly.
To learn the clean-conditional scores $\nabla_{\mathbf{x}_t} \log q_t(\mathbf{x}_t \!\mid\! y)$ from data paired only with $z$, we need to bridge the weak and clean conditional distributions. Inspired by the relationships between noisy and clean distributions explored in noise-robust diffusion training~\citep{na2024label}, we formally establish the link between the observable \textit{weak-conditional score} and the latent \textit{clean-conditional scores} (Appendix~\ref{proof:decouple}):

\begin{theorem}[Score Decomposition under Weak Annotations]
\label{theorem:decouple}
Under the class-conditional independence assumption ($X \perp Z \mid Y$), the score of the perturbed data density given a weak annotation $z$ can be decomposed as a posterior-weighted combination of clean-conditional scores:

\begin{equation} 
\label{eq:convex_combination}
\nabla_{\mathbf{x}_t}\!\log q_t(\mathbf{x}_t \!\mid\! z)
= \sum\nolimits_{y=1}^{c} p(y \!\mid\! \mathbf{x}_t, z)\, \nabla_{\mathbf{x}_t}\!\log q_t(\mathbf{x}_t \!\mid\! y).
\end{equation}
\end{theorem}

\textbf{Posterior-Weighted Score Matching.}
Theorem~\ref{theorem:decouple} reveals an asymmetry that motivates our loss design. The clean-conditional scores on the right-hand side are precisely what a class-conditional generative model needs to learn, but they cannot be targeted directly from weakly annotated data. The weak-conditional score on the left-hand side has the opposite property: it is not a desirable learning target since fitting it would produce mixed-class generations, yet it admits a closed-form training target through standard DSM against $(\mathbf{x}, z)$ pairs. Eq.~(\ref{eq:convex_combination}) thus suggests a strategy: substituting the labels $y$ in the standard LCDM objective Eq.~(\ref{eq:LL_cond}) with weak annotations $z$ and replacing the score network output $\mathbf{s}_\theta(\mathbf{x}_t, y, t)$ with its posterior-weighted aggregation, so that the clean-conditional network can be trained indirectly against the observable weak-conditional target:

\begin{equation}
\label{eq:WDSM}
\begin{aligned}
\mathcal{L}_\text{Gen}(\theta) & = \mathbb{E}_{t, \mathbf{x}, z, \mathbf{x}_t}\bigg[
   w_t \Big\|
   \sum\nolimits_{y=1}^c p_\phi(y \!\mid\! \mathbf{x}_t, z)\,\mathbf{s}_\theta(\mathbf{x}_t,y,t) - \nabla_{\mathbf{x}_t}\log q(\mathbf{x}_t \!\mid\! \mathbf{x}, z) 
   \Big\|_2^2
\bigg],
\end{aligned}
\end{equation}

where the target $\nabla_{\mathbf{x}_t}\log q(\mathbf{x}_t \!\mid\! \mathbf{x}, z) = \nabla_{\mathbf{x}_t}\log q(\mathbf{x}_t \!\mid\! \mathbf{x})$ is the closed-form forward perturbation score. Since the true posterior $p(y \!\mid\! \mathbf{x}_t, z)$ in Eq.~(\ref{eq:convex_combination}) is unobservable, we replace it with the EMA-parameterized estimate $p_\phi(y \!\mid\! \mathbf{x}_t, z)$ derived from the diffusion model itself via Bayes' rule, as described next.
Intuitively, when $p_\phi(y \!\mid\! \mathbf{x}_t, z)$ is confident, the gradient primarily updates $\mathbf{s}_\theta$ for the most likely class, while an uncertain posterior spreads the gradient across multiple classes and prevents the model from anchoring to a potentially wrong label.

\begin{proposition}[Identifiability of Clean-Conditional Scores]
\label{proposition:WDSM}
When the posterior estimate converges to the true posterior, i.e., $p_\phi(y \!\mid\! \mathbf{x}_t, z) = p(y \!\mid\! \mathbf{x}_t, z)$, the score network $\mathbf{s}_{\theta^*}$ that minimizes the posterior-weighted score matching objective recovers the clean-conditional scores:

\begin{equation}
\label{eq:xxwdd}
    \mathbf{s}_{\theta^*}(\mathbf{x}_t, y, t) = \nabla_{\mathbf{x}_t}\log q_t(\mathbf{x}_t \!\mid\! y), \quad \forall y \in \mathcal{Y}.
\end{equation}
\end{proposition}

The proof is in Appendix~\ref{proof:WDSM}. In practice, $\phi$ is maintained as an EMA copy of $\theta$, so $p_\phi$ is progressively refined alongside the model in an EM-style cycle: a better $\theta$ yields a more accurate $p_\phi$, which in turn improves $\theta$ via Eq.~(\ref{eq:WDSM}). This iteration progressively drives $p_\phi$ toward the true posterior, eventually satisfying the condition of Proposition~\ref{proposition:WDSM}.

\subsection{Classification via Diffusion-Based Posterior Inference}
\label{subsec:classification}
Both Eq.~(\ref{eq:WDSM}) and the \textit{Classification Term} of Eq.~(\ref{eq:VLB_LL}) require computing class posteriors: $p_\phi(y \!\mid\! \mathbf{x}_t, z)$ for score weighting and $p_\theta(y \!\mid\! \mathbf{x}, z)$ for prediction. Training an external classifier on the same weak annotations would face an identical supervision problem and would introduce objective mismatch. We therefore compute these posteriors directly from the diffusion model itself: since $\log p_\theta(\mathbf{x}_t \!\mid\! y)$ is already modeled by the generative network, Bayes' rule yields the posterior with no extra parameters, allowing generation and classification to mutually reinforce each other~\citep{li2023your}.

\textbf{Diffusion Classifier.}
Following \citet{chen2024diffusion}, we obtain the exact posterior by applying Bayes' rule to the class-conditional log-likelihood and normalizing over all classes via a softmax, $p_\theta(y \!\mid\! \mathbf{x}_t) = \exp\{\log p_\theta(\mathbf{x}_t \!\mid\! y) + \log p(y)\} / \sum_{y' \in \mathcal{Y}} \exp\{\log p_\theta(\mathbf{x}_t \!\mid\! y') + \log p(y')\}$, where the conditional ELBO approximates the log-likelihood as
\begin{equation}
\label{eq:elbo_approx}
\log p_\theta(\mathbf{x}_t \!\mid\! y) \approx -\!\!\sum_{\tau=t+1}^{T-1} \!\! w_\tau\, \hbar(\tau, y),\quad \hbar(\tau,y) = \mathbb{E}_{\mathbf{x}_\tau \sim q(\mathbf{x}_\tau \mid \hat{\mathbf{x}})}\!\left[\|\mathbf{h}_\theta(\mathbf{x}_\tau, y, \tau) - \hat{\mathbf{x}}\|_2^2\right],
\end{equation}
where $\mathbf{h}_\theta(\mathbf{x}_\tau, y, \tau) = (\mathbf{x}_\tau + \sigma_\tau^2\, \mathbf{s}_\theta(\mathbf{x}_\tau, y, \tau)) / \alpha_\tau$ is the class-conditional denoised estimator, $\hat{\mathbf{x}}$ is the unconditional estimate of the clean sample from $\mathbf{x}_t$ (so $\hbar(\tau, y)$ implicitly depends on $\mathbf{x}_t$ through $\hat{\mathbf{x}}$), and $w_\tau$ is the EDM log-normal weight. We assume a uniform class prior $p(y) = 1/c$ throughout, with the extension to non-uniform priors deferred to Appendix~\ref{exp:prior}.
For brevity, we write $f_\theta(\mathbf{x}_t) \in \Delta^{c-1}$ for the resulting normalized posterior, with $[f_\theta(\mathbf{x}_t)]_y = p_\theta(y \!\mid\! \mathbf{x}_t)$, and let $f_\phi$ denote its EMA counterpart, so that $p_\phi(y \!\mid\! \mathbf{x}_t) = [f_\phi(\mathbf{x}_t)]_y$.

\begin{table*}[!t]
    \centering
    \scriptsize
    \tabcolsep 0.04in
    \renewcommand{\arraystretch}{0.5}
    \caption{Generative results under various weak annotations. The gray-shaded `Clean' column represents the best performance trained on the fully clean dataset. `uncond'/`cond' denote unconditional/conditional metrics. \textbf{Bold} numbers indicate better performance.}
    \label{tab:gen_results}
    \resizebox{\textwidth}{!}{%
    \begin{tabular}{cc|lc|cc|cc|cc|cc|cc|cc|>{\columncolor{gray!10}}c}
    \toprule
    \multicolumn{2}{c|}{\multirow{6}{*}{}} & \multicolumn{1}{c}{\multirow{6}{*}{\quad\quad\ Metric}} & & \multicolumn{4}{c|}{Noisy label} & \multicolumn{4}{c|}{Ambiguous label} & \multicolumn{4}{c|}{Semi-supervised data} & \cellcolor{white}{}\\
    \cmidrule(lr){5-8} \cmidrule(lr){9-12} \cmidrule(lr){13-16} 
    & & & & \multicolumn{2}{c}{{Sym-40\%}} & \multicolumn{2}{c|}{{Asym-40\%}} & \multicolumn{2}{c}{{Random}} & \multicolumn{2}{c|}{{Class-dept.}} & \multicolumn{2}{c}{{1\% Labels}} & \multicolumn{2}{c|}{{10\% Labels}} & \multicolumn{1}{c}{\cellcolor{white}{Clean}}\\
    \cmidrule(lr){5-6} \cmidrule(lr){7-8} \cmidrule(lr){9-10} \cmidrule(lr){11-12} \cmidrule(lr){13-14} \cmidrule(lr){15-16}
    & & & & \textsl{Vanilla} & \textsl{DELTA} & \textsl{Vanilla} & \textsl{DELTA} & \textsl{Vanilla} & \textsl{DELTA} & \textsl{Vanilla} & \textsl{DELTA} & \textsl{Vanilla} & \textsl{DELTA} & \textsl{Vanilla} & \textsl{DELTA} & \cellcolor{white}{}\\
    \midrule
    \multirow{12}{*}{\rotatebox{90}{CIFAR-10}}& \multirow{5}{*}{\rotatebox{90}{uncond}} & \multicolumn{1}{|l}{FID} & $(\downarrow)$
    & \textbf{3.33} & 3.47 & 3.23 & \textbf{3.10} & 7.76 & \textbf{2.26} & 11.75 & \textbf{2.77} & 3.16 & \textbf{3.12} & 2.93 & \textbf{2.89} & {2.05}\\
    &  & \multicolumn{1}{|l}{IS} & $(\uparrow)$
    & 9.56 & \textbf{9.68} & 9.02 & \textbf{9.73} & 9.09 & \textbf{9.80} & 9.62 & \textbf{9.68} & 10.03 & \textbf{10.57} & 9.80 & \textbf{9.83} & {10.61}\\
    &  & \multicolumn{1}{|l}{Density} & $(\uparrow)$
    & 101.39 & \textbf{109.75} & 100.06 & \textbf{109.69} & 103.21 & \textbf{106.49} & {108.76} & \textbf{109.06} & 97.19 & \textbf{108.18} & 99.96 & \textbf{108.87} & {112.59}\\
    \vspace{2pt}
    &  & \multicolumn{1}{|l}{Coverage} & $(\uparrow)$
    & 81.12 & \textbf{81.21} & 80.71 & \textbf{81.30} & 68.45 & \textbf{82.69} & 64.90 & \textbf{81.52} & 78.44 & \textbf{81.00} & 81.85 & \textbf{82.00} & {83.27}\\
    & \multirow{4}{*}{\rotatebox{90}{cond}} & \multicolumn{1}{|l}{CW-FID} & $(\downarrow)$
    & 29.84 & \textbf{13.85} & 14.70 & \textbf{13.24} & 27.18 & \textbf{10.65} & 32.44 & \textbf{11.56} & 16.25 & \textbf{16.12} & 11.84 & \textbf{11.77} & {9.83}\\
    & & \multicolumn{1}{|l}{CW-Density} & $(\uparrow)$
    & 72.98 & \textbf{107.23} & 90.85 & \textbf{107.07} & 102.04 & \textbf{105.75} & 102.43 & \textbf{108.66} & 89.99 & \textbf{100.73} & 96.29 & \textbf{107.94} & {111.70}\\
    & & \multicolumn{1}{|l}{CW-Coverage} & $(\uparrow)$
    & 73.39 & \textbf{80.11} & 79.63 & \textbf{79.65} & 65.45 & \textbf{82.09} & 61.45 & \textbf{81.24} & 75.03 & \textbf{76.84} & 80.80 & \textbf{81.12} & {83.91}\\
    \midrule
    \multirow{12}{*}{\rotatebox{90}{ImageNette}}& \multirow{5}{*}{\rotatebox{90}{uncond}} & \multicolumn{1}{|l}{FID} & $(\downarrow)$ & 14.11 & \textbf{13.44} & 13.93 & \textbf{13.91} & 79.13 & \textbf{72.62} & 91.28 & \textbf{79.12} & 23.88 & \textbf{19.26} & 14.32 & \textbf{12.84} & {11.52} \\
             &  & \multicolumn{1}{|l}{IS} & $(\uparrow)$ & 12.69 & \textbf{13.21} & 12.51 & \textbf{13.73} & 9.19 & \textbf{9.40} & \textbf{9.27} & 9.11 & 12.23 & \textbf{13.72} & 12.80 & \textbf{13.16} & {13.81}\\
             &  & \multicolumn{1}{|l}{Density} & $(\uparrow)$ & 109.31 & \textbf{112.52} & \textbf{111.66} & 106.78 & 95.33 & \textbf{99.83} & 94.29 & \textbf{102.58} & 115.94 & \textbf{125.68} & 105.27 & \textbf{109.23} & {117.23}\\
             \vspace{2pt}
             &  & \multicolumn{1}{|l}{Coverage} & $(\uparrow)$ & 76.62 & \textbf{76.81} & 78.32 & \textbf{79.81} & 21.44 & \textbf{32.48} & 16.69 & \textbf{22.30} & 53.53 & \textbf{55.39} & 73.79 & \textbf{75.55} & {80.12}\\
             & \multirow{4}{*}{\rotatebox{90}{cond}} & \multicolumn{1}{|l}{CW-FID} & $(\downarrow)$ & 80.31 & \textbf{60.12} & 62.26 & \textbf{58.20} & 157.76 & \textbf{63.58} & 163.45 & \textbf{67.92} & 71.66 & \textbf{70.27} & 49.22 & \textbf{44.31} & {40.20}\\
             & & \multicolumn{1}{|l}{CW-Density} & $(\uparrow)$ & 73.99 & \textbf{81.12} & 93.53 & \textbf{94.58} & 93.38 & \textbf{95.83} & 91.50 & \textbf{95.21} & 115.90 & \textbf{118.69} & 103.41 & \textbf{115.67} & {120.35}\\
             & & \multicolumn{1}{|l}{CW-Coverage} & $(\uparrow)$ & 67.89 & \textbf{71.94} & 74.18 & \textbf{75.82} & 19.76 & \textbf{24.35} & 15.88 & \textbf{18.93} & 51.73 & \textbf{52.15} & 72.61 & \textbf{74.85} &{78.48}\\
    \bottomrule
    \end{tabular}}
    \end{table*}

\textbf{Classification Term under Weak Annotations.}
Under the class-conditional assumption of Section~\ref{subsec:learning}, the model posterior factorizes as $p_\theta(y \!\mid\! \mathbf{x}, z) = p(z \!\mid\! y)\, p_\theta(y \!\mid\! \mathbf{x}) / p_\theta(z \!\mid\! \mathbf{x})$. Since $p(z \!\mid\! y)$ is $\theta$-independent and $\log p_\theta(z \!\mid\! \mathbf{x}) \leq 0$, the \textit{Classification Term} in Eq.~(\ref{eq:VLB_LL}) admits the tractable lower bound $\mathbb{E}_{p_\phi(y \mid \mathbf{x}, z)}[\log p_\theta(y \!\mid\! \mathbf{x})]$ (derivation and boundedness of the dropped term in Appendix~\ref{proof:cls_upper_bound}), whose maximization is equivalent to minimizing the posterior-weighted cross-entropy below. 
To improve optimization robustness under weak annotations, we replace the raw posterior estimate $p_\phi(y \!\mid\! \mathbf{x}, z)$ with a rectified target $\tilde{p}_\phi(y \!\mid\! \mathbf{x}, z)$. For each weak-annotation regime, the rectification exploits its specific structure, yielding a unified family of objectives that generalizes three regularizers:
\begin{equation}
\label{eq:cls_unified}
\begin{aligned}
\mathcal{L}_\text{Cls}(\theta) &= -\mathbb{E}_{\mathbf{x},z}\Big[\sum\nolimits_{y \in \mathcal{Y}} \tilde{p}_\phi(y \!\mid\! \mathbf{x}, z)\, \log [f_\theta(\mathbf{x})]_y\Big], \quad \text{with}\\[2pt]
\tilde{p}_\phi(y \!\mid\! \mathbf{x}, z) &= 
\begin{dcases}
\operatorname{sg}\!\big(\mathbb{I}(y\!=\!z) - [f_\theta(\mathbf{x})]_y\, (\delta - [f_\phi(\mathbf{x})]_y)/(1 - \delta)\big) & \text{(Noisy)} \\
\mathbb{I}(y \in z)\, [f_\phi(\mathbf{x})]_y \,\big/\, \textstyle\sum_{y' \in z} [f_\phi(\mathbf{x})]_{y'} & \text{(Ambiguous)} \\
\mathbb{I}(z \in \mathcal{Y})\, \mathbb{I}(y = z) + \mathbb{I}(z = \emptyset)\, [f_\phi(\mathbf{x})]_y & \text{(Semi-sup.)}
\end{dcases}
\end{aligned}
\end{equation}
For \textit{noisy labels}, $\operatorname{sg}(\cdot)$ denotes the stop-gradient operator and $\delta = \langle f_\theta(\mathbf{x}), f_\phi(\mathbf{x}) \rangle$ is the inner product between the current and EMA posteriors; the closed-form rectification is a correction term we derive to suppress noisy memorization (Appendix~\ref{exp:elr}), generalizing early-learning regularization~\citep{liu2020early}. For \textit{ambiguous labels}, our masked renormalization of $f_\phi(\mathbf{x})$ over the candidate set $z$ extends EMA-stabilized progressive identification~\citep{lv2020progressive}. For \textit{semi-supervised data}, our indicator switch reduces to mean-teacher self-training~\citep{tarvainen2017mean} on the unlabeled portion.

\subsection{Efficient Posterior Estimation via Optimized Timestep Sampling}
\label{subsec:timereduction}
Computing $p_\theta(y \!\mid\! \mathbf{x}_t)$ via Eq.~(\ref{eq:elbo_approx}) requires evaluating the conditional ELBO across all $T$ timesteps and classes, which is the dominant per-iteration cost. \citet{chen2024diffusion} alleviate this by sharing $\mathbf{x}_\tau$ across classes and sampling $\tau$ uniformly from $\{t+1,\dots,T-1\}$, yet Figure~\ref{fig:timestep_selection}(b) shows uniform sampling is far from efficient. The single-noise-level accuracy curve in Figure~\ref{fig:timestep_selection}(a) explains why: low noise levels $\sigma_\tau$ induce trivial reconstructions with low label sensitivity, while high noise levels are noise-dominated and unreliable, so only an intermediate band carries meaningful class signal.

\textbf{Optimal Timestep Sampling.}
We therefore seek a contiguous sub-range $\{l, \dots, r\}$ of fixed width $\Delta = r - l$ such that its weighted average approximates the full sum:
\begin{equation}
\label{eq:range}
    \Big\|
        \frac{1}{\Delta}\sum_{\tau=l}^{r} w_\tau \hbar(\tau,y)
        - \frac{1}{T\!-\!1}\sum_{\tau=1}^{T-1} w_\tau \hbar(\tau,y)
    \Big\|_2^2.
\end{equation}
Since $\hbar(\tau, y)$ is data-dependent and evolves during training, directly optimizing Eq.~(\ref{eq:range}) is impractical. We instead adopt a $\hbar$-agnostic surrogate: choose the sub-range $\{l, \dots, r\}$ that maximizes the total weight $\sum_{\tau=l}^{r} w_\tau$, so that the partial sum closely approximates the full sum, as illustrated in Figure~\ref{fig:timestep_selection}(a) (orange areas: the partial sum; gray areas: the full sum). For the EDM log-normal weight, this maximization is modeled through the interval-mass objective and solved numerically, yielding a window centered around the median region of the weight density, as formalized below.

\begin{theorem}[Median-Centered Interval via Weight Mass]
\label{theorem:lr}
For the EDM log-normal weight $w(\tau)$ with cumulative weight $\mathcal{F}_w(b) = \int_{0}^{b} w(\tau) \,\mathrm{d}\tau$ and a fixed length $\Delta$, the interval $[l, r]$ that maximizes the covered weight mass $\mathcal{F}_w(r) - \mathcal{F}_w(l)$ is uniquely determined by
\begin{equation}
    \mathcal{F}_w(l) + \mathcal{F}_w(l + \Delta) = 1, \qquad r = l + \Delta.
\end{equation}
\end{theorem}

The optimal $l$ is solved by numerical root-finding~\citep{brent2013algorithms}. As $\Delta \!\to\! 0$ the interval collapses to the log-normal weight-density median $e^{-1.2}$ ($\log \sigma_t=-1.2$) and matches Figure~\ref{fig:timestep_selection}(a). The proof and an analogous result for DDPM~\citep{ho2020denoising} are given in Appendix~\ref{proof:lr}.

\section{Experiments}
\label{sec:experiments}

We evaluate \textsl{DELTA} on two categories of datasets.
\textbf{Synthetic benchmarks} apply controlled corruptions to Fashion-MNIST~\citep{xiao2017fashion}, CIFAR-10~\citep{krizhevsky2009learning}, and ImageNette~\citep{5206848}.
\textbf{Real-world benchmarks} use naturally occurring weak annotations: the noisy-label sets CIFAR10-N and CIFAR100-N~\citep{wei2022learning}, and the ambiguous-label set PLCIFAR10~\citep{wang2025realistic}.
All models are trained from scratch under the EDM framework; full implementation details are in Appendix~\ref{sec:implementation}.
As an internal ablation, \textsl{Vanilla} uses the same joint objective structure as \textsl{DELTA} but replaces the posterior-weighted generative term (Eq.~\ref{eq:WDSM}) with a naive formulation that trains directly on weak pairs $(\mathbf{x}, z)$ without posterior reweighting.

\textbf{Weak Annotation Protocols for Synthetic Benchmarks.}
We corrupt synthetic benchmarks as follows:
(1) \textbf{Noisy Label}: labels are flipped with 40\% probability via symmetric noise ({`Sym-40\%'}, uniform flipping to any incorrect class) or asymmetric noise ({`Asym-40\%'}, flipping to semantically similar classes);
(2) \textbf{Ambiguous Label}: candidate sets are generated via a random policy ({`Random'}, including non-target labels with 50\% probability) or a class-dependent policy ({`Class-Dept.'}, selecting semantically similar labels with specified probabilities);
(3) \textbf{Semi-Supervised Data}: labeled samples are retained at a fixed fraction per class, denoted as {`1\% Labels'} and {`10\% Labels'}.

\subsection{Image Generation}
\label{subsec:exp_gen}

\textbf{Setup.}
We assess generation quality with four unconditional metrics: FID, IS, Density (abbreviated as `Dens.'), and Coverage (abbreviated as `Cov.')~\citep{chaodenoising}, and three class-conditional metrics: CW-FID, CW-Density, and CW-Coverage~\citep{kaneko2019label}.
Detailed metric definitions are in Appendix~\ref{exp:metrics}.

\textbf{Comparison with Vanilla.}
Table~\ref{tab:gen_results} reports generation quality across all six weak annotation settings on CIFAR-10 and ImageNette.
\textsl{DELTA} consistently outperforms \textsl{Vanilla} on both unconditional and conditional metrics.
The improvement is most pronounced under ambiguous-label settings, where the naive posterior collapses to a mixture, causing class-image mismatch and per-class homogenization. For example, \textsl{Vanilla}'s CW-FID on CIFAR-10 (random ambiguous) is 27.18 versus \textsl{DELTA}'s 10.65.
Across all settings, \textsl{DELTA}'s results approach the clean-label upper bound (gray column), confirming that posterior-weighted score matching effectively resolves the mixture bias.

\begin{figure*}[!t]
    \centering
    \begin{minipage}[b]{0.3\textwidth}
        \centering
        \includegraphics[width=\linewidth]{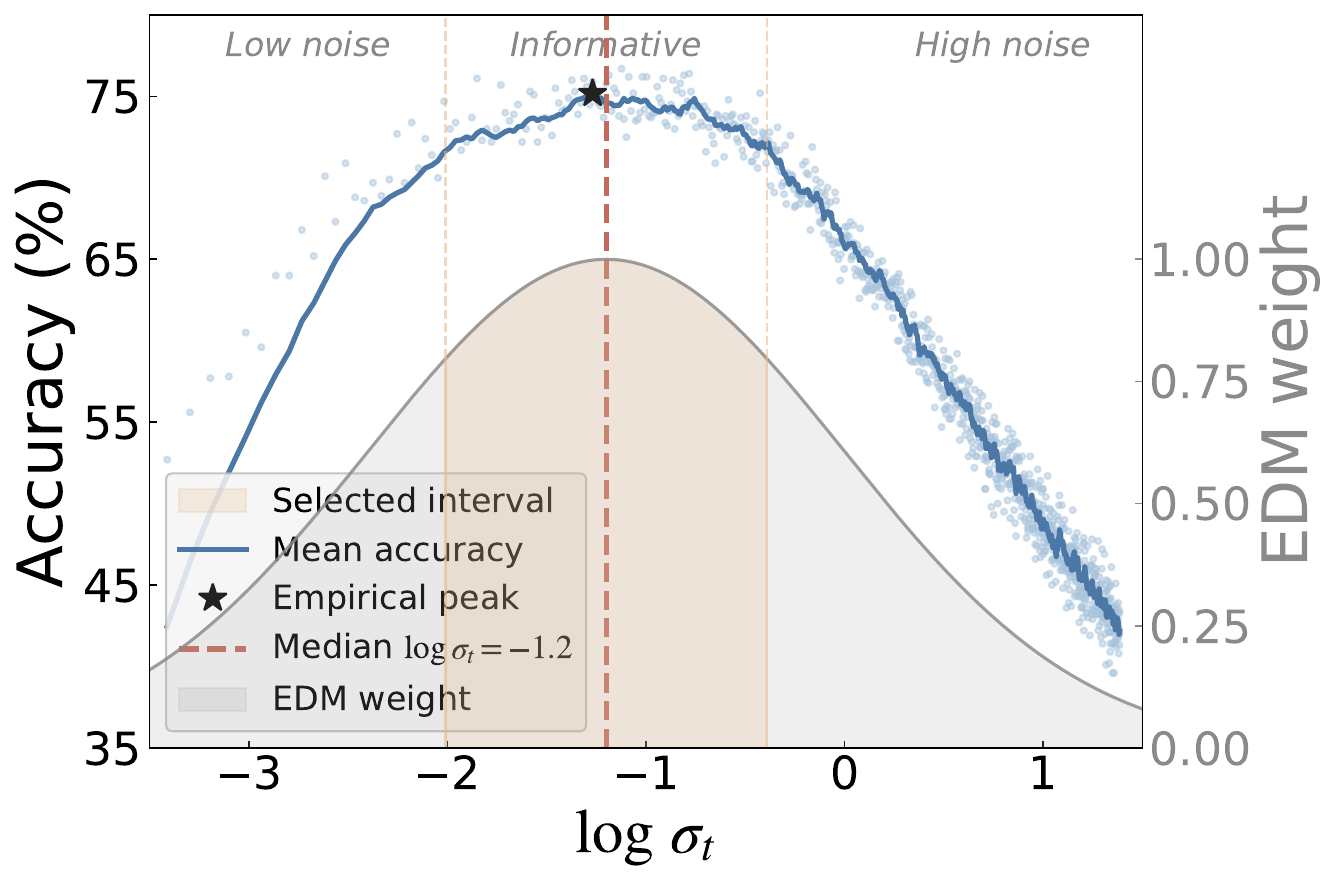}
        \centering    
        \footnotesize (a) EDM weight \& Acc. distribution.
    \end{minipage}
    \hspace{10pt}
    \begin{minipage}[b]{0.3\textwidth}
        \includegraphics[width=\linewidth]{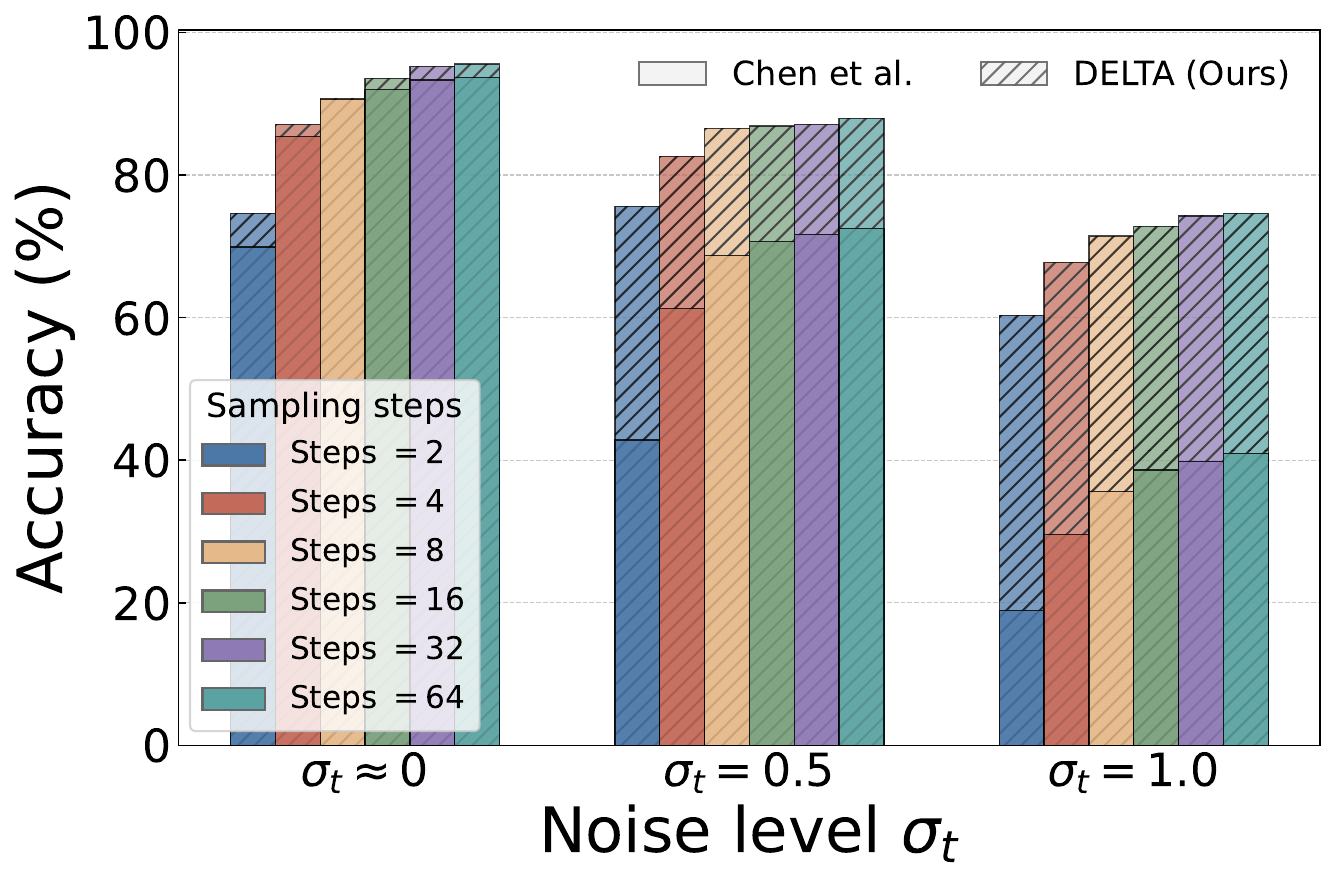}
        \centering    
        \footnotesize (b) Accuracy across sampling budgets. \!\!\!\!\!\!\!\!\!
    \end{minipage}
    \hspace{10pt}
    \begin{minipage}[b]{0.3\textwidth}
        \centering
        \includegraphics[width=\linewidth]{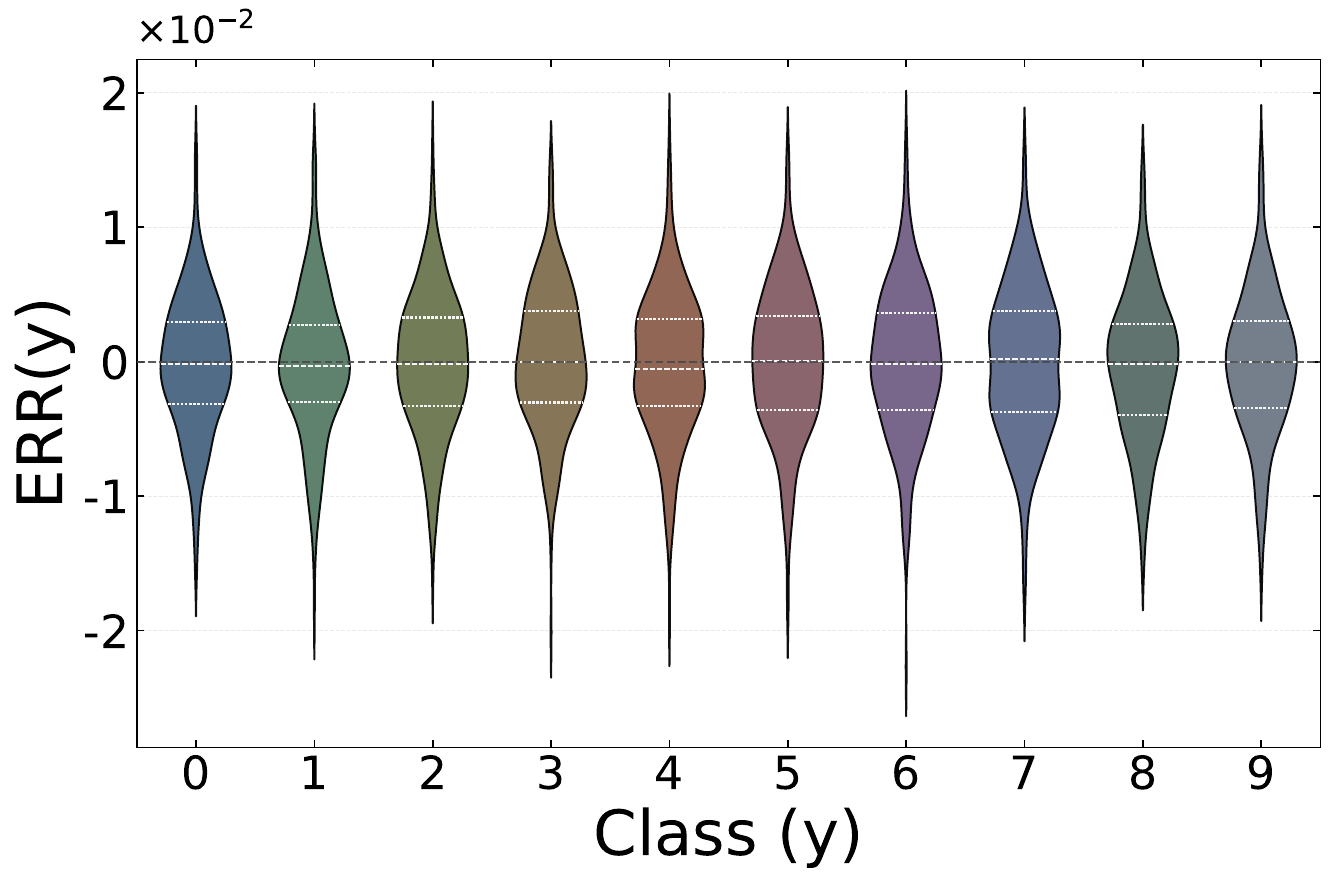}
        \centering    
        \footnotesize (c) Class-wise $\textsc{Err}$ distribution. \!\!\!\!\!\!\!\!\!\!\!\!\!
    \end{minipage}
    \caption{Noise-level sampling analysis on CIFAR-10.
    (a): Single-timestep accuracy (blue) and EDM weight (gray).
    (b): Our median-centered scheme outperforms the uniform sampling of \citet{chen2024diffusion} at all budgets.
    (c): Violin distribution of the optimality residual $\textsc{Err}$; closer to zero is better.
    }
    \label{fig:timestep_selection}
\end{figure*}

\noindent\textbf{Comparison with Noise-Robust Diffusion Baselines.} Table~\ref{tab:noise_baseline} benchmarks \textsl{DELTA} against four dedicated noise-robust diffusion methods (CAD~\citep{dufour2024don}, TDSM~\citep{na2024label}, SBDC~\citep{cong2025guiding}, and RTDC~\citep{chen2025robust}) on CIFAR-10 with 40\% noisy labels.
These baselines each assume access to task-specific auxiliary supervision such as transition matrices, oracle confidence scores, or auxiliary discriminators, whereas \textsl{DELTA} operates without such priors.
Despite this strictly weaker assumption, \textsl{DELTA} achieves the best results on nearly all metrics, and in particular dominates the class-conditional metrics that directly measure class--image alignment, demonstrating that our posterior-weighted formulation is a practically stronger alternative.
Notably, \textsl{Vanilla} attains a competitive unconditional FID under Sym-40\% yet the worst CW-FID, confirming that unconditional metrics alone are insensitive to the label-condition mismatch that robust training aims to resolve.

\begin{figure}[!t]
    \centering
    \begin{minipage}[t]{0.235\linewidth}
        \centering
        \includegraphics[width=\linewidth]{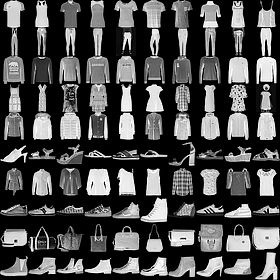}
    \end{minipage}\hfill
    \begin{minipage}[t]{0.235\linewidth}
        \centering
        \includegraphics[width=\linewidth]{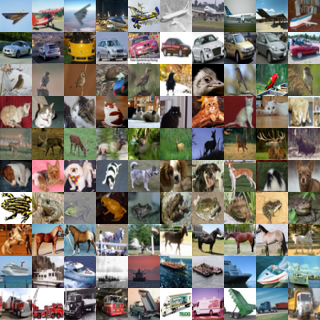}
    \end{minipage}\hfill
    \begin{minipage}[t]{0.235\linewidth}
        \centering
        \includegraphics[width=\linewidth]{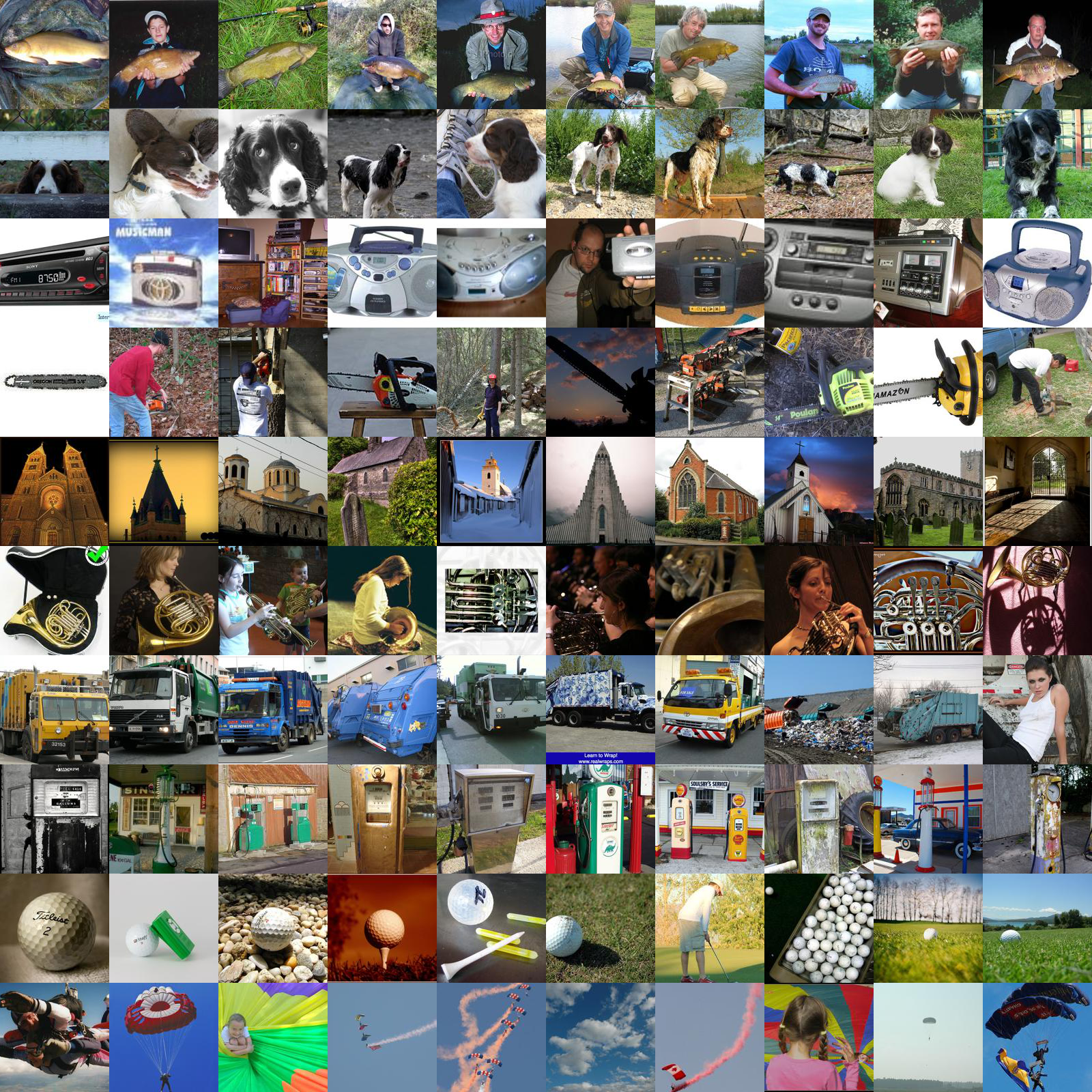}
    \end{minipage}\hfill
    \begin{minipage}[t]{0.235\linewidth}
        \centering
        \raisebox{-4mm}[0pt][0pt]{\includegraphics[width=\linewidth]{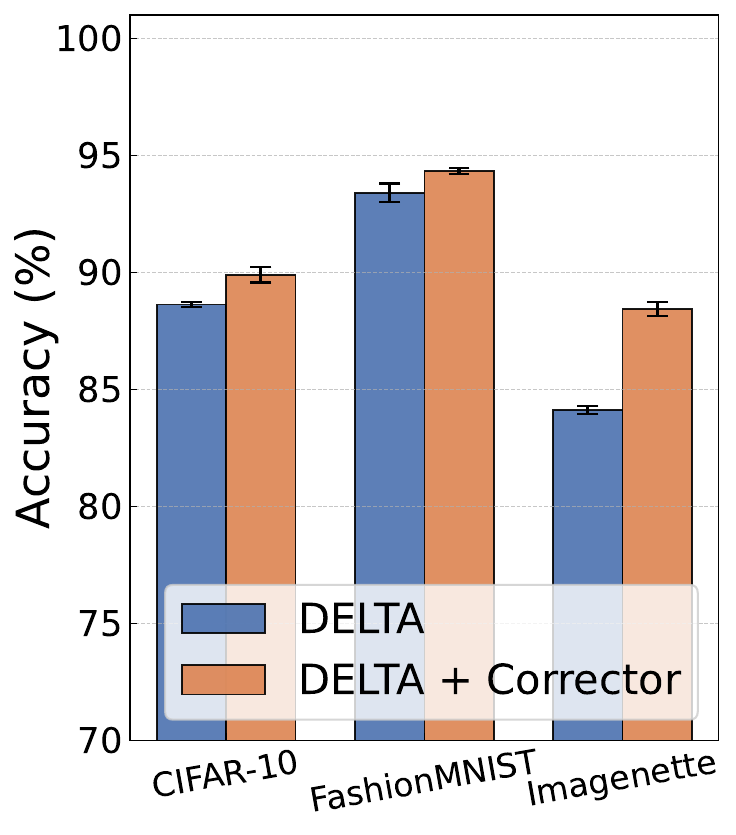}}
    \end{minipage}
    \caption{The three left panels show random samples from \textsl{DELTA} trained under symmetric $40\%$ label noise (Sym-40\%), on Fashion-MNIST, CIFAR-10, and ImageNette (left to right). The rightmost panel reports classification accuracy of \textsl{DELTA} with and without pseudo-label correction.}
    \label{fig:qualitative}
\end{figure}

\begin{table}[!t]
    \centering
    \scriptsize
    \renewcommand{\arraystretch}{0.9}
    \caption{Comparison against noise-robust diffusion baselines on CIFAR-10.}
    \label{tab:noise_baseline}
    \resizebox{\textwidth}{!}{%
    \begin{tabular}{l|ccccccc|ccccccc}
    \toprule
    & \multicolumn{7}{c|}{Sym-40\%} & \multicolumn{7}{c}{Asym-40\%} \\
    \cmidrule(lr){2-8} \cmidrule(lr){9-15}
    Method & FID$\downarrow$ & IS$\uparrow$ & Dens.$\uparrow$ & Cov.$\uparrow$
           & CW-FID$\downarrow$ & CW-Dens.$\uparrow$ & CW-Cov.$\uparrow$
           & FID$\downarrow$ & IS$\uparrow$ & Dens.$\uparrow$ & Cov.$\uparrow$
           & CW-FID$\downarrow$ & CW-Dens.$\uparrow$ & CW-Cov.$\uparrow$ \\
    \midrule
    \textsl{Vanilla} & \textbf{3.33} & 9.56 & 101.39 & 81.12 & 29.84 & 72.98 & 73.39
                     & 3.23 & 9.02 & 100.06 & 80.71 & 14.70 & 90.85 & 79.63 \\
    \midrule
    \textsl{CAD}~\citep{dufour2024don}  & 4.10 & 9.08 & 101.48 & 77.04 & 19.66 & 93.51 & 76.43
                                        & 3.87 & 9.16 & 98.27 & 75.81 & 14.33 & 99.25 & 78.12 \\
    \textsl{TDSM}~\citep{na2024label} & 3.85 & 9.40 & 106.32 & 79.66 & 16.42 & 101.27 & 78.63
                                      & 3.96 & \textbf{10.12} & 107.85 & 80.47 & 14.96 & 102.67 & 78.99 \\
    \textsl{SBDC}~\citep{cong2025guiding} & 4.25 & 8.95 & 97.83  & 74.92 & 21.30 & 89.83 & 75.37
                                          & 4.02 & 9.05 & 95.14 & 73.29 & 15.70 & 96.14 & 77.25 \\
    \textsl{RTDC}~\citep{chen2025robust} & 4.00 & 9.20 & 104.51 & 78.35 & 17.92 & 98.13 & 77.70
                                         & 3.85 & 9.30 & 103.63 & 78.06 & 13.92 & 101.03 & 78.68 \\
    \midrule
    \textsl{DELTA} & 3.47 & \textbf{9.68} & \textbf{109.75} & \textbf{81.21} & \textbf{13.85} & \textbf{107.23} & \textbf{80.11}
                   & \textbf{3.10} & 9.73 & \textbf{109.69} & \textbf{81.30} & \textbf{13.24} & \textbf{107.07} & \textbf{79.65} \\
    \bottomrule
    \end{tabular}}
\end{table}

\begin{wraptable}{r}{0.66\textwidth}
\vspace{-8pt}
\centering
\scriptsize
\begin{minipage}[t]{0.34\textwidth}
    \centering
    \tabcolsep 0.03in
    \renewcommand{\arraystretch}{1.05}
    \caption{Performance on real-world weakly annotated datasets.}
    \label{tab:complex_supervision}
    \begin{tabular}{l|cccc}
    \toprule
    Dataset & FID$\downarrow$ & IS$\uparrow$ & Dens.$\uparrow$ & Cov.$\uparrow$ \\
    \midrule
    PLCIFAR10  & 2.95 & 9.82  & 109.1 & 81.7 \\
    CIFAR10-N  & 3.22 & 9.66  & 108.3 & 80.9 \\
    CIFAR100-N & 4.85 & 10.42 & 96.4  & 71.2 \\
    \bottomrule
    \end{tabular}
\end{minipage}
\hfill
\begin{minipage}[t]{0.3\textwidth}
    \centering
    \tabcolsep 0.06in
    \renewcommand{\arraystretch}{1.05}
    \caption{Inference latency (CIFAR-10, single A100).}
    \label{tab:efficiency}
    \begin{tabular}{l|cc}
    \toprule
    Metric & Chen et al. & \textsl{DELTA} \\
    \midrule
    Time (min) & 18.43 & \textbf{0.67} \\
    Acc (\%)   & 87.13 & \textbf{88.63} \\
    Speedup    & 1$\times$ & \textbf{27.5$\times$} \\
    \bottomrule
    \end{tabular}
\end{minipage}
\vspace{-2pt}
\end{wraptable}

\noindent\textbf{Performance on Real-World Weak Annotations.} Table~\ref{tab:complex_supervision} shows that under real-world weak annotations (CIFAR10-N, PLCIFAR10, CIFAR100-N), \textsl{DELTA} achieves low FID, competitive IS scores, and strong generation quality, validating its effectiveness on real-world datasets.

\textbf{Qualitative Results.}
The three left panels of Figure~\ref{fig:qualitative} show random class-conditional samples from \textsl{DELTA} trained under symmetric $40\%$ label noise. Although the {training} labels are heavily corrupted, the {generated} images remain visually consistent with the class condition, preserve distinct modes rather than homogenizing, and retain clear structure instead of degrading into low-quality outputs.

\subsection{Probing the Learned Posterior}
\label{subsec:analysis}

In this subsection, we examine the quality of the learned class posteriors $p_\phi(y \!\mid\! \mathbf{x}, z)$ through two lenses: its evolution during training and its accuracy as a standalone classifier.

\begin{figure*}[!t]
    \centering
    \begin{minipage}[b]{0.16\textwidth}
        \centering
        \includegraphics[width=\linewidth]{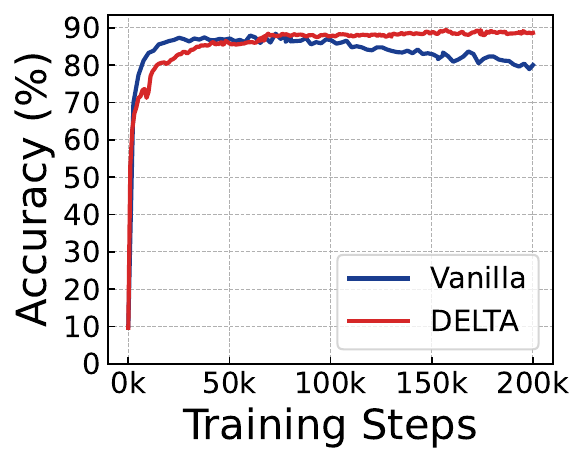}
        \scriptsize (a) NL Sym-40\%
    \end{minipage}
    \hfill
    \begin{minipage}[b]{0.16\textwidth}
        \centering
        \includegraphics[width=\linewidth]{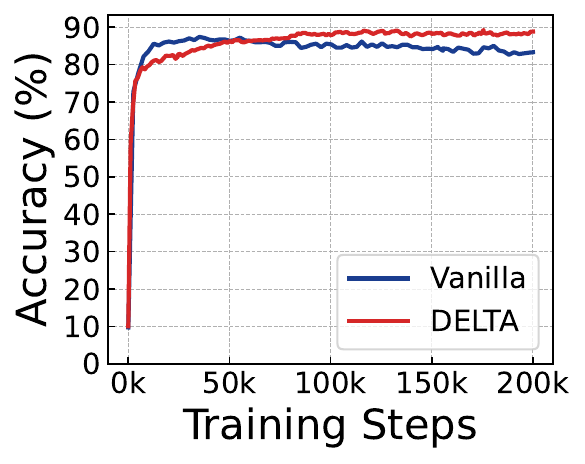}
        \scriptsize (b) NL Asym-40\%
    \end{minipage}
    \hfill
    \begin{minipage}[b]{0.16\textwidth}
        \centering
        \includegraphics[width=\linewidth]{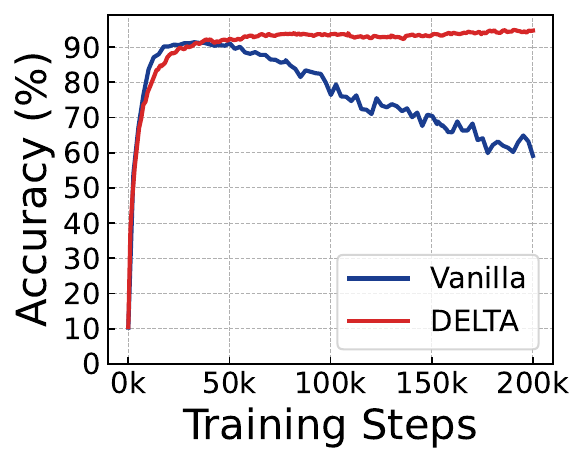}
        \scriptsize (c) AL Random
    \end{minipage}
    \hfill
    \begin{minipage}[b]{0.16\textwidth}
        \centering
        \includegraphics[width=\linewidth]{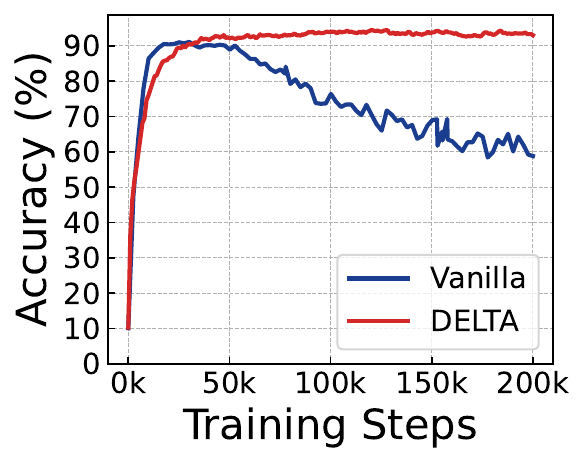}
        \scriptsize (d) AL Class-Dept.
    \end{minipage}
    \hfill
    \begin{minipage}[b]{0.16\textwidth}
        \centering
        \includegraphics[width=\linewidth]{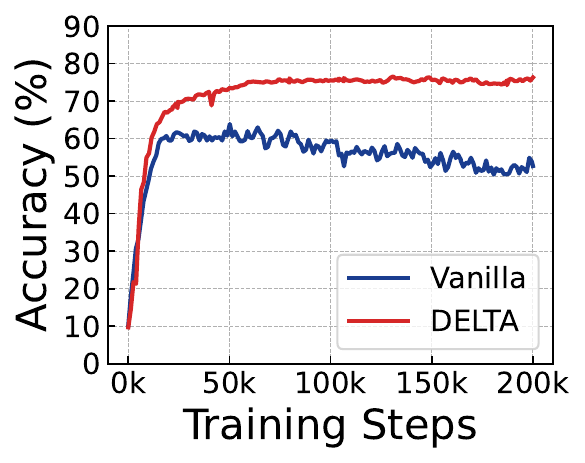}
        \scriptsize (e) SS 1\% Labels
    \end{minipage}
    \hfill
    \begin{minipage}[b]{0.16\textwidth}
        \centering
        \includegraphics[width=\linewidth]{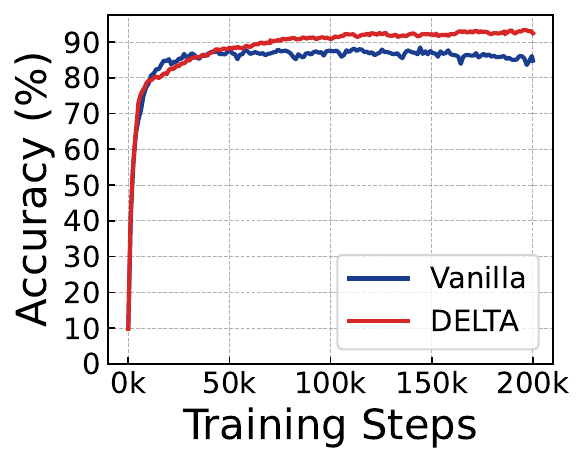}
        \scriptsize (f) SS 10\% Labels
    \end{minipage}
    \caption{Test accuracy curves of \textsl{Vanilla} versus \textsl{DELTA} on CIFAR-10 under various weak annotations: Noisy Labels (NL), Ambiguous Labels (AL), and Semi-Supervised data (SS). }
    \label{fig:curves}
\end{figure*}

\textbf{Evolution of Learned Class Posteriors.}
Figure~\ref{fig:curves} traces the classification accuracy of $p_\phi$ throughout training on CIFAR-10 under all six weak annotation settings.
\textsl{Vanilla}'s accuracy rises initially before degrading, reflecting that a naive posterior collapses under sustained label noise.
In contrast, \textsl{DELTA}'s posterior improves monotonically and stabilizes, validating the EM-style refinement described in Section~\ref{sec:method}: a better generative model yields a more accurate $p_\phi$, which in turn guides a better update.

\textbf{Posterior as a Classifier.}
A converged $p_\phi$ can be directly evaluated as a discriminative classifier.
Table~\ref{tab:weakly} compares \textsl{DELTA} against advanced specialized discriminative baselines across all three weak annotation types on Fashion-MNIST and CIFAR-10.
For fairness, all discriminative classifiers use Wide-ResNet-40-10 (55.84M parameters), matching the parameter count of our diffusion model (55.73M), and all models are trained from scratch.
Despite being a generative framework, \textsl{DELTA} consistently achieves the best accuracy across all settings, demonstrating that diffusion-based posterior inference is a strong discriminative signal in its own right.

\begin{table*}[!t]
    \centering
    \scriptsize
    \tabcolsep 0.128in
    \renewcommand{\arraystretch}{0.5}
    \caption{Classification results (test accuracy, \%) under various weak annotations {($\spadesuit$: noisy label, $\heartsuit$: ambiguous label, $\clubsuit$: semi-supervised data)}. \textbf{Bold} denotes best performance.}
    \label{tab:weakly}
    \resizebox{\textwidth}{!}{%
    \begin{tabular}{c|c|ccccccccccccc}
    \toprule[1pt]
    $\text{Dataset}$ & Type & \textsl{CE} & \textsl{Mixup}~\citep{zhang2018mixup} & \textsl{Coteaching}~\citep{han2018co} & \textsl{ELR}~\citep{liu2020early} & \textsl{PENCIL}~\citep{yi2022pencil} & \textsl{Vanilla} & \textsl{DELTA} \\
    \midrule
    \multirow{2}{*}{$\text{Fashion-MNIST}^{\spadesuit}$}
    & {Sym-40\%} & 76.18\tiny{$\pm$0.26} & 92.21\tiny{$\pm$0.03} & 92.17\tiny{$\pm$0.34} & 93.13\tiny{$\pm$0.13} & 90.85\tiny{$\pm$0.58} & 90.11\tiny{$\pm$1.24} & \textbf{93.40\tiny{$\pm$0.40}} \\
    & {Asym-40\%} & 82.01\tiny{$\pm$0.06} & 92.01\tiny{$\pm$1.02} & 92.78\tiny{$\pm$0.25} & 92.82\tiny{$\pm$0.09} & 91.77\tiny{$\pm$0.69} & 85.41\tiny{$\pm$0.96} & \textbf{93.20\tiny{$\pm$0.30}} \\
    \midrule
    \multirow{2}{*}{$\text{CIFAR-10}^{\spadesuit}$}
    & {Sym-40\%} & 67.22\tiny{$\pm$0.26} & 84.26\tiny{$\pm$0.64} & 86.54\tiny{$\pm$0.57} & 85.68\tiny{$\pm$0.13} & 85.91\tiny{$\pm$0.26} & 80.22\tiny{$\pm$0.10}& \textbf{88.63\tiny{$\pm$0.12}} \\
    & {Asym-40\%} & 76.98\tiny{$\pm$0.42} & 83.21\tiny{$\pm$0.85} & 79.38\tiny{$\pm$0.39} & 81.32\tiny{$\pm$0.31} & 84.89\tiny{$\pm$0.49}  & 86.31\tiny{$\pm$0.10} & \textbf{88.83\tiny{$\pm$0.33}} \\
    \midrule[1pt]
     & Type & \textsl{PRODEN}~\citep{lv2020progressive} & \textsl{IDGP}~\citep{qiaodecompositional} & \textsl{PiCO}~\citep{wang2022pico+} & \textsl{CRDPLL}~\citep{wu2022revisiting} & \textsl{DIRK}~\citep{wu2024distilling} & \textsl{Vanilla} & \textsl{DELTA} \\
    \midrule
    \multirow{2}{*}{$\text{Fashion-MNIST}^{\heartsuit}$}
    & {Random} & {93.31}\tiny{$\pm$0.07}  & {92.26}\tiny{$\pm$0.25} & {93.32}\tiny{$\pm$0.12} & {94.03}\tiny{$\pm$0.14} & {94.11}\tiny{$\pm$0.22} & {80.20}\tiny{$\pm$1.29} & \textbf{{94.27}\tiny{$\pm$0.55}} \\
    & {Class-Dept.} & {93.44}\tiny{$\pm$0.21} & {93.07}\tiny{$\pm$0.16}  & {93.32}\tiny{$\pm$0.33} & {93.80}\tiny{$\pm$0.23} & {93.99}\tiny{$\pm$0.24} & {66.03}\tiny{$\pm$1.43} & \textbf{{94.20}\tiny{$\pm$0.15}} \\
    \midrule
    \multirow{2}{*}{$\text{CIFAR-10}^{\heartsuit}$}
    & {Random} & {90.02}\tiny{$\pm$0.22} & {89.65}\tiny{$\pm$0.53}   & {86.40}\tiny{$\pm$0.89} & {92.74}\tiny{$\pm$0.26} & {93.48}\tiny{$\pm$0.14} & {60.25}\tiny{$\pm$0.17}& \textbf{{94.70}\tiny{$\pm$0.49}} \\
    & {Class-Dept.} & {90.44}\tiny{$\pm$0.44} & {90.83}\tiny{$\pm$0.34}   &{87.51}\tiny{$\pm$0.66} & {92.89}\tiny{$\pm$0.27} & {93.22}\tiny{$\pm$0.37} &  {56.34}\tiny{$\pm$0.50}& \textbf{{93.53}\tiny{$\pm$0.12}} \\
    \midrule[1pt]
    & Type & \textsl{Dash}~\citep{xu2021dash} & \textsl{CoMatch}~\citep{li2021comatch} & \textsl{FlexMatch}~\citep{zhang2021flexmatch} & \textsl{SimMatch}~\citep{zheng2022simmatch} & \textsl{SoftMatch}~\citep{chensoftmatch} & \textsl{Vanilla} & \textsl{DELTA} \\
    \midrule
    \multirow{2}{*}{$\text{Fashion-MNIST}^{\clubsuit}$}& {{1\% Labels}} 
    & {84.73}\tiny{$\pm$0.09} 
    & {85.31}\tiny{$\pm$0.29} 
    & {84.43}\tiny{$\pm$0.30} 
    & {84.69}\tiny{$\pm$0.17} 
    & {84.72}\tiny{$\pm$0.23} 
    & {78.37}\tiny{$\pm$0.72} 
    & \textbf{{85.92}\tiny{$\pm$0.13}} \\
    & {{10\% Labels}} 
    & {91.16}\tiny{$\pm$0.20} 
    & {90.52}\tiny{$\pm$0.12} 
    & {90.69}\tiny{$\pm$0.03} 
    & {91.18}\tiny{$\pm$0.13} 
    & {91.22}\tiny{$\pm$0.11} 
    & {90.50}\tiny{$\pm$1.00} 
    & \textbf{{92.97}\tiny{$\pm$0.21}} \\
    \midrule
    \multirow{2}{*}{$\text{CIFAR-10}^{\clubsuit}$}& {{1\% Labels}} & {70.14}\tiny{$\pm$0.69} & {61.45}\tiny{$\pm$1.46} & {70.72}\tiny{$\pm$0.93} & {73.33}\tiny{$\pm$1.02} & {73.74}\tiny{$\pm$0.82}  & {53.49}\tiny{$\pm$0.15} & \textbf{{76.40}\tiny{$\pm$0.54}} \\
    & {{10\% Labels}} & {81.50}\tiny{$\pm$0.68} & {77.79}\tiny{$\pm$0.53} & {81.35}\tiny{$\pm$0.48} & {82.90}\tiny{$\pm$0.43} & {88.66}\tiny{$\pm$0.60} &  {85.13}\tiny{$\pm$0.12} & \textbf{{92.47}\tiny{$\pm$0.39}} \\
    \bottomrule[1pt]
    \end{tabular}}
    \end{table*}

\subsection{Inference Efficiency of Timestep Sampling}
\label{subsec:efficiency}
Section~\ref{subsec:timereduction} proposes a median-centered interval timestep sampling strategy that concentrates posterior estimation on the most informative noise levels.
We validate its efficiency and effectiveness empirically.
Table~\ref{tab:efficiency} confirms that it reduces per-step posterior estimation cost by $27.5\times$ over full-timestep computation~\citep{chen2024diffusion} with no accuracy loss.
Figure~\ref{fig:timestep_selection}(b) further shows that the median-centered scheme consistently outperforms uniform sampling at every step budget, with the gap widening as the budget shrinks.
Figure~\ref{fig:timestep_selection}(c) reports the class-wise residual $\textsc{Err}(y)$ as a violin plot, where $\textsc{Err}(y)$ measures the deviation from the necessary condition of Eq.~(\ref{eq:range}) in Appendix~\ref{app:necessary_condition_err}; smaller absolute values indicate less deviation.
The distribution is tightly concentrated near zero across all classes, confirming that the median-centered timestep sampling strategy achieves effective coverage in practice.

\subsection{Synergy with Pseudo-label Correction}
\label{subsec:corrector}

Our generative framework is orthogonal to discriminative label correction strategies.
To validate this complementarity, we use DivideMix~\citep{li2020dividemix} to assign pseudo-labels to the noisy-label training set and then train \textsl{DELTA} on those pseudo-labeled examples.
Pseudo-labels are never fully reliable, so residual label noise remains and the supervision stays noisy.
As shown in the rightmost panel of Figure~\ref{fig:qualitative} on CIFAR-10 under symmetric $40\%$ label noise, \textsl{DELTA} consistently improves upon the corrector's standalone performance, validating that our method effectively complements existing approaches by acting as a robust denoiser for residual noise.

\section{Discussion and Limitations}
\label{sec:discussion}

Beyond the experiments above, a potential extension of \textsl{DELTA} is \emph{weakly supervised} dataset condensation~\cite{yu2023dataset}, which aims to distill large weakly annotated datasets into clean and compact synthetic counterparts for efficient downstream training, such as neural architecture search~\cite{zhao2021dataset,du2023minimizing}.
Most prior methods~\cite{zhao2023improved,he2024multisize,lee2022dataset,kim2022dataset,MTT} assume clean labels.
By contrast, \textsl{DELTA} acts as a \emph{generative compressor}, compressing the original, weakly annotated training data into a model that can synthesize clean, class-representative samples on demand.
Appendix~\ref{sec:condensation} provides further discussion and experiments.

Our limitations are mainly twofold.
On one hand, we assume a fixed and finite label space; we therefore do not treat unseen categories at test time or settings with a gradually expanding label space, and extending \textsl{DELTA} to such cases is left for future work.
On the other hand, our theory relies on class-conditional weak annotations, where the annotation mechanism depends on the latent class but not directly on instance-specific visual factors.
In practice, data-collection pipelines may also depend on image difficulty or style.
Relaxing this assumption would be a useful direction for future work.

\section{Conclusion}

We presented \textsl{DELTA}, a unified framework for training LCDMs under weak annotations. By treating the unobserved clean label as a latent variable, \textsl{DELTA} optimizes a single variational objective that couples posterior-weighted score matching with diffusion-based posterior refinement, allowing generation and label inference to reinforce each other within one model. A median-centered timestep sampling strategy further makes the per-sample posterior estimation tractable during joint training. Unlike prior approaches that target a single form of weak annotation and rely on auxiliary priors, \textsl{DELTA} handles all weak annotations within one objective while requiring less prior knowledge, offering a step toward making diffusion models broadly robust to imperfect supervision.

\section*{Acknowledgments}
DDW was supported by the RIKEN Junior Research Associate Program.
MS was supported by JST ASPIRE Grant Number JPMJAP2405.

{
\small
\bibliographystyle{unsrtnat}
\bibliography{reference}

\begin{thebibliography}{124}
\providecommand{\natexlab}[1]{#1}
\providecommand{\url}[1]{\texttt{#1}}
\expandafter\ifx\csname urlstyle\endcsname\relax
  \providecommand{\doi}[1]{doi: #1}\else
  \providecommand{\doi}{doi: \begingroup \urlstyle{rm}\Url}\fi

\bibitem[Ho et~al.(2020)Ho, Jain, and Abbeel]{ho2020denoising}
Jonathan Ho, Ajay Jain, and Pieter Abbeel.
\newblock Denoising diffusion probabilistic models.
\newblock In \emph{Advances in Neural Information Processing Systems},
  volume~33, pages 6840--6851, 2020.

\bibitem[Song et~al.(2021)Song, Sohl-Dickstein, Kingma, Kumar, Ermon, and
  Poole]{songscore}
Yang Song, Jascha Sohl-Dickstein, Diederik~P. Kingma, Abhishek Kumar, Stefano
  Ermon, and Ben Poole.
\newblock Score-based generative modeling through stochastic differential
  equations.
\newblock In \emph{International Conference on Learning Representations}, 2021.

\bibitem[Fuest et~al.(2026)Fuest, Ma, Gui, Schusterbauer, Hu, and
  Ommer]{fuest2026diffusion}
Michael Fuest, Pingchuan Ma, Ming Gui, Johannes Schusterbauer, Vincent~Tao Hu,
  and Bj{\"o}rn Ommer.
\newblock Diffusion models and representation learning: {A} survey.
\newblock \emph{{IEEE} Transactions on Pattern Analysis and Machine
  Intelligence}, 48\penalty0 (7):\penalty0 7209--7228, 2026.

\bibitem[Rombach et~al.(2022)Rombach, Blattmann, Lorenz, Esser, and
  Ommer]{rombach2022high}
Robin Rombach, Andreas Blattmann, Dominik Lorenz, Patrick Esser, and Bj{\"o}rn
  Ommer.
\newblock High-resolution image synthesis with latent diffusion models.
\newblock In \emph{Proceedings of the IEEE/CVF Conference on Computer Vision
  and Pattern Recognition}, pages 10674--10685, 2022.

\bibitem[Ho et~al.(2022)Ho, Salimans, Gritsenko, Chan, Norouzi, and
  Fleet]{ho2022video}
Jonathan Ho, Tim Salimans, Alexey Gritsenko, William Chan, Mohammad Norouzi,
  and David~J. Fleet.
\newblock Video diffusion models.
\newblock In \emph{Advances in Neural Information Processing Systems},
  volume~35, pages 8633--8646, 2022.

\bibitem[Wu et~al.(2025)Wu, Zhu, Liu, Zhao, Zhai, Cao, and
  Zha]{wu2024improvedvideovaelatent}
Pingyu Wu, Kai Zhu, Yu~Liu, Liming Zhao, Wei Zhai, Yang Cao, and Zheng-Jun Zha.
\newblock Improved video {VAE} for latent video diffusion model.
\newblock In \emph{Proceedings of the IEEE/CVF Conference on Computer Vision
  and Pattern Recognition}, pages 18124--18133, 2025.

\bibitem[Fei et~al.(2025)Fei, Li, Yang, Gao, Xu, Ma, Yang, and
  Zhou]{fei2025diffusion}
Ben Fei, Yixuan Li, Weidong Yang, Hengjun Gao, Jingyi Xu, Lipeng Ma, Yatian
  Yang, and Pinghong Zhou.
\newblock A diffusion model for universal medical image enhancement.
\newblock \emph{Communications Medicine}, 5\penalty0 (1):\penalty0 294, 2025.

\bibitem[Esser et~al.(2024)Esser, Kulal, Blattmann, Entezari, M{\"u}ller,
  Saini, Levi, Lorenz, Sauer, Boesel, Podell, Dockhorn, English, and
  Rombach]{esser2024scaling}
Patrick Esser, Sumith Kulal, Andreas Blattmann, Rahim Entezari, Jonas
  M{\"u}ller, Harry Saini, Yam Levi, Dominik Lorenz, Axel Sauer, Frederic
  Boesel, Dustin Podell, Tim Dockhorn, Zion English, and Robin Rombach.
\newblock Scaling rectified flow transformers for high-resolution image
  synthesis.
\newblock In \emph{Proceedings of the 41st International Conference on Machine
  Learning}, volume 235, pages 12606--12633, 2024.

\bibitem[Cao et~al.(2026)Cao, Zhou, Song, and Yang]{cao2025controllable}
Pu~Cao, Feng Zhou, Qing Song, and Lu~Yang.
\newblock Controllable generation with text-to-image diffusion models: {A}
  survey.
\newblock \emph{{IEEE} Transactions on Pattern Analysis and Machine
  Intelligence}, 48\penalty0 (4):\penalty0 4771--4791, 2026.

\bibitem[Ho and Salimans(2022)]{ho2022classifier}
Jonathan Ho and Tim Salimans.
\newblock Classifier-free diffusion guidance.
\newblock \emph{arXiv preprint arXiv:2207.12598}, 2022.

\bibitem[Shenoy et~al.(2026)Shenoy, Pan, Balakrishnan, Cheng, Jeon, Yang, and
  Kim]{shenoy2026gradient}
Rahul Shenoy, Zhihong Pan, Kaushik Balakrishnan, Qisen Cheng, Yongmoon Jeon,
  Heejune Yang, and Jaewon Kim.
\newblock Gradient-free classifier guidance for diffusion model sampling.
\newblock In \emph{Proceedings of the {IEEE}/{CVF} Winter Conference on
  Applications of Computer Vision}, pages 3162--3171, 2026.

\bibitem[Na et~al.(2024)Na, Kim, Bae, Lee, Kwon, Kang, and Moon]{na2024label}
Byeonghu Na, Yeongmin Kim, HeeSun Bae, Jung~Hyun Lee, Se~Jung Kwon, Wanmo Kang,
  and Il-Chul Moon.
\newblock Label-noise robust diffusion models.
\newblock In \emph{The Twelfth International Conference on Learning
  Representations}, 2024.

\bibitem[Zhu et~al.(2023)Zhu, Li, Wang, He, and Yao]{zhu2023conditional}
Yuanzhi Zhu, Zhaohai Li, Tianwei Wang, Mengchao He, and Cong Yao.
\newblock Conditional text image generation with diffusion models.
\newblock In \emph{Proceedings of the {IEEE}/{CVF} Conference on Computer
  Vision and Pattern Recognition}, pages 14235--14244, 2023.

\bibitem[Zhan et~al.(2025)Zhan, Chen, Mei, Zhao, Chen, Chen, Lyu, and
  Wang]{zhanconditional}
Zheyuan Zhan, Defang Chen, Jian-Ping Mei, Zhenghe Zhao, Jiawei Chen, Chun Chen,
  Siwei Lyu, and Can Wang.
\newblock Conditional image synthesis with diffusion models: {A} survey.
\newblock \emph{Transactions on Machine Learning Research}, 2025.

\bibitem[Noguchi et~al.(2024)Noguchi, Fukuda, and Yamanaka]{noguchi2024scene}
Chihiro Noguchi, Shun Fukuda, and Masao Yamanaka.
\newblock Scene text image super-resolution based on text-conditional diffusion
  models.
\newblock In \emph{Proceedings of the {IEEE/CVF} Winter Conference on
  Applications of Computer Vision}, pages 1485--1495, 2024.

\bibitem[Mahajan et~al.(2018)Mahajan, Girshick, Ramanathan, He, Paluri, Li,
  Bharambe, and van~der Maaten]{mahajan2018exploring}
Dhruv Mahajan, Ross Girshick, Vignesh Ramanathan, Kaiming He, Manohar Paluri,
  Yixuan Li, Ashwin Bharambe, and Laurens van~der Maaten.
\newblock Exploring the limits of weakly supervised pretraining.
\newblock In \emph{Proceedings of the European Conference on Computer Vision},
  pages 181--196, 2018.

\bibitem[Kazerouni et~al.(2023)Kazerouni, Aghdam, Heidari, Azad, Fayyaz,
  Hacihaliloglu, and Merhof]{kazerouni2023diffusion}
Amirhossein Kazerouni, Ehsan~Khodapanah Aghdam, Moein Heidari, Reza Azad,
  Mohsen Fayyaz, Ilker Hacihaliloglu, and Dorit Merhof.
\newblock Diffusion models in medical imaging: {A} comprehensive survey.
\newblock \emph{Medical Image Analysis}, 88:\penalty0 102846, 2023.

\bibitem[Yellapragada et~al.(2025)Yellapragada, Graikos, Li, Triaridis,
  Belagali, Kapse, Nandi, Madduri, Prasanna, Kurc, Gupta, Saltz, and
  Samaras]{yellapragada2025pixcell}
Srikar Yellapragada, Alexandros Graikos, Zilinghan Li, Kostas Triaridis, Varun
  Belagali, Saarthak Kapse, Tarak~Nath Nandi, Ravi~K. Madduri, Prateek
  Prasanna, Tahsin Kurc, Rajarsi~R. Gupta, Joel Saltz, and Dimitris Samaras.
\newblock {PixCell}: A generative foundation model for digital histopathology
  images.
\newblock \emph{arXiv preprint arXiv:2506.05127}, 2025.

\bibitem[Khanna et~al.(2024)Khanna, Liu, Zhou, Meng, Rombach, Burke, Lobell,
  and Ermon]{khannadiffusionsat}
Samar Khanna, Patrick Liu, Linqi Zhou, Chenlin Meng, Robin Rombach, Marshall
  Burke, David~B. Lobell, and Stefano Ermon.
\newblock {DiffusionSat}: {A} generative foundation model for satellite
  imagery.
\newblock In \emph{The Twelfth International Conference on Learning
  Representations}, 2024.

\bibitem[Tang et~al.(2024)Tang, Cao, Hou, Jiang, Liu, and Meng]{tang2024crs}
Datao Tang, Xiangyong Cao, Xingsong Hou, Zhongyuan Jiang, Junmin Liu, and Deyu
  Meng.
\newblock {CRS-Diff}: Controllable remote sensing image generation with
  diffusion model.
\newblock \emph{{IEEE} Transactions on Geoscience and Remote Sensing},
  62:\penalty0 1--14, 2024.

\bibitem[Li et~al.(2017)Li, Wang, Li, Agustsson, and Van~Gool]{li2017webvision}
Wen Li, Limin Wang, Wei Li, Eirikur Agustsson, and Luc Van~Gool.
\newblock {WebVision} database: Visual learning and understanding from web
  data.
\newblock \emph{arXiv preprint arXiv:1708.02862}, 2017.

\bibitem[Wang et~al.(2025{\natexlab{a}})Wang, Wu, Wang, Niu, Zhang, and
  Sugiyama]{wang2025realistic}
Wei Wang, Dong-Dong Wu, Jindong Wang, Gang Niu, Min-Ling Zhang, and Masashi
  Sugiyama.
\newblock Realistic evaluation of deep partial-label learning algorithms.
\newblock In \emph{The Thirteenth International Conference on Learning
  Representations}, 2025{\natexlab{a}}.

\bibitem[Wang et~al.(2025{\natexlab{b}})Wang, Mai, Ye, Lin, and
  Lin]{wang2023climage}
Hsiu-Hsuan Wang, Tan-Ha Mai, Nai-Xuan Ye, Wei-I Lin, and Hsuan-Tien Lin.
\newblock {CLImage}: Human-annotated datasets for complementary-label learning.
\newblock \emph{Transactions on Machine Learning Research}, 2025{\natexlab{b}}.

\bibitem[He et~al.(2023)He, Han, Nie, Wang, and Wang]{he2023species196}
Wei He, Kai Han, Ying Nie, Chengcheng Wang, and Yunhe Wang.
\newblock {Species196}: {A} one-million semi-supervised dataset for
  fine-grained species recognition.
\newblock In \emph{Advances in Neural Information Processing Systems}, pages
  44957--44975, 2023.

\bibitem[Sugiyama et~al.(2022)Sugiyama, Bao, Ishida, Lu, Sakai, and
  Niu]{sugiyama2022machine}
Masashi Sugiyama, Han Bao, Takashi Ishida, Nan Lu, Tomoya Sakai, and Gang Niu.
\newblock \emph{Machine Learning from Weak Supervision: {A}n Empirical Risk
  Minimization Approach}.
\newblock Adaptive Computation and Machine Learning. {MIT} Press, 2022.

\bibitem[Li et~al.(2025)Li, Luyten, and van~der Schaar]{li2024risk}
Yangming Li, Max~Ruiz Luyten, and Mihaela van~der Schaar.
\newblock Risk-sensitive diffusion: {R}obustly optimizing diffusion models with
  noisy samples.
\newblock In \emph{International Conference on Learning Representations}, 2025.

\bibitem[Dufour et~al.(2024)Dufour, Besnier, Kalogeiton, and
  Picard]{dufour2024don}
Nicolas Dufour, Victor Besnier, Vicky Kalogeiton, and David Picard.
\newblock Don't drop your samples! {C}oherence-aware training benefits
  conditional diffusion.
\newblock In \emph{Proceedings of the IEEE/CVF Conference on Computer Vision
  and Pattern Recognition}, pages 6264--6273, 2024.

\bibitem[Chen et~al.(2025)Chen, Dobbie, Wang, Liu, Wang, and
  Zhang]{chen2025robust}
Xin Chen, Gillian Dobbie, Xinyu Wang, Feng Liu, Di~Wang, and Jingfeng Zhang.
\newblock Robust learning of diffusion models with extremely noisy conditions.
\newblock \emph{arXiv preprint arXiv:2510.10149}, 2025.

\bibitem[You et~al.(2023)You, Zhong, Bao, Sun, Li, and Zhu]{you2023diffusion}
Zebin You, Yong Zhong, Fan Bao, Jiacheng Sun, Chongxuan Li, and Jun Zhu.
\newblock Diffusion models and semi-supervised learners benefit mutually with
  few labels.
\newblock In \emph{Advances in Neural Information Processing Systems}, pages
  43479--43495, 2023.

\bibitem[Cong et~al.(2025)Cong, Bao, and Hoang-Thanh]{cong2025guiding}
Dat~Nguyen Cong, Hieu~Tran Bao, and Tung Hoang-Thanh.
\newblock Guiding noisy label conditional diffusion models with score-based
  discriminator correction.
\newblock In \emph{Proceedings of the IEEE/CVF International Conference on
  Computer Vision}, pages 18531--18541, 2025.

\bibitem[Takahashi et~al.(2025)Takahashi, Iwata, Kumagai, Yamanaka, and
  Yamashita]{takahashi2025positive}
Hiroshi Takahashi, Tomoharu Iwata, Atsutoshi Kumagai, Yuuki Yamanaka, and
  Tomoya Yamashita.
\newblock Positive-unlabeled diffusion models for preventing sensitive data
  generation.
\newblock In \emph{The Thirteenth International Conference on Learning
  Representations}, 2025.

\bibitem[Yu et~al.(2026)Yu, Yang, Lin, Zhao, Wang, Gui, Fredrikson, and
  Jiang]{yu2024your}
Weichen Yu, Ziyan Yang, Shanchuan Lin, Qi~Zhao, Jianyi Wang, Liangke Gui, Matt
  Fredrikson, and Lu~Jiang.
\newblock Is your text-to-image model robust to caption noise?
\newblock In \emph{Proceedings of the IEEE/CVF Conference on Computer Vision
  and Pattern Recognition Findings}, pages 3789--3798, 2026.

\bibitem[Clark and Jaini(2023)]{clark2023text}
Kevin Clark and Priyank Jaini.
\newblock Text-to-image diffusion models are zero-shot classifiers.
\newblock In \emph{Advances in Neural Information Processing Systems},
  volume~36, pages 58921--58937, 2023.

\bibitem[Li et~al.(2023)Li, Prabhudesai, Duggal, Brown, and Pathak]{li2023your}
Alexander~C. Li, Mihir Prabhudesai, Shivam Duggal, Ellis Brown, and Deepak
  Pathak.
\newblock Your diffusion model is secretly a zero-shot classifier.
\newblock In \emph{Proceedings of the IEEE/CVF International Conference on
  Computer Vision}, pages 2206--2217, 2023.

\bibitem[Chen et~al.(2024{\natexlab{a}})Chen, Dong, Wang, Yang, Duan, Su, and
  Zhu]{chen2023robust}
Huanran Chen, Yinpeng Dong, Zhengyi Wang, Xiao Yang, Chengqi Duan, Hang Su, and
  Jun Zhu.
\newblock Robust classification via a single diffusion model.
\newblock In \emph{Proceedings of the 41st International Conference on Machine
  Learning}, volume 235, pages 6643--6665, 2024{\natexlab{a}}.

\bibitem[Chen et~al.(2024{\natexlab{b}})Chen, Dong, Shao, Hao, Yang, Su, and
  Zhu]{chen2024diffusion}
Huanran Chen, Yinpeng Dong, Shitong Shao, Zhongkai Hao, Xiao Yang, Hang Su, and
  Jun Zhu.
\newblock Diffusion models are certifiably robust classifiers.
\newblock In \emph{Advances in Neural Information Processing Systems}, pages
  50062--50097, 2024{\natexlab{b}}.

\bibitem[Wang and Chen(2025)]{wang2025noise}
Yanghao Wang and Long Chen.
\newblock Noise matters: {O}ptimizing matching noise for diffusion classifiers.
\newblock In \emph{Advances in Neural Information Processing Systems},
  volume~38, 2025.

\bibitem[Qi et~al.(2024)Qi, Liu, Zhang, Li, Xu, Xiong, and Xie]{qi2024simple}
Zipeng Qi, Buhua Liu, Shiyan Zhang, Bao Li, Zhiqiang Xu, Haoyi Xiong, and Zeke
  Xie.
\newblock A simple and efficient baseline for zero-shot generative
  classification.
\newblock \emph{arXiv preprint arXiv:2412.12594}, 2024.

\bibitem[Jeong et~al.(2025)Jeong, Uselis, Oh, and Rohrbach]{jeong2025diffusion}
Yujin Jeong, Arnas Uselis, Seong~Joon Oh, and Anna Rohrbach.
\newblock Diffusion classifiers understand compositionality, but conditions
  apply.
\newblock In \emph{Advances in Neural Information Processing Systems}, 2025.

\bibitem[Vincent(2011)]{vincent2011connection}
Pascal Vincent.
\newblock A connection between score matching and denoising autoencoders.
\newblock \emph{Neural Computation}, 23\penalty0 (7):\penalty0 1661--1674,
  2011.

\bibitem[Karras et~al.(2022)Karras, Aittala, Aila, and
  Laine]{karras2022elucidating}
Tero Karras, Miika Aittala, Timo Aila, and Samuli Laine.
\newblock Elucidating the design space of diffusion-based generative models.
\newblock In \emph{Advances in Neural Information Processing Systems}, pages
  26565--26577, 2022.

\bibitem[Han et~al.(2020)Han, Yao, Liu, Niu, Tsang, Kwok, and
  Sugiyama]{han2020survey}
Bo~Han, Quanming Yao, Tongliang Liu, Gang Niu, Ivor~W. Tsang, James~T. Kwok,
  and Masashi Sugiyama.
\newblock A survey of label-noise representation learning: {P}ast, present and
  future.
\newblock \emph{arXiv preprint arXiv:2011.04406}, 2020.

\bibitem[Tian et~al.(2023)Tian, Yu, and Fu]{tian2023partial}
Yingjie Tian, Xiaotong Yu, and Saiji Fu.
\newblock Partial label learning: {T}axonomy, analysis and outlook.
\newblock \emph{Neural Networks}, 161:\penalty0 708--734, 2023.

\bibitem[Yang et~al.(2023)Yang, Song, King, and Xu]{yang2022survey}
Xiangli Yang, Zixing Song, Irwin King, and Zenglin Xu.
\newblock A survey on deep semi-supervised learning.
\newblock \emph{{IEEE} Transactions on Knowledge and Data Engineering},
  35\penalty0 (9):\penalty0 8934--8954, 2023.

\bibitem[Liu et~al.(2023)Liu, Cheng, and Zhang]{liu2023identifiability}
Yang Liu, Hao Cheng, and Kun Zhang.
\newblock Identifiability of label noise transition matrix.
\newblock In \emph{Proceedings of the 40th International Conference on Machine
  Learning}, volume 202, pages 21475--21496, 2023.

\bibitem[Liu et~al.(2020)Liu, Niles-Weed, Razavian, and
  Fernandez-Granda]{liu2020early}
Sheng Liu, Jonathan Niles-Weed, Narges Razavian, and Carlos Fernandez-Granda.
\newblock Early-learning regularization prevents memorization of noisy labels.
\newblock In \emph{Advances in Neural Information Processing Systems},
  volume~33, pages 20331--20342, 2020.

\bibitem[Lv et~al.(2020)Lv, Xu, Feng, Niu, Geng, and
  Sugiyama]{lv2020progressive}
Jiaqi Lv, Miao Xu, Lei Feng, Gang Niu, Xin Geng, and Masashi Sugiyama.
\newblock Progressive identification of true labels for partial-label learning.
\newblock In \emph{Proceedings of the 37th International Conference on Machine
  Learning}, volume 119, pages 6500--6510, 2020.

\bibitem[Tarvainen and Valpola(2017)]{tarvainen2017mean}
Antti Tarvainen and Harri Valpola.
\newblock Mean teachers are better role models: {W}eight-averaged consistency
  targets improve semi-supervised deep learning results.
\newblock In \emph{Advances in Neural Information Processing Systems}, pages
  1195--1204, 2017.

\bibitem[Brent(2013)]{brent2013algorithms}
Richard~P Brent.
\newblock \emph{Algorithms for Minimization without Derivatives}.
\newblock Courier Corporation, 2013.

\bibitem[Xiao et~al.(2017)Xiao, Rasul, and Vollgraf]{xiao2017fashion}
Han Xiao, Kashif Rasul, and Roland Vollgraf.
\newblock {Fashion-MNIST}: A novel image dataset for benchmarking machine
  learning algorithms.
\newblock \emph{arXiv preprint arXiv:1708.07747}, 2017.

\bibitem[Krizhevsky et~al.(2009)Krizhevsky, Hinton,
  et~al.]{krizhevsky2009learning}
Alex Krizhevsky, Geoffrey Hinton, et~al.
\newblock Learning multiple layers of features from tiny images.
\newblock Technical report, 2009.

\bibitem[Deng et~al.(2009)Deng, Dong, Socher, Li, Li, and Fei-Fei]{5206848}
Jia Deng, Wei Dong, Richard Socher, Li-Jia Li, Kai Li, and Li~Fei-Fei.
\newblock {ImageNet}: A large-scale hierarchical image database.
\newblock In \emph{2009 {IEEE} Conference on Computer Vision and Pattern
  Recognition}, pages 248--255, 2009.

\bibitem[Wei et~al.(2022)Wei, Zhu, Cheng, Liu, Niu, and Liu]{wei2022learning}
Jiaheng Wei, Zhaowei Zhu, Hao Cheng, Tongliang Liu, Gang Niu, and Yang Liu.
\newblock Learning with noisy labels revisited: {A} study using real-world
  human annotations.
\newblock In \emph{International Conference on Learning Representations}, 2022.

\bibitem[Chao et~al.(2022)Chao, Sun, Cheng, Lo, Chang, Liu, Chang, Chen, and
  Lee]{chaodenoising}
Chen-Hao Chao, Wei-Fang Sun, Bo-Wun Cheng, Yi-Chen Lo, Chia-Che Chang, Yu-Lun
  Liu, Yu-Lin Chang, Chia-Ping Chen, and Chun-Yi Lee.
\newblock Denoising likelihood score matching for conditional score-based data
  generation.
\newblock In \emph{International Conference on Learning Representations}, 2022.

\bibitem[Kaneko et~al.(2019)Kaneko, Ushiku, and Harada]{kaneko2019label}
Takuhiro Kaneko, Yoshitaka Ushiku, and Tatsuya Harada.
\newblock Label-noise robust generative adversarial networks.
\newblock In \emph{Proceedings of the {IEEE/CVF} Conference on Computer Vision
  and Pattern Recognition}, pages 2467--2476, 2019.

\bibitem[Zhang et~al.(2018)Zhang, Ciss\'{e}, Dauphin, and
  Lopez-Paz]{zhang2018mixup}
Hongyi Zhang, Moustapha Ciss\'{e}, Yann~N. Dauphin, and David Lopez-Paz.
\newblock mixup: {B}eyond empirical risk minimization.
\newblock In \emph{International Conference on Learning Representations}, 2018.

\bibitem[Han et~al.(2018)Han, Yao, Yu, Niu, Xu, Hu, Tsang, and
  Sugiyama]{han2018co}
Bo~Han, Quanming Yao, Xingrui Yu, Gang Niu, Miao Xu, Weihua Hu, Ivor~W. Tsang,
  and Masashi Sugiyama.
\newblock {Co-teaching}: Robust training of deep neural networks with extremely
  noisy labels.
\newblock In \emph{Advances in Neural Information Processing Systems}, pages
  8527--8537, 2018.

\bibitem[Yi and Wu(2019)]{yi2022pencil}
Kun Yi and Jianxin Wu.
\newblock Probabilistic end-to-end noise correction for learning with noisy
  labels.
\newblock In \emph{Proceedings of the {IEEE}/CVF Conference on Computer Vision
  and Pattern Recognition}, pages 7017--7025, 2019.

\bibitem[Qiao et~al.(2023)Qiao, Xu, and Geng]{qiaodecompositional}
Congyu Qiao, Ning Xu, and Xin Geng.
\newblock Decompositional generation process for instance-dependent partial
  label learning.
\newblock In \emph{The Eleventh International Conference on Learning
  Representations}, 2023.

\bibitem[Wang et~al.(2024)Wang, Xiao, Li, Feng, Niu, Chen, and
  Zhao]{wang2022pico+}
Haobo Wang, Ruixuan Xiao, Yixuan Li, Lei Feng, Gang Niu, Gang Chen, and Junbo
  Zhao.
\newblock {PiCO}+: {C}ontrastive label disambiguation for robust partial label
  learning.
\newblock \emph{{IEEE} Transactions on Pattern Analysis and Machine
  Intelligence}, 46\penalty0 (5):\penalty0 3183--3198, 2024.

\bibitem[Wu et~al.(2022)Wu, Wang, and Zhang]{wu2022revisiting}
Dong-Dong Wu, Deng-Bao Wang, and Min-Ling Zhang.
\newblock Revisiting consistency regularization for deep partial label
  learning.
\newblock In \emph{Proceedings of the 39th International Conference on Machine
  Learning}, volume 162, pages 24212--24225, 2022.

\bibitem[Wu et~al.(2024{\natexlab{a}})Wu, Wang, and Zhang]{wu2024distilling}
Dong-Dong Wu, Deng-Bao Wang, and Min-Ling Zhang.
\newblock Distilling reliable knowledge for instance-dependent partial label
  learning.
\newblock In \emph{Proceedings of the 38th {AAAI} Conference on Artificial
  Intelligence}, volume~38, pages 15888--15896, 2024{\natexlab{a}}.

\bibitem[Xu et~al.(2021)Xu, Shang, Ye, Qian, Li, Sun, Li, and Jin]{xu2021dash}
Yi~Xu, Lei Shang, Jinxing Ye, Qi~Qian, Yu-Feng Li, Baigui Sun, Hao Li, and Rong
  Jin.
\newblock {D}ash: {S}emi-supervised learning with dynamic thresholding.
\newblock In \emph{Proceedings of the 38th International Conference on Machine
  Learning}, volume 139, pages 11525--11536, 2021.

\bibitem[Li et~al.(2021{\natexlab{a}})Li, Xiong, and Hoi]{li2021comatch}
Junnan Li, Caiming Xiong, and Steven C.~H. Hoi.
\newblock {CoMatch}: Semi-supervised learning with contrastive graph
  regularization.
\newblock In \emph{Proceedings of the IEEE/CVF International Conference on
  Computer Vision}, pages 9475--9484, 2021{\natexlab{a}}.

\bibitem[Zhang et~al.(2021)Zhang, Wang, Hou, Wu, Wang, Okumura, and
  Shinozaki]{zhang2021flexmatch}
Bowen Zhang, Yidong Wang, Wenxin Hou, Hao Wu, Jindong Wang, Manabu Okumura, and
  Takahiro Shinozaki.
\newblock {FlexMatch}: Boosting semi-supervised learning with curriculum pseudo
  labeling.
\newblock In \emph{Advances in Neural Information Processing Systems}, pages
  18408--18419, 2021.

\bibitem[Zheng et~al.(2022)Zheng, You, Huang, Wang, Qian, and
  Xu]{zheng2022simmatch}
Mingkai Zheng, Shan You, Lang Huang, Fei Wang, Chen Qian, and Chang Xu.
\newblock {SimMatch}: Semi-supervised learning with similarity matching.
\newblock In \emph{Proceedings of the {IEEE}/{CVF} Conference on Computer
  Vision and Pattern Recognition}, pages 14471--14481, 2022.

\bibitem[Chen et~al.(2023)Chen, Tao, Fan, Wang, Wang, Schiele, Xie, Raj, and
  Savvides]{chensoftmatch}
Hao Chen, Ran Tao, Yue Fan, Yidong Wang, Jindong Wang, Bernt Schiele, Xing Xie,
  Bhiksha Raj, and Marios Savvides.
\newblock {SoftMatch}: {A}ddressing the quantity-quality tradeoff in
  semi-supervised learning.
\newblock In \emph{International Conference on Learning Representations}, 2023.

\bibitem[Li et~al.(2020)Li, Socher, and Hoi]{li2020dividemix}
Junnan Li, Richard Socher, and Steven C.~H. Hoi.
\newblock {DivideMix}: Learning with noisy labels as semi-supervised learning.
\newblock In \emph{International Conference on Learning Representations}, 2020.

\bibitem[Yu et~al.(2024)Yu, Liu, and Wang]{yu2023dataset}
Ruonan Yu, Songhua Liu, and Xinchao Wang.
\newblock Dataset distillation: {A} comprehensive review.
\newblock \emph{{IEEE} Transactions on Pattern Analysis and Machine
  Intelligence}, 46\penalty0 (1):\penalty0 150--170, 2024.

\bibitem[Zhao et~al.(2021)Zhao, Mopuri, and Bilen]{zhao2021dataset}
Bo~Zhao, Konda~Reddy Mopuri, and Hakan Bilen.
\newblock Dataset condensation with gradient matching.
\newblock In \emph{International Conference on Learning Representations}, 2021.

\bibitem[Du et~al.(2023)Du, Jiang, Tan, Zhou, and Li]{du2023minimizing}
Jiawei Du, Yidi Jiang, Vincent Y.~F. Tan, Joey~Tianyi Zhou, and Haizhou Li.
\newblock Minimizing the accumulated trajectory error to improve dataset
  distillation.
\newblock In \emph{Proceedings of the IEEE/CVF Conference on Computer Vision
  and Pattern Recognition}, pages 3749--3758, 2023.

\bibitem[Zhao et~al.(2023)Zhao, Li, Qin, and Yu]{zhao2023improved}
Ganlong Zhao, Guanbin Li, Yipeng Qin, and Yizhou Yu.
\newblock Improved distribution matching for dataset condensation.
\newblock In \emph{Proceedings of the {IEEE}/{CVF} Conference on Computer
  Vision and Pattern Recognition}, pages 7856--7865, 2023.

\bibitem[He et~al.(2024)He, Xiao, Zhou, and Tsang]{he2024multisize}
Yang He, Lingao Xiao, Joey~Tianyi Zhou, and Ivor~W. Tsang.
\newblock Multisize dataset condensation.
\newblock In \emph{The Twelfth International Conference on Learning
  Representations}, 2024.

\bibitem[Lee et~al.(2022)Lee, Chun, Jung, Yun, and Yoon]{lee2022dataset}
Saehyung Lee, Sanghyuk Chun, Sangwon Jung, Sangdoo Yun, and Sungroh Yoon.
\newblock Dataset condensation with contrastive signals.
\newblock In \emph{Proceedings of the 39th International Conference on Machine
  Learning}, volume 162, pages 12352--12364, 2022.

\bibitem[Kim et~al.(2022)Kim, Kim, Oh, Yun, Song, Jeong, Ha, and
  Song]{kim2022dataset}
Jang-Hyun Kim, Jinuk Kim, Seong~Joon Oh, Sangdoo Yun, Hwanjun Song, Joonhyun
  Jeong, Jung-Woo Ha, and Hyun~Oh Song.
\newblock Dataset condensation via efficient synthetic-data parameterization.
\newblock In \emph{Proceedings of the 39th International Conference on Machine
  Learning}, volume 162, pages 11102--11118, 2022.

\bibitem[Cazenavette et~al.(2022)Cazenavette, Wang, Torralba, Efros, and
  Zhu]{MTT}
George Cazenavette, Tongzhou Wang, Antonio Torralba, Alexei~A. Efros, and
  Jun-Yan Zhu.
\newblock Dataset distillation by matching training trajectories.
\newblock In \emph{Proceedings of the IEEE/CVF Conference on Computer Vision
  and Pattern Recognition}, pages 10718--10727, 2022.

\bibitem[Paszke et~al.(2019)Paszke, Gross, Massa, Lerer, Bradbury, Chanan,
  Killeen, Lin, Gimelshein, Antiga, Desmaison, Köpf, Yang, DeVito, Raison,
  Tejani, Chilamkurthy, Steiner, Fang, Bai, and Chintala]{paszke2019pytorch}
Adam Paszke, Sam Gross, Francisco Massa, Adam Lerer, James Bradbury, Gregory
  Chanan, Trevor Killeen, Zeming Lin, Natalia Gimelshein, Luca Antiga, Alban
  Desmaison, Andreas Köpf, Edward Yang, Zachary DeVito, Martin Raison, Alykhan
  Tejani, Sasank Chilamkurthy, Benoit Steiner, Lu~Fang, Junjie Bai, and Soumith
  Chintala.
\newblock {PyTorch}: An imperative style, high-performance deep learning
  library.
\newblock In \emph{Advances in Neural Information Processing Systems},
  volume~32, pages 8024--8035, 2019.

\bibitem[Song and Ermon(2020)]{song2020improved}
Yang Song and Stefano Ermon.
\newblock Improved techniques for training score-based generative models.
\newblock In \emph{Advances in Neural Information Processing Systems},
  volume~33, pages 12438--12448, 2020.

\bibitem[Heusel et~al.(2017)Heusel, Ramsauer, Unterthiner, Nessler, and
  Hochreiter]{heusel2017gans}
Martin Heusel, Hubert Ramsauer, Thomas Unterthiner, Bernhard Nessler, and Sepp
  Hochreiter.
\newblock {GAN}s trained by a two time-scale update rule converge to a local
  {N}ash equilibrium.
\newblock In \emph{Advances in Neural Information Processing Systems}, pages
  6626--6637, 2017.

\bibitem[Salimans et~al.(2016)Salimans, Goodfellow, Zaremba, Cheung, Radford,
  and Chen]{salimans2016improved}
Tim Salimans, Ian Goodfellow, Wojciech Zaremba, Vicki Cheung, Alec Radford, and
  Xi~Chen.
\newblock Improved techniques for training {GAN}s.
\newblock In \emph{Advances in Neural Information Processing Systems}, pages
  2234--2242, 2016.

\bibitem[Naeem et~al.(2020)Naeem, Oh, Uh, Choi, and Yoo]{naeem2020reliable}
Muhammad~Ferjad Naeem, Seong~Joon Oh, Youngjung Uh, Yunjey Choi, and Jaejun
  Yoo.
\newblock Reliable fidelity and diversity metrics for generative models.
\newblock In \emph{Proceedings of the 37th International Conference on Machine
  Learning}, pages 7176--7185, 2020.

\bibitem[Miyato and Koyama(2018)]{miyato2018cgans}
Takeru Miyato and Masanori Koyama.
\newblock c{GAN}s with projection discriminator.
\newblock In \emph{International Conference on Learning Representations}, 2018.

\bibitem[Song and Ermon(2019)]{song2019generative}
Yang Song and Stefano Ermon.
\newblock Generative modeling by estimating gradients of the data distribution.
\newblock In \emph{Advances in Neural Information Processing Systems}, pages
  11895--11907, 2019.

\bibitem[Wu et~al.(2026{\natexlab{a}})Wu, Kou, Zeng, Huang, Liu, Gu, Jia,
  Jiang, Liu, and Geng]{wu2026trustworthyfederatedlabeldistribution}
Junxiang Wu, Zhiqiang Kou, Hongwei Zeng, Wenke Huang, Biao Liu, Hanlin Gu,
  Yuheng Jia, Di~Jiang, Yang Liu, and Xin Geng.
\newblock Trustworthy federated label distribution learning under annotation
  quality disparity.
\newblock In \emph{Forty-third International Conference on Machine Learning},
  2026{\natexlab{a}}.

\bibitem[Kou et~al.(2026{\natexlab{a}})Kou, Chen, Cai, Xia, Xie, Wu, Liu, Jia,
  Geng, Sugiyama, and Chua]{kou2026pu}
Zhiqiang Kou, Junyang Chen, Xin-Qiang Cai, Xiaobo Xia, Ming-Kun Xie, Dong-Dong
  Wu, Biao Liu, Yuheng Jia, Xin Geng, Masashi Sugiyama, and Tat-Seng Chua.
\newblock Positive–unlabeled reinforcement learning distillation for
  on-premise small models.
\newblock In \emph{Proceedings of the 43rd International Conference on Machine
  Learning}, volume 306, pages 60447--60479, 2026{\natexlab{a}}.

\bibitem[Fan et~al.(2026{\natexlab{a}})Fan, Li, Zhong, Ren, Jiang, Gan, Gu, and
  Huang]{fan2026evidential}
Jinfu Fan, Jiangnan Li, Xiaohui Zhong, Kangrui Ren, Zhencun Jiang, Min Gan,
  Tianhao Gu, and Linqing Huang.
\newblock Evidential deep partial label learning to quantify disambiguation
  uncertainty.
\newblock In \emph{Proceedings of the IEEE/CVF Conference on Computer Vision
  and Pattern Recognition}, pages 24770--24779, 2026{\natexlab{a}}.

\bibitem[Fan et~al.(2026{\natexlab{b}})Fan, Li, Huang, Gan, and
  Chen]{fan2026partial}
Jinfu Fan, Jiangnan Li, Linqing Huang, Min Gan, and CL~Philip Chen.
\newblock Partial label zero-shot learning with semantic mining and
  instance--label alignment.
\newblock \emph{Engineering Applications of Artificial Intelligence},
  177:\penalty0 114854, 2026{\natexlab{b}}.

\bibitem[Luo(2022)]{luo2022understanding}
Calvin Luo.
\newblock Understanding diffusion models: {A} unified perspective.
\newblock \emph{arXiv preprint arXiv:2208.11970}, 2022.

\bibitem[Efron(2011)]{Efron_2011}
Bradley Efron.
\newblock {T}weedie's formula and selection bias.
\newblock \emph{Journal of the American Statistical Association}, 106\penalty0
  (496):\penalty0 1602--1614, 2011.

\bibitem[Luo et~al.(2025)Luo, Chen, Liu, Han, Niu, Sugiyama, Tao, and
  Gong]{luo2024estimating}
Wenshui Luo, Shuo Chen, Tongliang Liu, Bo~Han, Gang Niu, Masashi Sugiyama,
  Dacheng Tao, and Chen Gong.
\newblock Estimating per-class statistics for label noise learning.
\newblock \emph{{IEEE} Transactions on Pattern Analysis and Machine
  Intelligence}, 47\penalty0 (1):\penalty0 305--322, 2025.

\bibitem[Wang et~al.(2022{\natexlab{a}})Wang, Xia, Li, Mao, Feng, Chen, and
  Zhao]{wang2022solar}
Haobo Wang, Mingxuan Xia, Yixuan Li, Yuren Mao, Lei Feng, Gang Chen, and Junbo
  Zhao.
\newblock {SoLar}: {S}inkhorn label refinery for imbalanced partial-label
  learning.
\newblock In \emph{Advances in Neural Information Processing Systems}, pages
  8104--8117, 2022{\natexlab{a}}.

\bibitem[Kou et~al.(2025{\natexlab{a}})Kou, Xie, Wang, Chen, Wang, Xie, Chen,
  Jia, Liu, and Geng]{kou2025nips}
Zhiqiang Kou, Yucheng Xie, Hailin Wang, Junyang Chen, Jingq Wang, Ming-Kun Xie,
  Shuo Chen, Yuheng Jia, Tongliang Liu, and Xin Geng.
\newblock Rankmatch: A novel approach to semi-supervised label distribution
  learning leveraging rank correlation between labels.
\newblock In \emph{Proceedings of the 39th Conference on Neural Information
  Processing Systems}, 2025{\natexlab{a}}.

\bibitem[Kou et~al.(2025{\natexlab{b}})Kou, Qin, Wang, Wang, Xie, Chen, Jia,
  Liu, Sugiyama, and Geng]{koubldl}
Zhiqiang Kou, Si~Qin, Hailin Wang, Jing Wang, Mingkun Xie, Shuo Chen, Yuheng
  Jia, Tongliang Liu, Masashi Sugiyama, and Xin Geng.
\newblock Label distribution learning with biased annotations assisted by
  multi-label learning.
\newblock In \emph{Proceedings of the Thirty-Fourth International Joint
  Conference on Artificial Intelligence}, 2025{\natexlab{b}}.

\bibitem[Kou et~al.(2024)Kou, Wang, Tang, Jia, Shi, and Geng]{koumll}
Zhiqiang Kou, Jing Wang, Jiawei Tang, Yuheng Jia, Boyu Shi, and Xin Geng.
\newblock Exploiting multi-label correlation in label distribution learning.
\newblock In \emph{Proceedings of the Thirty-Third International Joint
  Conference on Artificial Intelligence}, pages 4326--4334, 2024.

\bibitem[Wu et~al.(2024{\natexlab{b}})Wu, Fu, Wu, Xia, Zhang, Zhou, and
  Zhang]{wu2024efficient}
Dong-Dong Wu, Chilin Fu, Weichang Wu, Wenwen Xia, Xiaolu Zhang, Jun Zhou, and
  Min-Ling Zhang.
\newblock Efficient model stealing defense with noise transition matrix.
\newblock In \emph{2024 IEEE/CVF Conference on Computer Vision and Pattern
  Recognition (CVPR)}, pages 24305--24315. IEEE, 2024{\natexlab{b}}.

\bibitem[Wu et~al.(2026{\natexlab{b}})Wu, Li, Li, and Shen]{wu2026making}
Dong-Dong Wu, Zhaoyi Li, Xiang Li, and Zhiqiang Shen.
\newblock Making partial-label datasets easier: A simple yet highly effective
  data augmentation for deep partial-label learning.
\newblock In \emph{European Conference on Computer Vision}, pages 506--524.
  Springer, 2026{\natexlab{b}}.

\bibitem[Shi et~al.(2023)Shi, Wu, Geng, and Zhang]{shi2023robust}
Yu~Shi, Dong-Dong Wu, Xin Geng, and Min-Ling Zhang.
\newblock Robust representation learning for unreliable partial label learning.
\newblock \emph{arXiv preprint arXiv:2308.16718}, 2023.

\bibitem[Yao et~al.(2020)Yao, Liu, Han, Gong, Deng, Niu, and
  Sugiyama]{yao2020dual}
Yu~Yao, Tongliang Liu, Bo~Han, Mingming Gong, Jiankang Deng, Gang Niu, and
  Masashi Sugiyama.
\newblock Dual {T}: {R}educing estimation error for transition matrix in
  label-noise learning.
\newblock In \emph{Advances in Neural Information Processing Systems},
  volume~33, pages 7260--7271, 2020.

\bibitem[Kou et~al.(2026{\natexlab{b}})Kou, Wu, Huang, He, Xie, Wang, Jia,
  Jiang, Liu, Geng, and Yang]{kou2026fedharmony}
Zhiqiang Kou, Junxiang Wu, Wenke Huang, Wenwen He, Ming-Kun Xie, Changwei Wang,
  Yuheng Jia, Di~Jiang, Yang Liu, Xin Geng, and Qiang Yang.
\newblock Fedharmony: Harmonizing heterogeneous label correlations in federated
  multi-label learning.
\newblock In \emph{Proceedings of the IEEE/CVF Conference on Computer Vision
  and Pattern Recognition}, 2026{\natexlab{b}}.

\bibitem[Kou et~al.(2025{\natexlab{c}})Kou, Wang, Jia, Liu, and Geng]{koutnnls}
Zhiqiang Kou, Jing Wang, Yuheng Jia, Biao Liu, and Xin Geng.
\newblock Instance-dependent inaccurate label distribution learning.
\newblock \emph{IEEE Transactions on Neural Networks and Learning Systems},
  36\penalty0 (1):\penalty0 1425--1437, 2025{\natexlab{c}}.

\bibitem[Fan et~al.(2026{\natexlab{c}})Fan, Zhong, Ren, Li, Huang, Gan, and
  Chen]{fan2026partial2}
JinFu Fan, Xiaohui Zhong, Kangrui Ren, JiangNan Li, Linqing Huang, Min Gan, and
  CL~Philip Chen.
\newblock Diffusion disambiguation models for partial label learning.
\newblock \emph{IEEE Transactions on Circuits and Systems for Video
  Technology}, 36\penalty0 (10):\penalty0 14621--14636, 2026{\natexlab{c}}.

\bibitem[Fan et~al.(2025)Fan, Huang, Gong, You, Gan, and Wang]{fan2024kmt}
Jinfu Fan, Linqing Huang, Chaoyu Gong, Yang You, Min Gan, and Zhongjie Wang.
\newblock Kmt-pll: K-means cross-attention transformer for partial label
  learning.
\newblock \emph{IEEE Transactions on Neural Networks and Learning Systems},
  36\penalty0 (2):\penalty0 2789--2800, 2025.

\bibitem[Li et~al.(2021{\natexlab{b}})Li, Liu, Han, Niu, and
  Sugiyama]{li2021provably}
Xuefeng Li, Tongliang Liu, Bo~Han, Gang Niu, and Masashi Sugiyama.
\newblock Provably end-to-end label-noise learning without anchor points.
\newblock In \emph{Proceedings of the 38th International Conference on Machine
  Learning}, volume 139, pages 6403--6413, 2021{\natexlab{b}}.

\bibitem[Yu et~al.(2019)Yu, Han, Yao, Niu, Tsang, and Sugiyama]{yu2019does}
Xingrui Yu, Bo~Han, Jiangchao Yao, Gang Niu, Ivor~W. Tsang, and Masashi
  Sugiyama.
\newblock How does disagreement help generalization against label corruption?
\newblock In \emph{Proceedings of the 36th International Conference on Machine
  Learning}, volume~97, pages 7164--7173, 2019.

\bibitem[Wang et~al.(2019)Wang, Ma, Chen, Luo, Yi, and
  Bailey]{wang2019symmetric}
Yisen Wang, Xingjun Ma, Zaiyi Chen, Yuan Luo, Jinfeng Yi, and James Bailey.
\newblock Symmetric cross entropy for robust learning with noisy labels.
\newblock In \emph{Proceedings of the IEEE/CVF International Conference on
  Computer Vision}, pages 322--330, 2019.

\bibitem[Zhang and Sabuncu(2018)]{zhang2018generalized}
Zhilu Zhang and Mert~R. Sabuncu.
\newblock Generalized cross entropy loss for training deep neural networks with
  noisy labels.
\newblock In \emph{Advances in Neural Information Processing Systems}, pages
  8792--8802, 2018.

\bibitem[Malach and Shalev-Shwartz(2017)]{malach2017decoupling}
Eran Malach and Shai Shalev-Shwartz.
\newblock Decoupling ``when to update'' from ``how to update''.
\newblock In \emph{Advances in Neural Information Processing Systems}, pages
  960--970, 2017.

\bibitem[Wei et~al.(2020)Wei, Feng, Chen, and An]{wei2020combating}
Hongxin Wei, Lei Feng, Xiangyu Chen, and Bo~An.
\newblock Combating noisy labels by agreement: {A} joint training method with
  co-regularization.
\newblock In \emph{Proceedings of the IEEE/CVF Conference on Computer Vision
  and Pattern Recognition}, pages 13726--13735, 2020.

\bibitem[Zhang et~al.(2022)Zhang, Feng, Han, Liu, Niu, Qin, and
  Sugiyama]{zhang2021exploiting}
Fei Zhang, Lei Feng, Bo~Han, Tongliang Liu, Gang Niu, Tao Qin, and Masashi
  Sugiyama.
\newblock Exploiting class activation value for partial-label learning.
\newblock In \emph{International Conference on Learning Representations}, 2022.

\bibitem[Xu et~al.(2023)Xu, Liu, Lv, Qiao, and Geng]{xu2023progressive}
Ning Xu, Biao Liu, Jiaqi Lv, Congyu Qiao, and Xin Geng.
\newblock Progressive purification for instance-dependent partial label
  learning.
\newblock In \emph{Proceedings of the 40th International Conference on Machine
  Learning}, volume 202, pages 38551--38565, 2023.

\bibitem[Feng et~al.(2020)Feng, Lv, Han, Xu, Niu, Geng, An, and
  Sugiyama]{feng2020provably}
Lei Feng, Jiaqi Lv, Bo~Han, Miao Xu, Gang Niu, Xin Geng, Bo~An, and Masashi
  Sugiyama.
\newblock Provably consistent partial-label learning.
\newblock In \emph{Advances in Neural Information Processing Systems},
  volume~33, pages 10948--10960, 2020.

\bibitem[Xia et~al.(2022)Xia, Lv, Xu, and Geng]{xia2022ambiguity}
Shi-Yu Xia, Jiaqi Lv, Ning Xu, and Xin Geng.
\newblock Ambiguity-induced contrastive learning for instance-dependent partial
  label learning.
\newblock In \emph{Proceedings of the Thirty-First International Joint
  Conference on Artificial Intelligence}, pages 3615--3621, 2022.

\bibitem[Wang et~al.(2022{\natexlab{b}})Wang, Chen, Fan, Sun, Tao, Hou, Wang,
  Yang, Zhou, Guo, Qi, Wu, Li, Nakamura, Ye, Savvides, Raj, Shinozaki, Schiele,
  Wang, Xie, and Zhang]{wang2022usb}
Yidong Wang, Hao Chen, Yue Fan, Wang Sun, Ran Tao, Wenxin Hou, Renjie Wang,
  Linyi Yang, Zhi Zhou, Lan-Zhe Guo, Heli Qi, Zhen Wu, Yu-Feng Li, Satoshi
  Nakamura, Wei Ye, Marios Savvides, Bhiksha Raj, Takahiro Shinozaki, Bernt
  Schiele, Jindong Wang, Xing Xie, and Yue Zhang.
\newblock {USB}: {A} unified semi-supervised learning benchmark for
  classification.
\newblock In \emph{Advances in Neural Information Processing Systems}, pages
  3938--3961, 2022{\natexlab{b}}.

\bibitem[Lee(2013)]{lee2013pseudo}
Dong-Hyun Lee.
\newblock Pseudo-label: {T}he simple and efficient semi-supervised learning
  method for deep neural networks.
\newblock In \emph{Workshop on Challenges in Representation Learning,
  International Conference on Machine Learning}, pages 1--6, 2013.

\bibitem[Miyato et~al.(2019)Miyato, Maeda, Koyama, and
  Ishii]{miyato2018virtual}
Takeru Miyato, Shin-ichi Maeda, Masanori Koyama, and Shin Ishii.
\newblock Virtual adversarial training: {A} regularization method for
  supervised and semi-supervised learning.
\newblock \emph{{IEEE} Transactions on Pattern Analysis and Machine
  Intelligence}, 41\penalty0 (8):\penalty0 1979--1993, 2019.

\bibitem[Xie et~al.(2020)Xie, Dai, Hovy, Luong, and Le]{xie2020unsupervised}
Qizhe Xie, Zihang Dai, Eduard Hovy, Minh-Thang Luong, and Quoc~V. Le.
\newblock Unsupervised data augmentation for consistency training.
\newblock In \emph{Advances in Neural Information Processing Systems},
  volume~33, pages 6256--6268, 2020.

\bibitem[Sohn et~al.(2020)Sohn, Berthelot, Carlini, Zhang, Zhang, Raffel,
  Cubuk, Kurakin, and Li]{sohn2020fixmatch}
Kihyuk Sohn, David Berthelot, Nicholas Carlini, Zizhao Zhang, Han Zhang,
  Colin~A. Raffel, Ekin~D. Cubuk, Alexey Kurakin, and Chun-Liang Li.
\newblock {F}ix{M}atch: {S}implifying semi-supervised learning with consistency
  and confidence.
\newblock In \emph{Advances in Neural Information Processing Systems},
  volume~33, pages 596--608, 2020.

\bibitem[Wang et~al.(2023)Wang, Chen, Heng, Hou, Fan, Wu, Wang, Savvides,
  Shinozaki, Raj, Schiele, and Xie]{wang2022freematch}
Yidong Wang, Hao Chen, Qiang Heng, Wenxin Hou, Yue Fan, Zhen Wu, Jindong Wang,
  Marios Savvides, Takahiro Shinozaki, Bhiksha Raj, Bernt Schiele, and Xing
  Xie.
\newblock {F}ree{M}atch: Self-adaptive thresholding for semi-supervised
  learning.
\newblock In \emph{Eleventh International Conference on Learning
  Representations}, 2023.

\bibitem[Tschandl et~al.(2018)Tschandl, Rosendahl, and
  Kittler]{tschandl2018ham10000}
Philipp Tschandl, Cliff Rosendahl, and Harald Kittler.
\newblock The {HAM10000} dataset, a large collection of multi-source
  dermatoscopic images of common pigmented skin lesions.
\newblock \emph{Scientific Data}, 5\penalty0 (1):\penalty0 180161, 2018.

\bibitem[Kim and Song(2025)]{kim2025robust}
Nicole Hee-Yeon Kim and Hwanjun Song.
\newblock Robust dataset condensation using supervised contrastive learning.
\newblock In \emph{Proceedings of the {IEEE}/{CVF} International Conference on
  Computer Vision}, pages 2857--2866, 2025.

\bibitem[Zhao and Bilen(2023)]{zhao2023dataset}
Bo~Zhao and Hakan Bilen.
\newblock Dataset condensation with distribution matching.
\newblock In \emph{Proceedings of the IEEE/CVF Winter Conference on
  Applications of Computer Vision}, pages 6514--6523, 2023.

\bibitem[Zhao and Bilen(2021)]{zhao2021datasetDSA}
Bo~Zhao and Hakan Bilen.
\newblock Dataset condensation with differentiable {S}iamese augmentation.
\newblock In \emph{Proceedings of the 38th International Conference on Machine
  Learning}, volume 139, pages 12674--12685, 2021.

\bibitem[Sun et~al.(2024)Sun, Shi, Yu, and Lin]{sun2024diversity}
Peng Sun, Bei Shi, Daiwei Yu, and Tao Lin.
\newblock On the diversity and realism of distilled dataset: {A}n efficient
  dataset distillation paradigm.
\newblock In \emph{Proceedings of the {IEEE}/{CVF} Conference on Computer
  Vision and Pattern Recognition}, pages 9390--9399, 2024.

\bibitem[Yin et~al.(2023)Yin, Xing, and Shen]{yin2024squeeze}
Zeyuan Yin, Eric Xing, and Zhiqiang Shen.
\newblock Squeeze, recover and relabel: {D}ataset condensation at {ImageNet}
  scale from a new perspective.
\newblock In \emph{Advances in Neural Information Processing Systems},
  volume~36, pages 73582--73603, 2023.

\end{thebibliography}
}

\clearpage
\appendix
\etocdepthtag.toc{mtappendix}
\etocsettagdepth{mtmain}{none}
\etocsettagdepth{mtappendix}{subsection}
\renewcommand{\contentsname}{Appendix}
\tableofcontents

\clearpage
\section{Notation and Definitions}
\label{sec:symbols}
Table~\ref{tab:symbols} summarizes the main notation used throughout the paper.
\begin{table}[!h]
\centering
\small
\renewcommand{\arraystretch}{1.25}
\caption{List of common mathematical symbols used in this paper.}
\label{tab:symbols}
\begin{tabular}{p{3cm}<{\centering}p{9.5cm}}
\toprule
\textbf{Symbol} & \textbf{Definition} \\
\midrule
$\mathbf{x}$ & Data sample \\
$X$ & Random variable of data samples \\
$y$ & Latent clean class label \\
$Y$ & Random variable of clean class labels \\
$z$ & Observed weak annotation \\
$Z$ & Random variable of weak annotations \\
$\hat{y}$ & Observed noisy label \\
$\hat{Y}$ & Random variable of noisy labels \\
$\emptyset$ & Missing-label symbol for unlabeled samples \\
$\mathcal{X}$ & Input space \\
$\mathcal{Y}$ & Class-label space \\
$c$ & Number of classes \\
$\mathcal{D}$ & Weakly annotated training dataset \\
$\theta$ & Current trainable LCDM parameters \\
$\phi$ & EMA copy of the LCDM parameters \\
$q(\mathbf{x})$ & Real data density \\
$q_t(\mathbf{x}_t\mid y)$ & Perturbed clean class-conditional density \\
$q(\mathbf{x}_t \mid \mathbf{x})$ & Forward Gaussian perturbation distribution \\
$p_\theta(\cdot)$ & Current class-posterior estimate \\
$p_\phi(\cdot)$ & EMA class-posterior estimate \\
$\mathbf{x}_t$ & Noisy latent state at timestep $t$ \\
$t$ & Diffusion timestep \\
$T$ & Total number of diffusion timesteps \\
$\alpha_t$ & Signal scale in the forward process \\
$\sigma_t$ & Noise scale in the forward process \\
$w_t$ & Timestep-dependent weight in score matching \\
$\mathbf{s}_\theta(\mathbf{x}_t,y,t)$ & Class-conditional score estimator \\
$\mathbf{h}_\theta(\mathbf{x}_\tau,y,\tau)$ & Class-conditional denoised estimator \\
$f_\theta$ & Current diffusion-classifier posterior estimator \\
$f_\phi$ & EMA diffusion-classifier posterior estimator \\
$\mathcal{L}_\text{Gen}(\theta)$ & Generative loss for posterior-weighted score matching \\
$\mathcal{L}_\text{Cls}(\theta)$ & Classification loss for posterior refinement \\

\bottomrule
\end{tabular}
\end{table}

\clearpage
\section{Implementation and Evaluation Protocols}
\label{sec:implementation}
This section details the experimental protocols used in the main paper, including weak-annotation construction, model training, and evaluation metrics.
Our implementation is based on PyTorch~1.12~\citep{paszke2019pytorch}, and all experiments were conducted on NVIDIA Tesla A100 GPUs with CUDA~12.4. 

\subsection{Weak-Annotation Construction}
\label{exp:weak_construction}
For all class-dependent ambiguous-label datasets, we construct a $10 \times 10$ circulant transition matrix $T$ where each row maps a true label to a candidate set of labels with varying probabilities. We set $q=0.5$ and define the matrix as:
\begingroup
\small
\[
T = 
\setlength{\arraycolsep}{3pt}
\begin{bmatrix}
1 & q\!+\!0.2 & q & \cdots & q\!+\!0.2 & q & q\!-\!0.2 \\
q\!-\!0.2 & 1 & q\!+\!0.2 & \cdots & q & q\!-\!0.2 & q\!+\!0.2 \\
q & q\!-\!0.2 & 1 & \cdots & q\!-\!0.2 & q\!+\!0.2 & q \\
\vdots & \vdots & \vdots & \ddots & \vdots & \vdots & \vdots \\
q\!+\!0.2 & q & q\!-\!0.2  & \cdots & q & q\!-\!0.2 & 1
\end{bmatrix}.
\]
\endgroup
For noisy-label datasets with asymmetric noise (40\% flip probability), we adopt the following mappings:
\textit{Fashion-MNIST:} `Pullover'$\to$`Sneaker', `Dress'$\to$`Bag', `Sandal'$\to$`Shirt', `Shirt'$\to$`Sandal'.  
\textit{CIFAR-10:} `Truck'$\to$`Automobile', `Bird'$\to$`Airplane', `Deer'$\to$`Horse', `Cat'$\to$`Dog', `Dog'$\to$`Cat'.  
\textit{ImageNette:} `Tench'$\to$`English springer', `Cassette player'$\to$`Garbage truck', `Chain saw'$\to$`Church', `Golf ball'$\to$`Parachute', `Parachute'$\to$`Golf ball'.

\subsection{Model and Training Setup}
\label{exp:model_setup}
We build our framework upon the \textsc{Edm} formulation~\citep{karras2022elucidating}, utilizing the DDPM++ network architecture with a U-Net backbone~\citep{song2020improved} for all experiments.
Optimization is performed using the Adam optimizer with a learning rate of $1 \times 10^{-3}$, $(\beta_1, \beta_2) = (0.9, 0.999)$, and $\epsilon = 1 \times 10^{-8}$.
We apply an Exponential Moving Average (EMA) decay of $0.5$.
The batch sizes are set to 128 for Fashion-MNIST, 64 for CIFAR-10, and 16 for ImageNette.
Regarding the proposed diffusion classifier, the timestep interval length $\Delta$ is set to $6.4$.
All models are trained from scratch for a total of $200\text{k}$ iterations.

\subsection{Evaluation Metrics}
\label{exp:metrics}
We evaluate the trained CDMs using four unconditional metrics, Fr\'echet Inception Distance (FID)~\citep{heusel2017gans}, Inception Score (IS)~\citep{salimans2016improved}, Density, and Coverage~\citep{naeem2020reliable}, together with three class-wise conditional metrics, CW-FID, CW-Density, and CW-Coverage~\citep{chaodenoising}.
All metrics are computed using the official implementation of DLSM~\citep{chaodenoising}.

\noindent\textbf{Unconditional metrics.}
Unconditional metrics ignore class labels when comparing real and generated distributions.
In our experiments, samples are generated conditionally per class and then pooled before evaluation, following prior conditional-generation protocols~\citep{kaneko2019label,chaodenoising}.
FID measures feature-space distribution distance, IS reflects realism and class separability, and Density and Coverage are robust analogs of Precision and Recall.

\noindent\textbf{Conditional metrics.}
To measure class-conditional consistency, we use class-wise variants of the above metrics.
Each metric is computed within each class and then averaged across classes.
CW-FID, also known as intra-FID, is a standard measure of conditional distribution quality~\citep{miyato2018cgans,kaneko2019label}.

\noindent\textbf{Remark.}
Fashion-MNIST is not suitable for these image-generation metrics, so we do not report FID, IS, Density, Coverage, or their class-wise variants on Fashion-MNIST.

\clearpage
\section{Theoretical Derivations and Proofs}
\label{sec:proof}

\noindent
This section collects the derivations and proofs supporting the main method.
We first relate our diffusion notation to common formulations, then prove the score decomposition, the posterior-weighted score-matching property, the timestep-sampling rule, and the variational objectives used in training.

\subsection{Connections among Different Diffusion Models.}
\label{proof:convert}
The diffusion model used in this paper admits a flexible parameterization that aligns with common diffusion frameworks, including DDPM~\citep{ho2020denoising}, \textsc{Smld}~\citep{song2019generative}, \textsc{Ve-sde}~\citep{songscore}, and \textsc{Vp-sde}~\citep{songscore}.
This compatibility allows us to analyze these frameworks through specific choices of the schedules $\alpha_t$ and $\sigma_t$.

\noindent\textbf{DDPM.}
DDPM defines a sequence $\{\beta_t\}_{t=0}^{T}$ and 
$\mathbf{x}_t = \sqrt{\prod_{i=0}^{t}(1-\beta_i)}\mathbf{x}_0 + \sqrt{1-\prod_{i=0}^{t}(1-\beta_i)}\boldsymbol{\epsilon}$,
which can be seen as a special case of Eq.~(\ref{eq:forward}) where we can set $\alpha_t=\sqrt{\prod_{i=0}^{t}(1-\beta_i)}$ and $\sigma_t=\sqrt{1-\prod_{i=0}^{t}(1-\beta_i)}$.

\noindent\textbf{\textsc{Smld}.}
\textsc{Smld} defines a noise schedule $\sigma(t)_{t=0}^T$ and $\mathbf{x}_t=\mathbf{x}_0+\sigma(t)\boldsymbol{\epsilon}$, with $\sigma(1)<\sigma(2)<\cdots<\sigma(T)$.
In this scenario, Eq.~(\ref{eq:forward}) reduces to $\alpha_t=1$, $\sigma_t=\sigma(t)$.

\noindent\textbf{\textsc{Vp-sde}.} 
\textsc{Vp-sde} is the continuous-time counterpart of DDPM and defines the stochastic differential equation
\begin{equation*}
dX_t = -\frac{1}{2}\beta(t)X_t \,\mathrm{d}t + \sqrt{\beta(t)} \,\mathrm{d}W_t,\quad t \in [0,1],
\end{equation*}
where $\beta(t)=\beta_{t\cdot T}\cdot T$. 
In this setup, $\alpha_t=\sqrt{\exp\left(-\int_0^t\beta(s)\,\mathrm{d}s\right)}$, and $\sigma_t=\sqrt{1-\exp\left(-\int_0^t\beta(s)\,\mathrm{d}s\right)}$. %

\noindent\textbf{\textsc{Ve-sde}.} \textsc{Ve-sde} is the continuous-time counterpart of \textsc{Smld}, with forward process
\begin{equation*}
dX_t=\sqrt{\frac{\mathrm{d}\sigma(t)^2}{\mathrm{d}t}}\,\mathrm{d}W_t.
\end{equation*}
In this setup, $\alpha_t=1$ and $\sigma_t=\sqrt{\sigma^2(t)-\sigma^2(0)}$.

\noindent\textbf{Weighting function $w_t$.}
Different frameworks typically adopt specific prediction targets (e.g., $\epsilon$-prediction in DDPM or score-prediction in SMLD), which implicitly corresponds to the usage of specific weighting functions $w_t$ in the score matching objective. 
For instance, when the model is trained to predict $\boldsymbol{\epsilon}$, the resulting objective is equivalent to a score-matching loss with weight $w_t \propto \sigma_t^2$; training to predict $\mathbf{x}_0$ instead induces a different rescaling, and hence a different effective weighting.
While the models above each define a specific framework, EDM~\citep{karras2022elucidating} provides a unified structure and optimizes these choices, including network input and output preconditioning, for stable training.
We therefore adopt EDM as our default implementation framework.
We provide the conversion logic in Algorithm \ref{alg:linear_to_edm} to demonstrate how our score network $\mathbf{s}_\theta$ adapts to different prediction targets ($\mathbf{x}_0$, $\boldsymbol{\epsilon}$, or $\boldsymbol{v}$), thereby effectively recovering the distinct loss weightings used in prior works.

\begin{algorithm}[!b]
\caption{Formulation Conversion (Our Model to EDM)}
\label{alg:linear_to_edm}
\begin{algorithmic}[1]
\REQUIRE A score network $\mathbf{s}_\theta$, a noisy input $\mathbf{x}_t$, timestep $t$, and forward-process schedules $\{\alpha_i\}_{i=1}^{T}$ and $\{\sigma_i\}_{i=1}^{T}$.
\STATE Estimate the clean sample using $\mathbf{s}_\theta$:
$\hat{\mathbf{x}}_0=(\mathbf{x}_t+\sigma_t^2 \mathbf{s}_\theta(\mathbf{x}_t,t))/\alpha_t$
\IF{performing $\mathbf{x}_0$-prediction}
    \STATE Return $\hat{\mathbf{x}}_0$.
\ENDIF
\STATE Estimate the noise component:
$\hat{\boldsymbol{\epsilon}}=\frac{\mathbf{x}_t-\alpha_t\hat{\mathbf{x}}_0}{\sigma_t}$
\IF{performing $\boldsymbol{\epsilon}$-prediction}
    \STATE Return $\hat{\boldsymbol{\epsilon}}$.
\ENDIF
\STATE Estimate the velocity component:
$\hat{\boldsymbol{v}}=\alpha_t\hat{\boldsymbol{\epsilon}}-\sigma_t\hat{\mathbf{x}}_0$
\IF{performing $\boldsymbol{v}$-prediction}
    \STATE Return $\hat{\boldsymbol{v}}$.
\ENDIF
\end{algorithmic}
\end{algorithm}

\subsection{Derivation of Eq.~(\ref{eq:VLB_LL})}
\label{proof:VLB}
To maximize the log-likelihood $\log p_{\theta}(X,Z)$, we introduce an approximate posterior distribution $p_{\phi}(Y \!\mid\! X,Z)$ over the latent variable $Y$.
Applying Jensen's inequality, we derive the ELBO as follows:
\begin{align*}
\log p_{\theta}(X,Z) &= \log \sum_{Y\in\mathcal{Y}} p_{\theta}(X,Y,Z) \\
&= \log \sum_{Y\in\mathcal{Y}} p_{\phi}(Y \!\mid\! X,Z) \frac{p_{\theta}(X,Y,Z)}{p_{\phi}(Y \!\mid\! X,Z)} \\
&\geq \sum_{Y\in\mathcal{Y}} p_{\phi}(Y \!\mid\! X,Z) \log \frac{p_{\theta}(X,Y,Z)}{p_{\phi}(Y \!\mid\! X,Z)} \\
&= \mathbb{E}_{p_{\phi}(Y \mid X,Z)} \Big[ \log p_{\theta}(X,Y,Z) - \log p_{\phi}(Y \!\mid\! X,Z) \Big].
\end{align*}
Since the entropy term $-\mathbb{E}_{p_{\phi}}[\log p_{\phi}]$ does not depend on $\theta$, maximizing the ELBO w.r.t. $\theta$ simplifies to

\begin{equation*}
\arg\max_{\theta}\ \mathbb{E}_{p_{\phi}(Y \mid X,Z)} \big[ \log p_{\theta}(X,Y,Z) \big].
\end{equation*}

Using the chain rule of probability, we can decompose the log-likelihood as
\begin{align*}
\log p_{\theta}(X,Y,Z) &= \log p_{\theta}(X \!\mid\! Z) + \log p_{\theta}(Y \!\mid\! X,Z) + \log p_{\theta}(Z).
\end{align*}
Substituting this back into the expectation:
\begin{align*}
{\theta^*} &= \arg\max_{\theta}\ \mathbb{E}_{p_{\phi}(Y \mid X,Z)} \Big[ \log p_{\theta}(X \!\mid\! Z) + \log p_{\theta}(Y \!\mid\! X,Z) + \log p_{\theta}(Z) \Big] \\
&= \arg\max_{\theta}\ \Big\{ \mathbb{E}_{p_{\phi}(Y \mid X,Z)} \big[ \log p_{\theta}(X \!\mid\! Z) \big] + \mathbb{E}_{p_{\phi}(Y \mid X,Z)} \big[ \log p_{\theta}(Y \!\mid\! X,Z) \big] \Big\}.
\end{align*}

The term $\log p_{\theta}(Z)$ is constant w.r.t. the optimization and thus omitted, yielding the objective in Eq.~(\ref{eq:VLB_LL}).

\subsection{Proof of Theorem \ref{theorem:decouple}}
\label{proof:decouple}
The derivation is analogous to Theorem 1 in \citet{na2024label}, but we rewrite it here using the notation and assumptions of the current paper for completeness.
Under the class-conditional independence assumption $X \perp Z \mid Y$, the perturbed variable $\mathbf{x}_t$ also satisfies $\mathbf{x}_t \perp Z \mid Y$, because $\mathbf{x}_t$ is generated from $X$ through the forward diffusion process and therefore does not depend on $Z$ once $Y$ is fixed. Hence
\[
q_t(\mathbf{x}_t \mid y,z)=q_t(\mathbf{x}_t \mid y).
\]
Marginalizing over the latent clean label $Y$ gives
\[
q_t(\mathbf{x}_t \mid z)
= \sum_{y=1}^{c} q_t(\mathbf{x}_t,y \mid z)
= \sum_{y=1}^{c} q_t(\mathbf{x}_t \mid y,z)\,p(y \mid z)
= \sum_{y=1}^{c} q_t(\mathbf{x}_t \mid y)\,p(y \mid z).
\]
Taking the gradient of the log-density, we obtain
\[
\nabla_{\mathbf{x}_t}\log q_t(\mathbf{x}_t \mid z)
= \frac{\sum_{y=1}^{c} p(y \mid z)\,\nabla_{\mathbf{x}_t} q_t(\mathbf{x}_t \mid y)}{q_t(\mathbf{x}_t \mid z)}.
\]
Multiplying and dividing each term by $q_t(\mathbf{x}_t \mid y)$ yields
\[
\nabla_{\mathbf{x}_t}\log q_t(\mathbf{x}_t \mid z)
= \sum_{y=1}^{c}
\frac{p(y \mid z)\,q_t(\mathbf{x}_t \mid y)}{q_t(\mathbf{x}_t \mid z)}
\nabla_{\mathbf{x}_t}\log q_t(\mathbf{x}_t \mid y).
\]
Now apply Bayes' rule:
\[
p(y \mid \mathbf{x}_t,z)
= \frac{q_t(\mathbf{x}_t \mid y,z)\,p(y \mid z)}{q_t(\mathbf{x}_t \mid z)}
= \frac{q_t(\mathbf{x}_t \mid y)\,p(y \mid z)}{q_t(\mathbf{x}_t \mid z)},
\]
where the second equality again uses $\mathbf{x}_t \perp Z \mid Y$. Therefore,
\[
\frac{p(y \mid z)\,q_t(\mathbf{x}_t \mid y)}{q_t(\mathbf{x}_t \mid z)}
= p(y \mid \mathbf{x}_t,z).
\]
Substituting this identity into the previous display gives Eq.~(\ref{eq:convex_combination}):
\[
\nabla_{\mathbf{x}_t}\log q_t(\mathbf{x}_t \mid z)
= \sum_{y=1}^{c} p(y \mid \mathbf{x}_t,z)\,\nabla_{\mathbf{x}_t}\log q_t(\mathbf{x}_t \mid y).
\]
This completes the proof.

\subsection{Proof of Proposition \ref{proposition:WDSM}}
\label{proof:WDSM}

Assume that the posterior estimate is exact, i.e., $p_\phi(y \mid \mathbf{x}_t,z)=p(y \mid \mathbf{x}_t,z)$. Define the posterior-weighted aggregation of the class-wise scores as
\[
g_\theta(\mathbf{x}_t,z,t)
:= \sum_{y=1}^c p(y \mid \mathbf{x}_t,z)\,\mathbf{s}_\theta(\mathbf{x}_t,y,t).
\]
With this notation, Eq.~(\ref{eq:WDSM}) is exactly a denoising score matching objective for the weak-conditional distribution:
\[
\mathcal{L}_{\mathrm{DSM}}(\theta)
:= \mathbb{E}_{t,z}\Big[w_t\,
\mathbb{E}_{\mathbf{x}_t,\mathbf{x}\sim q(\mathbf{x}_t,\mathbf{x}\mid z)}
\big\|g_\theta(\mathbf{x}_t,z,t)-a(\mathbf{x}_t,\mathbf{x},z)\big\|_2^2
\Big],
\]
where
\[
a(\mathbf{x}_t,\mathbf{x},z)
:=\nabla_{\mathbf{x}_t}\log q_{t\mid0}(\mathbf{x}_t\mid \mathbf{x},z).
\]
Let
\[
b(\mathbf{x}_t,z):=\nabla_{\mathbf{x}_t}\log q_t(\mathbf{x}_t\mid z)
\]
be the explicit weak-conditional score. By the standard DSM identity,
\[
\mathbb{E}_{\mathbf{x}\sim q(\mathbf{x}\mid \mathbf{x}_t,z)}
\big[a(\mathbf{x}_t,\mathbf{x},z)\big]
= b(\mathbf{x}_t,z).
\]
Therefore, conditioning on $(\mathbf{x}_t,z)$ gives the bias--variance decomposition
\[
\mathbb{E}_{\mathbf{x}\sim q(\mathbf{x}\mid \mathbf{x}_t,z)}
\big\|g_\theta(\mathbf{x}_t,z,t)-a(\mathbf{x}_t,\mathbf{x},z)\big\|_2^2
= \big\|g_\theta(\mathbf{x}_t,z,t)-b(\mathbf{x}_t,z)\big\|_2^2
+ C(\mathbf{x}_t,z),
\]
where $C(\mathbf{x}_t,z)$ is independent of $\theta$. Hence $\mathcal{L}_{\mathrm{DSM}}(\theta)$ differs from the explicit score matching objective
\[
\mathcal{L}_{\mathrm{ESM}}(\theta)
:= \mathbb{E}_{t,z}\Big[w_t\,
\mathbb{E}_{\mathbf{x}_t\sim q(\mathbf{x}_t\mid z)}
\big\|g_\theta(\mathbf{x}_t,z,t)-\nabla_{\mathbf{x}_t}\log q_t(\mathbf{x}_t\mid z)\big\|_2^2
\Big]
\]
only by a constant independent of $\theta$. Thus the two objectives share the same minimizers, and any optimal aggregated predictor satisfies
\[
g_{\theta^*}(\mathbf{x}_t,z,t)=\nabla_{\mathbf{x}_t}\log q_t(\mathbf{x}_t\mid z).
\]
By Theorem~\ref{theorem:decouple}, this weak-conditional score decomposes as
\[
\nabla_{\mathbf{x}_t}\log q_t(\mathbf{x}_t\mid z)
= \sum_{y=1}^c p(y \mid \mathbf{x}_t,z)\,\nabla_{\mathbf{x}_t}\log q_t(\mathbf{x}_t\mid y).
\]
Combining the last two displays with the definition of $g_{\theta^*}$ gives
\[
\sum_{y=1}^c p(y \mid \mathbf{x}_t,z)\,\mathbf{s}_{\theta^*}(\mathbf{x}_t,y,t)
= \sum_{y=1}^c p(y \mid \mathbf{x}_t,z)\,\nabla_{\mathbf{x}_t}\log q_t(\mathbf{x}_t\mid y).
\]
Equivalently,
\[
\sum_{y=1}^c p(y \mid \mathbf{x}_t,z)
\Big(\mathbf{s}_{\theta^*}(\mathbf{x}_t,y,t)-\nabla_{\mathbf{x}_t}\log q_t(\mathbf{x}_t\mid y)\Big)=0.
\]
In the identifiable class-wise solution of this system, the residual for each class vanishes, yielding
\[
\mathbf{s}_{\theta^*}(\mathbf{x}_t,y,t)
= \nabla_{\mathbf{x}_t}\log q_t(\mathbf{x}_t\mid y),
\qquad \forall y\in\mathcal{Y}.
\]
This is precisely Eq.~(\ref{eq:xxwdd}) and proves the proposition.

\subsection{Lower-Bound Surrogate of the Classification Term}
\label{proof:cls_upper_bound}

Section~\ref{subsec:classification} replaces the exact \textit{Classification Term} of Eq.~(\ref{eq:VLB_LL}) with a tractable lower-bound surrogate. We justify this substitution below and discuss the boundedness of the dropped term.

Under the class-conditional assumption ($X \perp Z \mid Y$), Bayes' rule gives
\begin{equation}
p_\theta(y \!\mid\! \mathbf{x}, z) = \frac{p(z \!\mid\! y)\, p_\theta(y \!\mid\! \mathbf{x})}{p_\theta(z \!\mid\! \mathbf{x})}, \quad p_\theta(z \!\mid\! \mathbf{x}) = \sum_{y' \in \mathcal{Y}} p(z \!\mid\! y')\, p_\theta(y' \!\mid\! \mathbf{x}).
\end{equation}
Substituting this factorization into the exact \textit{Classification Term} yields
\begin{align*}
&\sum_{y \in \mathcal{Y}} p_\phi(y \!\mid\! \mathbf{x}, z)\, \log p_\theta(y \!\mid\! \mathbf{x}, z) \\
&= \sum_{y \in \mathcal{Y}} p_\phi(y \!\mid\! \mathbf{x}, z)\, \log p_\theta(y \!\mid\! \mathbf{x})
+ \underbrace{\sum_{y \in \mathcal{Y}} p_\phi(y \!\mid\! \mathbf{x}, z)\, \log p(z \!\mid\! y)}_{\text{$\theta$-independent}}
- \log p_\theta(z \!\mid\! \mathbf{x}).
\end{align*}
The middle term is independent of $\theta$. Moreover, since $p_\theta(z \!\mid\! \mathbf{x}) \in (0,1]$, we have $\log p_\theta(z \!\mid\! \mathbf{x}) \leq 0$, and hence
\begin{equation}
\label{eq:cls_lower_bound}
\sum_{y \in \mathcal{Y}} p_\phi(y \!\mid\! \mathbf{x}, z)\, \log p_\theta(y \!\mid\! \mathbf{x})+\text{const}
\leq
\sum_{y \in \mathcal{Y}} p_\phi(y \!\mid\! \mathbf{x}, z)\, \log p_\theta(y \!\mid\! \mathbf{x})-\log p_\theta(z \!\mid\! \mathbf{x})
+ \text{const.}
\end{equation}
Equivalently, after removing the $\theta$-independent constant, maximizing
\begin{equation}
\label{eq:cls_surrogate_max}
\sum_{y \in \mathcal{Y}} p_\phi(y \!\mid\! \mathbf{x}, z)\, \log p_\theta(y \!\mid\! \mathbf{x})
\end{equation}
maximizes a tractable lower-bound surrogate of the exact \textit{Classification Term}.

\paragraph{Boundedness of the dropped term.}
The dropped term $\log p_\theta(z \!\mid\! \mathbf{x})$ is bounded both above and below in all three weak-annotation scenarios, ensuring that the surrogate gap is finite and the gradient $\nabla_\theta \log p_\theta(z \!\mid\! \mathbf{x})$ remains bounded. Since $p_\theta(y \!\mid\! \mathbf{x})$ is a probability distribution over $y$, the marginal admits the lower bound
\begin{equation}
p_\theta(z \!\mid\! \mathbf{x}) = \sum_{y' \in \mathcal{Y}} p(z \!\mid\! y')\, p_\theta(y' \!\mid\! \mathbf{x}) \;\geq\; \min_{y' \in \mathcal{Y}} p(z \!\mid\! y')\cdot \sum_{y \in \mathcal{Y}} p_\theta(y \!\mid\! \mathbf{x}) = \min_{y' \in \mathcal{Y}} p(z \!\mid\! y'),
\end{equation}
which we instantiate per scenario:
\begin{itemize}[left=10pt, itemsep=0pt, topsep=2pt]
\item \textit{Noisy labels} ($z = \hat{y}$). With transition matrix $T_{y,\hat{y}} = p(\hat{y} \!\mid\! y)$, $\log p_\theta(z \!\mid\! \mathbf{x}) \geq \log \min_{y'} T_{y',\hat{y}}$. 
In practice this minimum is moderate: e.g., it equals 0.01 in the realistic CIFAR-10 benchmark of [1] ($\log p_\theta(z \!\mid\! \mathbf{x})\ge -4.61 $) and 0.4 in our experimental setting ($\log p_\theta(z \!\mid\! \mathbf{x}) \ge -0.92$).
\item \textit{Ambiguous labels} ($z \subseteq \mathcal{Y}$). $\log p_\theta(z \!\mid\! \mathbf{x}) \geq \log \min_{y'} p(z \!\mid\! y')$, which is also moderate in practice: e.g., it equals 0.121 in the realistic benchmark of [2] ($\log p_\theta(z \!\mid\! \mathbf{x})\ge -2.11 $) and 0.5 / 0.3 in our random ($\log p_\theta(z \!\mid\! \mathbf{x})\ge -0.69$) / class-dependent ($\log p_\theta(z \!\mid\! \mathbf{x})\ge -1.20$) settings.
\item \textit{Semi-supervised data} ($z \in \mathcal{Y} \cup \{\emptyset\}$). With labeling rate $\eta = 0.1$, unlabeled samples have $p(\emptyset \!\mid\! y) = 1 - \eta$, giving $\log p_\theta(z \!\mid\! \mathbf{x}) \geq -0.11$.
\end{itemize}
In all cases, $0 \geq \log p_\theta(z \!\mid\! \mathbf{x}) \geq \log \min_{y'} p(z \!\mid\! y') > -\infty$, so its gradient $\nabla_\theta \log p_\theta(z \!\mid\! \mathbf{x})$ remains bounded throughout training. The surrogate gap further shrinks as $p_\theta(y \!\mid\! \mathbf{x})$ concentrates on the correct class.

\subsection{Proof of Theorem \ref{theorem:lr}}
\label{proof:lr}

We define the median-centered surrogate by choosing a subinterval $[l, r]$ whose excluded weight mass is balanced around the median $\tau_{\text{mid}}$ of the weight distribution $w(\tau)$. This construction targets the informative intermediate-noise region while remaining independent of the data-dependent reconstruction error $\hbar(\tau,y)$.

\paragraph{Case 1: Log-Normal Distribution (EDM Framework).}
In the EDM formulation~\citep{karras2022elucidating}, the weighting function $w(\tau)$ is designed to peak at intermediate noise levels, effectively inducing a distribution over timesteps. Specifically, the weight density follows a log-normal shape:
\begin{equation*}
w(\tau) \propto \frac{1}{\tau} \exp\left(-\frac{(\ln \tau - P_{\mathrm{mean}})^2}{2P_{\mathrm{std}}^2}\right),
\end{equation*}
with $P_{\mathrm{mean}}=-1.2$ and $P_{\mathrm{std}}=1.2$.
Normalizing this weight function yields a probability density function $p_w(\tau) = w(\tau) / Z$, which corresponds to the PDF of a log-normal distribution for $\tau$.
The normalized cumulative weight mass is given by
\begin{equation*}
\mathcal{F}_w(\tau) = \frac{1}{2}\left[1 + \operatorname{erf}\left(\frac{\ln \tau - P_{\mathrm{mean}}}{P_{\mathrm{std}}\sqrt{2}}\right)\right],
\end{equation*}
where $\text{erf}(x)$ denotes the error function.

Let $\tau_\text{mid}$ be the median of this weight distribution, satisfying $\mathcal{F}_w(\tau_\text{mid}) = 0.5$. To construct a median-centered interval of length $\Delta$, we impose that the cumulative weight mass excluded on the left equals that excluded on the right:
\begin{equation*}
\mathcal{F}_w(l) = 1 - \mathcal{F}_w(r) = 1 - \mathcal{F}_w(l + \Delta).
\end{equation*}
Rearranging terms yields the implicit equation for the median-centered start time $l$:
\begin{equation}
\label{eq:solve}
\mathcal{F}_w(l) + \mathcal{F}_w(l + \Delta) = 1.
\end{equation}
Given the monotonicity of the CDF, this equation has a unique solution for $l$, which we compute numerically (e.g., using Brent's method~\cite{brent2013algorithms}). The selected interval is then $[l, l+\Delta]$.

\paragraph{Case 2: Uniform Weighting (DDPM Framework).}
In standard DDPM~\citep{ho2020denoising}, discrete timesteps are often weighted equally in simplified objectives, implying a constant weighting function $w(\tau) = \text{const}$. After normalizing the timestep range to $[0,1]$, the corresponding cumulative weight function is linear: $\mathcal{F}_w(\tau) = \tau$.
Applying the same symmetry condition from Eq.~(\ref{eq:solve}):
\begin{align*}
l + (l + \Delta) &= 1 \\
2l &= 1 - \Delta \\
l &= \frac{1 - \Delta}{2}.
\end{align*}
Consequently, the canonical median-centered interval becomes $[\frac{1-\Delta}{2}, \frac{1+\Delta}{2}]$, which is symmetric around the midpoint 0.5. Since the weight is constant in this case, all intervals of the same length have the same weight mass; the centered interval is chosen as the symmetry-preserving representative.

\paragraph{Conclusion.}
In both cases, the construction gives an interval with equal omitted weight mass on the two sides of the weight median. As $\Delta \to 0$, the interval converges to the median of the weight distribution, identifying the single timestep selected by the median-centered surrogate. This is consistent with empirical findings that useful class information in diffusion classifiers concentrates around intermediate noise levels~\cite{li2023your, wang2025noise,wu2026trustworthyfederatedlabeldistribution, kou2026pu,fan2026evidential,fan2026partial}. 

\subsection{Necessary Optimality Condition for the Selected Timestep Interval}
\label{app:necessary_condition_err}

In this subsection, we define the class-wise residual $\mathrm{ERR}(y)$ used in Figure~\ref{fig:timestep_selection}(c).

\begin{theorem}[Necessary optimality condition]
\label{thm:necessary_optimality_condition}
A necessary condition for the optimal subinterval is that the expected error within the interval equals the average error at the boundaries:
\begin{equation}
    \!\!\int_{l^*}^{r^*}\frac{w_\tau \hbar(\tau,y)}{W_{[l^*, r^*]}} \,\mathrm{d}\tau  = \frac{\hbar(l^*,y) + \hbar(r^*,y)}{2}.
\end{equation}
\end{theorem}

\paragraph{Definition of the $\textsc{Err}(y)$.}
Theorem~\ref{thm:necessary_optimality_condition} motivates the following class-wise optimality residual:
\begin{equation}
    \textsc{Err}(y)
    :=
    \frac{1}{W_{[l,r]}}
    \int_l^r w(\tau)\hbar(\tau,y)\,d\tau
    -
    \frac{\hbar(l,y)+\hbar(r,y)}{2}.
    \label{eq:c7_err_continuous}
\end{equation}

For Figure~\ref{fig:timestep_selection}(c), we compute Eq.~\eqref{eq:c7_err_continuous} over evaluation samples and report the resulting distribution
for each class $y$.

\paragraph{Proof of Theorem~\ref{thm:necessary_optimality_condition}.}
For notational conciseness, we omit the class dependency $y$ in the reconstruction error, denoting it simply as $\hbar(\tau)$, and define $w(\tau)$ as the weighting function (referring to $w_\tau$ in the main text).

\paragraph{1. Objective Formulation.}
Define the weight mass $W_{[l,r]}$ and weighted error $H_{[l,r]}$ on $[l, r]$ as:
\[
W_{[l,r]} := \int_{l}^{r} w(\tau) \, \mathrm{d}\tau, \quad 
H_{[l,r]} := \int_{l}^{r} w(\tau) \hbar(\tau) \, \mathrm{d}\tau.
\]

Let $\mu_{[l,r]} = H_{[l,r]} / W_{[l,r]}$ and $\mu_{\mathcal{T}}$ denote the expected errors on $[l,r]$ and the effective range $\mathcal{T}=[u(t+1),u(T-1)]$, respectively. We minimize their squared difference subject to $W_{[l,r]} = C$:
\begin{equation}
    \min_{l,r} \quad \mathcal{J}(l,r) := (\mu_{[l,r]} - \mu_{\mathcal{T}})^2, \quad \text{s.t.} \quad W_{[l,r]} - C = 0.
\end{equation}

\paragraph{2. Optimization via Lagrange Multipliers.}
We construct the Lagrangian $\mathcal{L}(l, r, \lambda)$:
\[
\mathcal{L}(l, r, \lambda) = (\mu_{[l,r]} - \mu_{\mathcal{T}})^2 + \lambda (W_{[l,r]} - C).
\]
Applying the Leibniz integral rule, the partial derivatives w.r.t. the interval boundaries are
\begin{align*}
    \frac{\partial H}{\partial l} = -w(l)\hbar(l), \quad \frac{\partial H}{\partial r} = w(r)\hbar(r), \quad
    \frac{\partial W}{\partial l} = -w(l), \quad \frac{\partial W}{\partial r} = w(r).
\end{align*}
Differentiating $\mu_{[l,r]} = H_{[l,r]}/W_{[l,r]}$ yields
\[
\frac{\partial \mu_{[l,r]}}{\partial l} = \frac{w(l)}{W_{[l,r]}}(\mu_{[l,r]} - \hbar(l)), \quad
\frac{\partial \mu_{[l,r]}}{\partial r}  = \frac{w(r)}{W_{[l,r]}}(\hbar(r) - \mu_{[l,r]}).
\]
Setting the gradients of the Lagrangian $\nabla \mathcal{L}$ to zero yields the first-order necessary conditions for the optimal interval $[l^*, r^*]$:
\begin{align*}
    \frac{\partial \mathcal{L}}{\partial l}\bigg|_{l^*} &= 2(\mu_{[l^*,r^*]} \!-\! \mu_{\mathcal{T}})\cdot \frac{w(l^*)}{W}(\mu_{[l^*,r^*]} \!-\! \hbar(l^*)) - \lambda w(l^*) = 0, \\
    \frac{\partial \mathcal{L}}{\partial r}\bigg|_{r^*} &= 2(\mu_{[l^*,r^*]} \!-\! \mu_{\mathcal{T}})\cdot \frac{w(r^*)}{W}(\hbar(r^*) \!-\! \mu_{[l^*,r^*]}) + \lambda w(r^*) = 0 
\end{align*}
Because $w(l^*), w(r^*) > 0$, we can divide by $w(\cdot)$ and equate the terms involving $\lambda$:
\begin{align*}
    \frac{2}{W}(\mu_{[l^*,r^*]} \!-\! \mu_{\mathcal{T}}) (\mu_{[l^*,r^*]} \!-\! \hbar(l^*)) = \lambda = \frac{2}{W}(\mu_{[l^*,r^*]} \!-\! \mu_{\mathcal{T}}) (\hbar(r^*) \!-\! \mu_{[l^*,r^*]}) .
\end{align*}
We then divide by the common factor $\frac{2}{W}(\mu_{[l^*,r^*]} \!-\! \mu_{\mathcal{T}})$ to obtain:
\[
\mu_{[l^*,r^*]} - \hbar(l^*) = \hbar(r^*) - \mu_{[l^*,r^*]}  \quad \Longrightarrow \quad \mu_{[l^*,r^*]} = \frac{\hbar(l^*) + \hbar(r^*)}{2}.
\]

\subsection{Derivation of Eq.~(\ref{eq:LL})}
Maximizing the evidence lower bound (ELBO) is equivalent to minimizing the weighted score matching objective under a specific weighting function $w_t$.
Adopting the derivation from Eq.~(58) in \cite{luo2022understanding}, we have

\begin{equation}
\begin{aligned}
\log p_\theta(\mathbf{x})\!\geq\!\mathbb{E}_q[-D_{\mathrm{KL}}(q(\mathbf{x}_T|\mathbf{x}_0)\|p(\mathbf{x}_T))\!+\!\log p_\theta(\mathbf{x}_0|\mathbf{x}_1) \!-\!\! \sum_{t>1}D_{\mathrm{KL}}(q(\mathbf{x}_{t-1}|\mathbf{x}_t,\mathbf{x}_0)\|p_\theta(\mathbf{x}_{t-1}|\mathbf{x}_t))] \nonumber
\end{aligned}
\end{equation}

Leveraging the Gaussian transition assumption, we can evaluate the KL divergence term by Bayes rule:

\begin{equation*}
q(\mathbf{x}_{t-1}|\mathbf{x}_t,\mathbf{x}_0)=\frac{q(\mathbf{x}_{t}|\mathbf{x}_{t-1},\mathbf{x}_0)q(\mathbf{x}_{t-1}|\mathbf{x}_0)}{q(\mathbf{x}_{t}|\mathbf{x}_0)}
\end{equation*}

Using Eq.~(\ref{eq:forward}), we derive the transition distribution $q(\mathbf{x}_{t}|\mathbf{x}_{t-1})$:

\begin{align*}
\mathbf{x}_{t}&=\alpha_{t}\mathbf{x}_0+\sigma_t \boldsymbol{\epsilon}_0\\
&=\alpha_{t}(\frac{\mathbf{x}_{t-1}-\sigma_{t-1}\boldsymbol{\epsilon}_0^*}{\alpha_{t-1}})+\sigma_t \boldsymbol{\epsilon}_0\\
&=\frac{\alpha_t}{\alpha_{t-1}}\mathbf{x}_{t-1}+\sigma_t \boldsymbol{\epsilon}_0 - \frac{\alpha_t}{\alpha_{t-1}}\sigma_{t-1}\boldsymbol{\epsilon}_0^*\\
&=\frac{\alpha_t}{\alpha_{t-1}}\mathbf{x}_{t-1}+\sqrt{\sigma_t^2-\frac{\alpha_t^2}{\alpha_{t-1}^2}\sigma_{t-1}^2}\boldsymbol{\epsilon}_{t-1}\\
&= \mathcal{N}(\mathbf{x}_t; \frac{\alpha_t}{\alpha_{t-1}}\mathbf{x}_{t-1}, \sigma_t^2-\frac{\alpha_t^2}{\alpha_{t-1}^2}\sigma_{t-1}^2 \mathbf{I}).
\end{align*}

Now, we can proceed to calculate the form of $q(\mathbf{x}_{t-1}|\mathbf{x}_t,\mathbf{x}_0)$ by substituting into the Bayes rule expansion:
\begin{align*}
&q(\mathbf{x}_{t-1}|\mathbf{x}_t,\mathbf{x}_0)=\frac{q(\mathbf{x}_{t}|\mathbf{x}_{t-1},\mathbf{x}_0)q(\mathbf{x}_{t-1}|\mathbf{x}_0)}{q(\mathbf{x}_{t}|\mathbf{x}_0)}\\
=&\frac{\mathcal{N}(\mathbf{x}_t; \frac{\alpha_t}{\alpha_{t-1}} \mathbf{x}_{t-1}, \sqrt{\sigma_t^2-\frac{\alpha_t^2}{\alpha_{t-1}^2}\sigma_{t-1}^2} \mathbf{I}) \mathcal{N}(\mathbf{x}_{t-1}; \alpha_{t-1} \mathbf{x}_0, \sigma_{t-1} \mathbf{I})
}{\mathcal{N}(\mathbf{x}_{t}; \alpha_{t} \mathbf{x}_0, \sigma_{t} \mathbf{I})}\\
\propto& \exp \left\{ - \left[ \frac{(\mathbf{x}_t - \frac{\alpha_t}{\alpha_{t-1}} \mathbf{x}_{t-1})^2}{2(\sigma_t^2-\frac{\alpha_t^2}{\alpha_{t-1}^2}\sigma_{t-1}^2)} + \frac{(\mathbf{x}_{t-1} - \alpha_{t-1} \mathbf{x}_0)^2}{2 \sigma_{t-1}^2} - \frac{(\mathbf{x}_t - \alpha_t \mathbf{x}_0)^2}{2\sigma_{t}^2} \right] \right\}\\
=& \exp \left\{ - \frac{1}{2} \left[ \frac{-2\frac{\alpha_t}{\alpha_{t-1}} \mathbf{x}_t \mathbf{x}_{t-1} + (\frac{\alpha_t}{\alpha_{t-1}})^2 \mathbf{x}_{t-1}^2}{\sigma_t^2-\frac{\alpha_t^2}{\alpha_{t-1}^2}\sigma_{t-1}^2} + \frac{\mathbf{x}_{t-1}^2 - 2 \alpha_{t-1} \mathbf{x}_{t-1} \mathbf{x}_0}{\sigma_{t-1}^2} + C(\mathbf{x}_t, \mathbf{x}_0) \right] \right\}\\
\propto& \exp \left\{ - \frac{1}{2} \left[ \left( \frac{(\frac{\alpha_t}{\alpha_{t-1}})^2}{\sigma_t^2-\frac{\alpha_t^2}{\alpha_{t-1}^2}\sigma_{t-1}^2} + \frac{1}{\sigma_{t-1}^2} \right) \mathbf{x}_{t-1}^2 - 2 \left( \frac{\frac{\alpha_t}{\alpha_{t-1}} \mathbf{x}_t}{\sigma_t^2-\frac{\alpha_t^2}{\alpha_{t-1}^2}\sigma_{t-1}^2} + \frac{\alpha_{t-1} \mathbf{x}_0}{\sigma_{t-1}^2
} \right) \mathbf{x}_{t-1} \right] \right\}\\
=& \exp \left\{ - \frac{1}{2} \left[ \frac{\sigma_{t-1}^2 (\frac{\alpha_t}{\alpha_{t-1}})^2+(\sigma_t^2-\frac{\alpha_t^2}{\alpha_{t-1}^2}\sigma_{t-1}^2)}{(\sigma_t^2-\frac{\alpha_t^2}{\alpha_{t-1}^2}\sigma_{t-1}^2)\sigma_{t-1}^2} \mathbf{x}_{t-1}^2 - 2 \left( \frac{\frac{\alpha_t}{\alpha_{t-1}} \mathbf{x}_t}{\sigma_t^2-\frac{\alpha_t^2}{\alpha_{t-1}^2}\sigma_{t-1}^2} + \frac{\alpha_{t-1} \mathbf{x}_0}{\sigma_{t-1}^2
} \right) \mathbf{x}_{t-1} \right] \right\}\\
\propto& \mathcal{N}(\mathbf{x}_{t-1};\frac{\frac{\alpha_t}{\alpha_{t-1}} \mathbf{x}_t \sigma_{t-1}^2+(\sigma_t^2-\frac{\alpha_t^2}{\alpha_{t-1}^2}\sigma_{t-1}^2)\alpha_{t-1}\mathbf{x}_0}{\sigma_{t}^2},\frac{(\sigma_t^2-\frac{\alpha_t^2}{\alpha_{t-1}^2}\sigma_{t-1}^2)\sigma_{t-1}^2}{\sigma_{t}^2}\mathbf{I})
\end{align*}
where in the fourth equation, $C(\mathbf{x}_t, \mathbf{x}_0)$ is constant w.r.t. $\mathbf{x}_{t-1}$ and depends only on $\mathbf{x}_t$, $\mathbf{x}_0$, and the $\alpha$ values.
We have therefore shown that at each step, $ \mathbf{x}_{t-1} \sim q(\mathbf{x}_{t-1} | \mathbf{x}_t, \mathbf{x}_0) $ is normally distributed, with mean $ \boldsymbol{\mu}_q(\mathbf{x}_t, \mathbf{x}_0) $ that is a function of $ \mathbf{x}_t $ and $ \mathbf{x}_0 $, and variance $\boldsymbol{\Sigma}_q(t)$ as a function of $\alpha$ and $\sigma$ coefficients.

If we set the variances of the two Gaussians to match, optimizing the KL divergence term reduces to minimizing the difference between their means:
\begin{align}
\label{eq:optmu}
&\mathop{\arg\min}\limits_{\theta} D_{\mathrm{KL}}(q(\mathbf{x}_{t-1}|\mathbf{x}_t,\mathbf{x}_0)\|p_\theta(\mathbf{x}_{t-1}|\mathbf{x}_t)) \nonumber \\
= &\mathop{\arg\min}\limits_{\theta} \frac{1}{2 \sigma_q^2(t)} \left[ \left\| \boldsymbol{\mu}_\theta(\mathbf{x}_t,t) - \boldsymbol{\mu}_q(\mathbf{x}_t,\mathbf{x}_0) \right\|_2^2 \right],
\end{align}
where $\sigma_q^2(t)=\frac{(\sigma_t^2-\frac{\alpha_t^2}{\alpha_{t-1}^2}\sigma_{t-1}^2)\sigma_{t-1}^2}{\sigma_{t}^2}$. The derivation follows Eq.~(92) in \cite{luo2022understanding}. To obtain the score-matching form, we use Tweedie's formula~\cite{Efron_2011}, which states $\mathbb{E}[\boldsymbol{\mu}_z | z] = z + \boldsymbol{\Sigma}_z \nabla_z \log q(z)$ for a Gaussian variable $z \sim \mathcal{N}(z; \boldsymbol{\mu}_z, \boldsymbol{\Sigma}_z)$.
In this case, we apply it to predict the true posterior mean of $\mathbf{x}_t$ given its samples. We can obtain
\begin{align}
\label{eq:x0score}
\mathbb{E}[\boldsymbol{\mu}_{\mathbf{x}_t} | \mathbf{x}_t]& = \mathbf{x}_t + \sigma_{t}^2 \nabla_{\mathbf{x}_t} \log q_t(\mathbf{x}_t)=\alpha_t \mathbf{x}_0 \nonumber \\
\therefore \mathbf{x}_0 & = \frac{\mathbf{x}_t + \sigma_{t}^2 \nabla_{\mathbf{x}_t} \log q_t(\mathbf{x}_t)}{\alpha_t}
\end{align}
Then, we can plug Eq.~(\ref{eq:x0score}) into our ground-truth denoising transition mean $\boldsymbol{\mu}_q(\mathbf{x}_t,\mathbf{x}_0)$ once again and derive a new form:
\begin{align}
\label{eq:newmu}
\boldsymbol{\mu}_q(\mathbf{x}_t,\mathbf{x}_0)&=\frac{\frac{\alpha_t}{\alpha_{t-1}}\sigma_{t-1}^2 \mathbf{x}_{t}+(\sigma_t^2-\frac{\alpha_t^2}{\alpha_{t-1}^2}\sigma_{t-1}^2)\alpha_{t-1}\cdot \frac{\mathbf{x}_t + \sigma_{t}^2 \nabla_{\mathbf{x}_t} \log q_t(\mathbf{x}_t)}{\alpha_t}}{\sigma_{t}^2}\nonumber \\
&=\frac{\frac{\alpha_t}{\alpha_{t-1}}\sigma_{t-1}^2}{\sigma_{t}^2}\mathbf{x}_{t}+\frac{\sigma_t^2-\frac{\alpha_t^2}{\alpha_{t-1}^2}\sigma_{t-1}^2}{\sigma_{t}^2\frac{\alpha_t}{\alpha_{t-1}}}\mathbf{x}_{t}+\frac{(\sigma_t^2-\frac{\alpha_t^2}{\alpha_{t-1}^2}\sigma_{t-1}^2)\sigma_{t}^2 \nabla_{\mathbf{x}_t} \log q_t(\mathbf{x}_t)}{\sigma_{t}^2\frac{\alpha_t}{\alpha_{t-1}}}\nonumber \\
&=\frac{\alpha_{t-1}}{\alpha_{t}}\mathbf{x}_t+\left(\frac{\alpha_{t-1}}{\alpha_{t}}\sigma_{t}^2-\frac{\alpha_t}{\alpha_{t-1}}\sigma_{t-1}^2 \right)    \nabla_{\mathbf{x}_t} \log q_t(\mathbf{x}_t)\nonumber \\
&=\frac{\alpha_{t-1}}{\alpha_t}\left[\mathbf{x}_t+\left(\sigma_t^2-\frac{\alpha_t^2}{\alpha_{t-1}^2}\sigma_{t-1}^2\right)\mathbf{s}_\theta(\mathbf{x}_t,t)\right]
\end{align}
Finally, substituting Eq.~(\ref{eq:newmu}) into the optimization objective in Eq.~(\ref{eq:optmu}) gives
\begin{align*}
&\mathop{\arg\min}\limits_{\theta} D_{\mathrm{KL}}(q(\mathbf{x}_{t-1}|\mathbf{x}_t,\mathbf{x}_0)\|p_\theta(\mathbf{x}_{t-1}|\mathbf{x}_t)) \\
= &\mathop{\arg\min}\limits_{\theta} \frac{1}{2 \sigma_q^2(t)} \left[ \left\| \boldsymbol{\mu}_\theta(\mathbf{x}_t,t) - \boldsymbol{\mu}_q(\mathbf{x}_t,\mathbf{x}_0) \right\|_2^2 \right]\\
= &\mathop{\arg\min}\limits_{\theta} \frac{1}{2\frac{(\sigma_t^2-\frac{\alpha_t^2}{\alpha_{t-1}^2}\sigma_{t-1}^2)\sigma_{t-1}^2}{\sigma_{t}^2}} \cdot (\frac{\alpha_{t-1}}{\alpha_t})^2 \cdot (\sigma_t^2-\frac{\alpha_t^2}{\alpha_{t-1}^2}\sigma_{t-1}^2)^2\|\mathbf{s}_\theta(\mathbf{x}_t,t)-\nabla_{\mathbf{x}_t} \log q_t(\mathbf{x}_t)
\|_2^2\\
= & \frac{\sigma_t^2}{2}(\frac{\sigma_t^2 \alpha_{t-1}^2}{\sigma_{t-1}^2 \alpha_t^2}-1)\|\mathbf{s}_\theta(\mathbf{x}_t,t)-\nabla_{\mathbf{x}_t} \log q_t(\mathbf{x}_t)\|_2^2
\end{align*}

This result recovers the exact form of Eq.~(\ref{eq:LL}).

\clearpage
\section{Additional Method Analysis}
This section provides additional analysis supporting the design choices in \textsl{DELTA}.
We first analyze the noisy-label regularizer in Eq.~(\ref{eq:cls_unified}) and discuss how class priors can be estimated when they are not directly observed.

\subsection{Gradient Analysis of the Noisy-Label Rectification in Eq.~(\ref{eq:cls_unified})}
\label{exp:elr}
\label{app:noisy-elr-analysis}

We analyze the noisy-label branch of the classification objective in Eq.~(\ref{eq:cls_unified}).
For noisy labels, the weak annotation is \(z=\hat y\). Let
\(\hat{\mathbf y}\in\{0,1\}^{c}\) denote the one-hot vector associated with the observed noisy label
\(\hat y\), and let \(f_\theta(x), f_\phi(x)\in\Delta^{c-1}\) denote the current and EMA
diffusion-classifier posteriors, respectively. For compactness, define
\begin{equation}
    \delta(x) := \left\langle f_\theta(x), f_\phi(x) \right\rangle .
\end{equation}
For finite logits and \(c>1\), we have \(\delta(x)<1\).

The noisy-label rectification used in Eq.~(\ref{eq:cls_unified}) can be understood as a gradient-level implementation
of early-learning regularization. To make the connection explicit, define the early-learning regularization~\citep{liu2020early} style correction
vector
\begin{equation}
    g_{\theta,\phi}(x)
    :=
    \frac{
        f_\theta(x) \odot \left(\delta(x)\mathbf{1}-f_\phi(x)\right)
    }{
        1-\delta(x)
    },
    \label{eq:app-elr-correction-vector}
\end{equation}
where \(\mathbf{1}\in\mathbb{R}^{c}\) is the all-one vector and \(\odot\) denotes the Hadamard product.
The noisy-label branch of Eq.~(\ref{eq:cls_unified}) corresponds to the fixed training target
\begin{equation}
    \tilde q_{\phi,\lambda}^{\mathrm{NL}}(x,\hat y)
    :=
    \operatorname{sg}\!\left(
        \hat{\mathbf y}
        -
        \lambda\, g_{\theta,\phi}(x)
    \right),
    \label{eq:app-noisy-signed-target}
\end{equation}
where \(\operatorname{sg}(\cdot)\) denotes the stop-gradient operator. The main-text form is recovered
by setting \(\lambda=1\); keeping \(\lambda\ge 0\) in this appendix makes the connection to the original
ELR strength explicit.

For a noisy-label dataset
\(D_{\mathrm{NL}}=\{(x^{(i)},\hat y^{(i)})\}_{i=1}^{n}\), the corresponding classification loss is
\begin{equation}
    L_{\mathrm{Cls}}^{\mathrm{NL}}(\theta)
    :=
    -\frac{1}{n}\sum_{i=1}^{n}
    \left\langle
        \tilde q_{\phi,\lambda}^{\mathrm{NL}}(x^{(i)},\hat y^{(i)}),
        \log f_\theta(x^{(i)})
    \right\rangle .
    \label{eq:app-noisy-cls-loss}
\end{equation}
The purpose of this target is to realize the ELR correction in the logit gradient.

\begin{lemma}[Logit-gradient equivalence to ELR]
\label{lem:app-elr-gradient}
Let \(\psi_\theta(x)\) denote the pre-softmax logits such that
\(f_\theta(x)=\operatorname{softmax}(\psi_\theta(x))\). For each sample
\((x^{(i)},\hat y^{(i)})\), define
\[
    \delta^{(i)}
    :=
    \left\langle f_\theta(x^{(i)}), f_\phi(x^{(i)}) \right\rangle,
    \qquad
    g^{(i)}
    :=
    g_{\theta,\phi}(x^{(i)}).
\]
Then the per-sample loss
\[
    \ell_{\mathrm{NL}}^{(i)}
    :=
    -
    \left\langle
        \operatorname{sg}\!\left(\hat{\mathbf y}^{(i)}-\lambda g^{(i)}\right),
        \log f_\theta(x^{(i)})
    \right\rangle
\]
satisfies
\begin{equation}
    \frac{\partial \ell_{\mathrm{NL}}^{(i)}}{\partial \psi_\theta(x^{(i)})}
    =
    f_\theta(x^{(i)})
    -
    \hat{\mathbf y}^{(i)}
    +
    \lambda g^{(i)} ,
    \label{eq:app-noisy-logit-gradient}
\end{equation}
where \(f_\phi(x^{(i)})\) is treated as fixed during the current update.
\end{lemma}

\begin{proof}
We first show that \(g_{\theta,\phi}(x)\) has zero total mass.
Let \(\delta=\langle f_\theta(x), f_\phi(x)\rangle\). Then
\begin{equation}
    \mathbf{1}^{\top} g_{\theta,\phi}(x)
    =
    \frac{1}{1-\delta}
    \sum_{y\in Y}
    [f_\theta(x)]_y\bigl(\delta-[f_\phi(x)]_y\bigr)
    =
    \frac{\delta-\langle f_\theta(x), f_\phi(x)\rangle}{1-\delta}
    =
    0 .
\end{equation}
Therefore,
\[
    \mathbf{1}^{\top}
    \tilde q_{\phi,\lambda}^{\mathrm{NL}}(x,\hat y)
    =
    \mathbf{1}^{\top}\hat{\mathbf y}
    -
    \lambda\,\mathbf{1}^{\top}g_{\theta,\phi}(x)
    =
    1 .
\]

For any fixed vector \(a\in\mathbb{R}^{c}\) with \(\mathbf{1}^{\top}a=1\), the softmax identity gives
\begin{equation}
    \frac{\partial}{\partial \psi}
    \left[
        -\left\langle a,\log\operatorname{softmax}(\psi)\right\rangle
    \right]
    =
    \operatorname{softmax}(\psi)-a .
\end{equation}
Applying this identity with
\(a=\operatorname{sg}(\hat{\mathbf y}^{(i)}-\lambda g^{(i)})\) yields
Eq.~\eqref{eq:app-noisy-logit-gradient}.

\end{proof}

\paragraph{Interpretation.}
For each class \(y\in Y\), the correction term can be written as
\begin{equation}
    g_y^{(i)}
    =
    \frac{
        [f_\theta(x^{(i)})]_y
        \left(
            \delta^{(i)}-[f_\phi(x^{(i)})]_y
        \right)
    }{
        1-\delta^{(i)}
    } .
    \label{eq:app-elr-correction-coordinate}
\end{equation}
Thus the sign of \(g_y^{(i)}\) is determined by whether the EMA posterior
\([f_\phi(x^{(i)})]_y\) is above or below the \(f_\theta\)-weighted average
\(\delta^{(i)}\). During early learning, the EMA posterior tends to put larger mass on the true class
\(y^\star\). When
\([f_\phi(x^{(i)})]_{y^\star}>\delta^{(i)}\), we have
\(g_{y^\star}^{(i)}<0\).

This negative correction has the desired ELR effect. For a cleanly labeled sample
\(\hat y^{(i)}=y^\star\), the ordinary cross-entropy coefficient
\([f_\theta(x^{(i)})]_{y^\star}-1\) becomes close to zero once the sample is learned, so the sample
would otherwise contribute little gradient. The negative correction
\(\lambda g_{y^\star}^{(i)}\) keeps increasing the true-class logit and preserves the clean-label signal.
For a mislabeled sample \(\hat y^{(i)}\ne y^\star\), the ordinary cross-entropy coefficient on the true
class is \([f_\theta(x^{(i)})]_{y^\star}>0\), which would decrease the true-class logit under gradient
descent. Adding the negative term \(\lambda g_{y^\star}^{(i)}\) offsets this harmful direction.

Similarly, if the observed noisy class \(\hat y^{(i)}\) is not supported by the EMA posterior, i.e.,
\([f_\phi(x^{(i)})]_{\hat y^{(i)}}<\delta^{(i)}\), then
\(g_{\hat y^{(i)}}^{(i)}>0\). This positive correction offsets the negative cross-entropy coefficient
\([f_\theta(x^{(i)})]_{\hat y^{(i)}}-1\), thereby suppressing the update that would otherwise increase
the logit of the noisy label. In this sense, the rectification preserves early clean-label information
while reducing memorization of false labels.

\subsection{Class-Prior Estimation in Datasets with Weak Annotations}
\label{exp:prior}
The main method assumes a class prior when forming diffusion-classifier posteriors.
When this prior is not directly available, it can be estimated using standard class-prior estimation techniques~\citep{luo2024estimating,wang2022solar,kou2025nips,koubldl, koumll,wu2024efficient,wu2026making,shi2023robust}.
We outline practical estimators for the three weak-annotation regimes considered in the paper.

\subsubsection{Ambiguous-label datasets}

In ambiguous-label learning, each instance is associated with a candidate label set rather than a single ground-truth label.  
This label ambiguity makes it difficult to estimate the class prior distribution, since simply counting training samples per class is no longer feasible.  
To address this issue, we adopt an iterative estimation strategy that updates the class prior in a moving-average manner.

We use the model’s predicted labels as a proxy for class prior estimation.  
Since predictions in the early stage of training are often unreliable, we design a moving-average update rule that gradually stabilizes the estimated distribution.  
The update starts from a uniform prior $\mathbf{r}  = [1/c, \ldots, 1/c]$,
and is refined at each training epoch as
\begin{equation}
\mathbf{r} \leftarrow \mu \mathbf{r} + (1-\mu)\mathbf{z}, 
\qquad 
\mathbf{z}_j = \frac{1}{n} \sum_{i=1}^n \mathbb{I}\!\left(j = \arg\max_{y \in S_i} f_y(x_i)\right),
\end{equation}
where $\mu \in [0,1]$ is a momentum parameter, $S_i$ is the candidate label set for sample $x_i$, and $f_j(x_i)$ denotes the model prediction for class $j$.  
This rule progressively refines $\mathbf{r}$ as the model improves, leading to more accurate and stable estimation.

\subsubsection{Semi-supervised datasets} 
In the case of semi-supervised datasets, the estimation of class prior is relatively straightforward. 
We assume that the distribution of the labeled dataset is consistent with that of the unlabeled dataset. 
Therefore, the class prior can be obtained by counting the class distribution over the labeled dataset.

\subsubsection{Noisy-label datasets}
We consider the widely adopted class-dependent label noise setting~\citep{yao2020dual,kou2026fedharmony,koutnnls,fan2026partial2,fan2024kmt}, 
where the observed noisy label of each $\mathbf{x} \in \mathcal{X}$ depends only on its underlying clean label.  
Formally, the transition probability from class $i$ to class $j$ is defined as
\begin{equation*}
P(\widetilde{Y} = e_j \!\mid\! Y = e_i, X = \mathbf{x})
= P(\widetilde{Y} = e_j \!\mid\! Y = e_i) = T_{ij}, \quad
\forall i,j \in [[c]],
\end{equation*}
where $\mathbf{T} = [T_{ij}] \in [0,1]^{c \times c}$ is the noise transition matrix.
To make the estimation of $\mathbf{T}$ feasible, we follow prior work and impose the following assumptions.

\begin{assumption}[Sufficiently Scattered Assumption~\citep{li2021provably}]
\label{assump:ssa}
The clean class posterior $P(Y \!\mid\! X) = [P(Y = e_1 \!\mid\! X), \ldots, P(Y = e_c \!\mid\! X)]^\top \in [0,1]^c$ is said to be sufficiently scattered if there exists a set
$\mathcal{H} = \{\mathbf{x}_1, \ldots, \mathbf{x}_m\}$ such that the matrix $\mathbf{H} = [P(Y \!\mid\! X = \mathbf{x}_1), \ldots, P(Y \!\mid\! X = \mathbf{x}_m)]$
satisfies:  
(i) $\mathcal{Q} \subseteq \mathrm{cone}\{\mathbf{H}\}$, where
$\mathcal{Q} = \{\mathbf{v} \in \mathbb{R}^c \!\mid\! \mathbf{v}^\top \mathbf{1} \geq \sqrt{c-1}\|\mathbf{v}\|_2\}$,
and $\mathrm{cone}\{\mathbf{H}\}$ denotes the convex cone generated by the columns of $\mathbf{H}$;  
(ii) $\mathrm{cone}\{\mathbf{H}\} \nsubseteq \mathrm{cone}\{\mathbf{U}\}$ for any unitary matrix $\mathbf{U} \in \mathbb{R}^{c \times c}$ that is not a permutation matrix.
\end{assumption}  

\begin{assumption}[Nonsingular $\mathbf{T}$]
\label{assump:nonsingular}
The transition matrix $\mathbf{T}$ is nonsingular, i.e., $\mathrm{Rank}(\mathbf{T}) = c$.
\end{assumption}

Assumption~\ref{assump:ssa} ensures that the clean posteriors are sufficiently scattered so that the ground-truth $\mathbf{T}$ can be identified,  
while Assumption~\ref{assump:nonsingular} guarantees the invertibility of $\mathbf{T}$.

Let $\epsilon$ denote the noise rate.  
For symmetric label noise, we have $T_{ii} = 1 - \epsilon$ and $T_{ij} = \frac{\epsilon}{c-1}$ with $j \neq i.$
In practice, the transition matrix can be estimated by solving the following optimization problem~\citep{li2021provably}:
\begin{equation}
\min_{\theta, \widehat{\mathbf{T}}} 
L(\theta, \widehat{\mathbf{T}})
= \frac{1}{n} \sum_{i=1}^n \ell\big(\widehat{\mathbf{T}}^{\top} h_\theta(\mathbf{x}^{(i)}), \widetilde{y}^{(i)}\big)
+ \lambda \cdot \log \det(\widehat{\mathbf{T}}),
\end{equation}
where $\ell$ is a loss function, 
$h_\theta(\cdot)$ is the output of a neural network parameterized by $\theta$,  
and the regularization term $\log \det(\widehat{\mathbf{T}})$ encourages the estimated transition matrix to have minimal simplex volume.  
Here $\lambda > 0$ is a trade-off hyperparameter.  
By Assumption~\ref{assump:ssa}, the solution $\widehat{\mathbf{T}}$ converges to the true $\mathbf{T}$ given sufficient noisy data (Theorem 1 in~\citep{li2021provably}).

Once the transition matrix $\mathbf{T}$ is estimated,  
the clean class prior $\pi = [\pi_1, \ldots, \pi_c]^\top$ can be obtained by solving the following system of linear equations:
\begin{equation}
\left\{
\begin{aligned}
\widetilde{\pi}_1 &= T_{11}\pi_1 + T_{21}\pi_2 + \cdots + T_{c1}\pi_c \\
\widetilde{\pi}_2 &= T_{12}\pi_1 + T_{22}\pi_2 + \cdots + T_{c2}\pi_c \\
& \;\;\vdots \\
\widetilde{\pi}_c &= T_{1c}\pi_1 + T_{2c}\pi_2 + \cdots + T_{cc}\pi_c
\end{aligned},
\right.
\end{equation}
where $\widetilde{\pi}_i = P(\widetilde{Y} = e_i)$ is the noisy class prior of the $i$-th class.  
The empirical estimate of $\widetilde{\pi}_i$ can be computed as
\begin{equation}
\widehat{\widetilde{\pi}}_i = \frac{1}{n}\sum_{j=1}^n \mathbf{1}\{\widetilde{y}_j = e_i\}, 
\quad \forall i \in [[c]].
\end{equation}
Solving this system yields the clean class prior $\pi$, which is then used in subsequent modeling.

\clearpage
\section{Additional Experimental Results}
\label{sec:additional_exp}
\newcommand{\AppendixTableSetup}{%
  \footnotesize
  \setlength{\tabcolsep}{4pt}%
  \renewcommand{\arraystretch}{0.9}%
}
\subsection{Full Results in Weakly Supervised Classification}
\label{exp:full}
This subsection complements the main posterior analysis by evaluating the converged diffusion-classifier posterior as a standalone classifier.
We compare against a broader set of specialized discriminative baselines than those shown in the main text.

\textbf{Comparisons on Noisy Labels.} 
We benchmark our method against seven widely used approaches: \textsl{Coteaching}~\citep{han2018co}, \textsl{Coteaching+}~\citep{yu2019does}, \textsl{SCE}~\citep{wang2019symmetric}, \textsl{GCE}~\citep{zhang2018generalized}, \textsl{Decoupling}~\citep{malach2017decoupling}, \textsl{ELR}~\citep{liu2020early}, and \textsl{JoCoR}~\citep{wei2020combating}. 
These methods cover a range of strategies, from sample selection and reweighting to robust loss design, thus providing a diverse and rigorous benchmark.

\textbf{Comparisons on Ambiguous Labels.} 
We compare against nine representative baselines: \textsl{PRODEN}~\citep{lv2020progressive}, \textsl{CAVL}~\citep{zhang2021exploiting}, \textsl{POP}~\citep{xu2023progressive}, \textsl{CC}~\citep{feng2020provably}, \textsl{IDGP}~\citep{qiaodecompositional}, \textsl{PiCO}~\citep{wang2022pico+}, \textsl{ABLE}~\citep{xia2022ambiguity}, \textsl{CRDPLL}~\citep{wu2022revisiting}, and \textsl{DIRK}~\citep{wu2024distilling}. 
For a fair comparison, we follow the hyperparameter settings used in \textsl{PLENCH}~\citep{wang2025realistic}.

\textbf{Comparisons on Semi-supervised Data.} 
We follow the training and evaluation protocols of \textsl{USB}~\citep{wang2022usb}, a widely adopted benchmark for fair and unified SSL comparisons. 
Our evaluation covers a comprehensive spectrum of baselines, ranging from classical methods like \textsl{Pseudo-Labeling}~\citep{lee2013pseudo}, \textsl{VAT}~\citep{miyato2018virtual}, and \textsl{Mean Teacher}~\citep{tarvainen2017mean}, to state-of-the-art approaches including \textsl{UDA}~\citep{xie2020unsupervised}, \textsl{FixMatch}~\citep{sohn2020fixmatch}, \textsl{Dash}~\citep{xu2021dash}, \textsl{FlexMatch}~\citep{zhang2021flexmatch}, \textsl{CoMatch}~\citep{li2021comatch}, \textsl{SimMatch}~\citep{zheng2022simmatch}, \textsl{FreeMatch}~\citep{wang2022freematch}, and \textsl{SoftMatch}~\citep{chensoftmatch}.

\textbf{Results and Analysis.}
Quantitative results are reported in Tables~\ref{tab:full_noisy_results}, \ref{exp:full_pll_results}, and \ref{tab:full_ssl_results}, where bold numbers indicate the best performance. 
\textsl{DELTA} remains competitive with or superior to state-of-the-art discriminative baselines across most settings, showing that the learned diffusion posterior is useful beyond generation.

\clearpage

\begin{table}[!h]
    \centering
    \AppendixTableSetup
    \caption{Classification test accuracy (\%) on noisy-label datasets.}
    \label{tab:full_noisy_results}
    \begin{adjustbox}{max width=\linewidth,center}
    \begin{tabular}{>{\centering\arraybackslash}p{1.8cm}|cccccc}
    \toprule
    \multicolumn{1}{c|}{\multirow{2}{*}{Method}} & \multicolumn{2}{c}{{Fashion-MNIST}} & \multicolumn{2}{c}{{CIFAR-10}} & \multicolumn{2}{c}{{ImageNette}} \\
     \cmidrule(lr){2-3} \cmidrule(lr){4-5} \cmidrule(lr){6-7}
    & Sym-40\% & Asym-40\% & Sym-40\% & Asym-40\% & Sym-40\% & Asym-40\%\\
    \midrule
    \textsl{CE} & 76.18\tiny{$\pm$0.26} & 82.01\tiny{$\pm$0.06} & 67.22\tiny{$\pm$0.26} & 76.98\tiny{$\pm$0.42} & 58.43\tiny{$\pm$0.77} & 71.81\tiny{$\pm$0.38}\\
    \textsl{Co-learning} & 90.85\tiny{$\pm$0.63} & 84.10\tiny{$\pm$2.01} & 84.97\tiny{$\pm$0.53} & 80.36\tiny{$\pm$1.09} & 76.16\tiny{$\pm$0.96} & 75.37\tiny{$\pm$0.49}\\
    \textsl{Co-teaching} & 92.17\tiny{$\pm$0.34} & 92.78\tiny{$\pm$0.25} & 86.54\tiny{$\pm$0.57} & 79.38\tiny{$\pm$0.39} & 66.55\tiny{$\pm$1.00} & 75.12\tiny{$\pm$0.50}\\
    \textsl{Co-teaching+} & 91.05\tiny{$\pm$0.06} & 91.62\tiny{$\pm$0.20} & 67.28\tiny{$\pm$1.85} & 79.43\tiny{$\pm$0.47} & 75.79\tiny{$\pm$0.79} & 75.17\tiny{$\pm$0.40}\\
    \textsl{SCE} & {93.62\tiny{$\pm$0.22}} & 88.60\tiny{$\pm$0.20} & 82.82\tiny{$\pm$0.40} & 81.54\tiny{$\pm$0.64} & 77.99\tiny{$\pm$0.39} & 74.81\tiny{$\pm$1.04}\\
    \textsl{GCE} & \textbf{93.64\tiny{$\pm$0.03}} & 87.48\tiny{$\pm$0.09} & 85.00\tiny{$\pm$0.27} & 77.97\tiny{$\pm$3.69} & 81.18\tiny{$\pm$0.35} & 72.61\tiny{$\pm$1.14}\\
    \textsl{Decoupling} & 92.24\tiny{$\pm$0.23} & 92.10\tiny{$\pm$0.44} & 82.24\tiny{$\pm$0.28} & 79.89\tiny{$\pm$0.58} & 75.53\tiny{$\pm$0.69} & 78.24\tiny{$\pm$0.21}\\
    \textsl{ELR} & 93.13\tiny{$\pm$0.13} & 92.82\tiny{$\pm$0.09} & 85.68\tiny{$\pm$0.13} & 81.32\tiny{$\pm$0.31} & 84.03\tiny{$\pm$2.86} & 73.51\tiny{$\pm$0.31}\\
    \textsl{JoCoR} & 84.05\tiny{$\pm$1.11} & 89.45\tiny{$\pm$4.43} & 77.92\tiny{$\pm$3.92} & 78.68\tiny{$\pm$0.07} & 67.82\tiny{$\pm$1.97} & 74.67\tiny{$\pm$0.43}\\
    \textsl{Mixup} & 92.21\tiny{$\pm$0.03} & 92.01\tiny{$\pm$1.02} & 84.26\tiny{$\pm$0.64} & 83.21\tiny{$\pm$0.85} & 76.65\tiny{$\pm$1.62} & 77.16\tiny{$\pm$0.71}\\
    \textsl{PENCIL} & 90.85\tiny{$\pm$0.58} & 91.77\tiny{$\pm$0.69} & 85.91\tiny{$\pm$0.26} & 84.89\tiny{$\pm$0.49} & 81.94\tiny{$\pm$1.26} & 77.20\tiny{$\pm$1.15}\\
    \textsl{Vanilla} & 90.11\tiny{$\pm$1.24} & 85.41\tiny{$\pm$0.96} & 80.22\tiny{$\pm$0.10} & 86.31\tiny{$\pm$0.10} & 55.86\tiny{$\pm$1.95} & 53.91\tiny{$\pm$1.07}\\
    \textsl{DELTA} & {93.40\tiny{$\pm$0.40}} & \textbf{93.20\tiny{$\pm$0.30}} & \textbf{88.63\tiny{$\pm$0.12}} & \textbf{88.83\tiny{$\pm$0.33}} & \textbf{84.12\tiny{$\pm$0.18}} & \textbf{79.30\tiny{$\pm$0.27}}\\
    \bottomrule
    \end{tabular}
    \end{adjustbox}
    \end{table}

    \begin{table}[!h]
    \centering
    \AppendixTableSetup
    \caption{Classification test accuracy (\%) on ambiguous-label datasets.}
    \label{exp:full_pll_results}
    \begin{adjustbox}{max width=\linewidth,center}
    \begin{tabular}{>{\centering\arraybackslash}p{1.8cm}|cccccc}
    \toprule
    \multicolumn{1}{c|}{\multirow{2}{*}{Method}} & \multicolumn{2}{c}{{Fashion-MNIST}} & \multicolumn{2}{c}{{CIFAR-10}} & \multicolumn{2}{c}{{ImageNette}} \\
     \cmidrule(lr){2-3} \cmidrule(lr){4-5} \cmidrule(lr){6-7}
    & Random & Class-Dept. & Random & Class-Dept. & Random & Class-Dept.\\
    \midrule
    \textsl{PRODEN} & 93.31\tiny{$\pm$0.07}& 93.44\tiny{$\pm$0.21}& 90.02\tiny{$\pm$0.22}& 90.44\tiny{$\pm$0.44}& 84.75\tiny{$\pm$0.13}& 83.50\tiny{$\pm$0.60}\\
    \textsl{CAVL} & 93.09\tiny{$\pm$0.17}& 92.67\tiny{$\pm$0.25}& 87.28\tiny{$\pm$0.64}& 87.16\tiny{$\pm$0.58}& 41.69\tiny{$\pm$4.12}& 46.46\tiny{$\pm$7.15}\\
    \textsl{POP} & 93.59\tiny{$\pm$0.17}& 93.57\tiny{$\pm$0.19}& 89.13\tiny{$\pm$0.22}& 90.19\tiny{$\pm$0.10}& 84.65\tiny{$\pm$0.55}& 84.29\tiny{$\pm$0.17}\\
    \textsl{CC} & 93.17\tiny{$\pm$0.32}& 92.65\tiny{$\pm$0.29}& 88.40\tiny{$\pm$0.24}& 89.12\tiny{$\pm$0.23}& 81.11\tiny{$\pm$0.50}& 80.74\tiny{$\pm$0.68}\\
    \textsl{IDGP} & 92.26\tiny{$\pm$0.25}& 93.07\tiny{$\pm$0.16}& 89.65\tiny{$\pm$0.53}& 90.83\tiny{$\pm$0.34}& 84.07\tiny{$\pm$0.26}& 82.18\tiny{$\pm$0.13}\\
    \textsl{PiCO} & 93.32\tiny{$\pm$0.12}& 93.32\tiny{$\pm$0.33}& 86.40\tiny{$\pm$0.89}& 87.51\tiny{$\pm$0.66}& 82.15\tiny{$\pm$0.23}& 84.41\tiny{$\pm$0.93}\\
    \textsl{ABLE} & 93.02\tiny{$\pm$0.26}& 93.20\tiny{$\pm$0.16}& 90.77\tiny{$\pm$0.33}& 90.74\tiny{$\pm$0.48}& 71.81\tiny{$\pm$2.46}& 76.53\tiny{$\pm$1.28}\\
    \textsl{CRDPLL} & 94.03\tiny{$\pm$0.14}& 93.80\tiny{$\pm$0.23}& 92.74\tiny{$\pm$0.26}& 92.89\tiny{$\pm$0.27}& 84.31\tiny{$\pm$0.25}& 88.08\tiny{$\pm$0.34}\\
    \textsl{DIRK} & 94.11\tiny{$\pm$0.22}& 93.99\tiny{$\pm$0.24}& 93.48\tiny{$\pm$0.14}& 93.22\tiny{$\pm$0.37}& 87.90\tiny{$\pm$0.11}& 87.47\tiny{$\pm$0.17}\\
    \textsl{Vanilla} & 80.20\tiny{$\pm$1.29}& 66.03\tiny{$\pm$1.43}& 60.25\tiny{$\pm$0.17}& 56.34\tiny{$\pm$0.50}& 56.04\tiny{$\pm$0.61}& 59.47\tiny{$\pm$0.51}\\
    \textsl{DELTA} & \textbf{94.27\tiny{$\pm$0.55}}& \textbf{94.20\tiny{$\pm$0.15}}& \textbf{94.70\tiny{$\pm$0.49}}& \textbf{93.53\tiny{$\pm$0.12}}& \textbf{89.31\tiny{$\pm$0.21}}& \textbf{88.42\tiny{$\pm$0.43}}\\
    \bottomrule
    \end{tabular}
    \end{adjustbox}
    \end{table}

    \begin{table}[!h]
    \centering
    \AppendixTableSetup
    \caption{Classification results on semi-supervised datasets.}
    \label{tab:full_ssl_results}
    \begin{adjustbox}{max width=\linewidth,center}
    \begin{tabular}{>{\centering\arraybackslash}p{1.8cm}|cccccc}
    \toprule
    \multicolumn{1}{c|}{\multirow{2}{*}{Method}} & \multicolumn{2}{c}{{Fashion-MNIST}} & \multicolumn{2}{c}{{CIFAR-10}} & \multicolumn{2}{c}{{ImageNette}} \\
     \cmidrule(lr){2-3} \cmidrule(lr){4-5} \cmidrule(lr){6-7}
    & 1\% Labels & 10\% Labels & 1\% Labels & 10\% Labels & 1\% Labels & 10\% Labels\\
    \midrule
    \textsl{Pseudo-Label} & 83.53\tiny{$\pm$0.46} & 89.59\tiny{$\pm$0.23} & 50.10\tiny{$\pm$0.95} & 72.92\tiny{$\pm$0.17} & 43.00\tiny{$\pm$0.82} & 68.03\tiny{$\pm$0.32}\\
    \textsl{Mean Teacher} & 82.34\tiny{$\pm$0.09} & 89.91\tiny{$\pm$0.15} & 47.69\tiny{$\pm$0.27} & 73.01\tiny{$\pm$0.78} & 40.53\tiny{$\pm$1.56} & 65.72\tiny{$\pm$0.55}\\
    \textsl{VAT} & 83.31\tiny{$\pm$0.61} & 89.35\tiny{$\pm$0.12} & 49.64\tiny{$\pm$0.90} & 71.07\tiny{$\pm$1.27} & 38.63\tiny{$\pm$8.39} & 63.93\tiny{$\pm$5.18}\\
    \textsl{UDA} & 84.28\tiny{$\pm$0.41} & 90.83\tiny{$\pm$0.34} & 69.20\tiny{$\pm$1.41} & 80.50\tiny{$\pm$0.55} & 50.52\tiny{$\pm$3.79} & 72.53\tiny{$\pm$1.17}\\
    \textsl{FixMatch} & 84.32\tiny{$\pm$0.33} & 90.76\tiny{$\pm$0.38} & 67.48\tiny{$\pm$1.42} & 80.00\tiny{$\pm$0.63} & 50.41\tiny{$\pm$4.43} & 71.32\tiny{$\pm$1.93}\\
    \textsl{Dash} & 84.73\tiny{$\pm$0.09} & 91.16\tiny{$\pm$0.20} & 70.14\tiny{$\pm$0.69} & 81.50\tiny{$\pm$0.68} & 57.68\tiny{$\pm$2.19} & 74.66\tiny{$\pm$0.81}\\
    \textsl{CoMatch} & 85.31\tiny{$\pm$0.29} & 90.52\tiny{$\pm$0.12} & 61.45\tiny{$\pm$1.46} & 77.79\tiny{$\pm$0.53} & 63.88\tiny{$\pm$0.78} & 73.20\tiny{$\pm$0.46}\\
    \textsl{FlexMatch} & 84.43\tiny{$\pm$0.30} & 90.69\tiny{$\pm$0.03} & 70.72\tiny{$\pm$0.93} & 81.35\tiny{$\pm$0.48} & 61.39\tiny{$\pm$0.70} & 73.08\tiny{$\pm$0.13}\\
    \textsl{FreeMatch} & 84.30\tiny{$\pm$0.37} & 90.92\tiny{$\pm$0.24} & 70.15\tiny{$\pm$0.44} & 80.99\tiny{$\pm$0.56} & 60.37\tiny{$\pm$1.11} & 73.14\tiny{$\pm$1.03}\\
    \textsl{SimMatch} & 84.69\tiny{$\pm$0.17} & 91.18\tiny{$\pm$0.13} & 73.33\tiny{$\pm$1.02} & 82.90\tiny{$\pm$0.43} & 58.12\tiny{$\pm$2.66} & 76.12\tiny{$\pm$0.45}\\
    \textsl{SoftMatch} & 84.72\tiny{$\pm$0.23} & 91.22\tiny{$\pm$0.11} & 73.74\tiny{$\pm$0.82} & 88.66\tiny{$\pm$0.60} & 58.50\tiny{$\pm$2.31} & 75.75\tiny{$\pm$0.25}\\
    \textsl{Vanilla} & 78.37\tiny{$\pm$0.72} & 90.50\tiny{$\pm$1.00} & 53.49\tiny{$\pm$0.15} & 85.13\tiny{$\pm$0.12} & 49.55\tiny{$\pm$0.99} & 74.70\tiny{$\pm$0.53}\\
    \textsl{DELTA} & \textbf{85.92\tiny{$\pm$0.13}} & \textbf{92.97\tiny{$\pm$0.21}} & \textbf{76.40\tiny{$\pm$0.54}} & \textbf{92.47\tiny{$\pm$0.39}} & \textbf{68.23\tiny{$\pm$0.19}} & \textbf{77.30\tiny{$\pm$0.15}}\\
    \bottomrule
    \end{tabular}
    \end{adjustbox}
    \end{table}

\subsection{Results under Coexisting Weak-Annotation Types}
\label{exp:mixed_weak}

The main experiments evaluate each weak-annotation regime separately.
To further verify that \textsl{DELTA} operates as a unified framework, we construct a CIFAR-10 setting in which all three mechanisms coexist: the training set is partitioned into four disjoint subsets with 20\% correctly labeled, 20\% wrongly labeled (uniformly flipped), 30\% ambiguously labeled (random policy), and the remaining 30\% unlabeled samples.
\textsl{DELTA} requires no algorithmic modification in this setting: the rectified target in Eq.~(\ref{eq:cls_unified}) is defined per sample according to the form of its annotation---a single label activates the noisy branch, a candidate set the ambiguous branch, and an empty annotation the semi-supervised branch.
Table~\ref{tab:mixed_weak} shows that \textsl{DELTA} remains stable when the three mechanisms coexist, substantially outperforming \textsl{Vanilla} on all generative metrics and in test accuracy.

\begin{table}[!h]
\centering
\AppendixTableSetup
\caption{Results on CIFAR-10 where noisy, ambiguous, and missing labels coexist.}
\label{tab:mixed_weak}
\begin{adjustbox}{max width=\linewidth,center}
\begin{tabular}{l|ccccccc|c}
\toprule
Method & FID$\downarrow$ & IS$\uparrow$ & Dens.$\uparrow$ & Cov.$\uparrow$
       & CW-FID$\downarrow$ & CW-Dens.$\uparrow$ & CW-Cov.$\uparrow$ & Test Acc.\ (\%)$\uparrow$ \\
\midrule
\textsl{Vanilla} & 10.92 & 9.05 & 97.64 & 69.83 & 30.17 & 80.46 & 68.52 & 61.24 \\
\textsl{DELTA} & \textbf{3.71} & \textbf{9.55} & \textbf{105.83} & \textbf{79.86} & \textbf{16.24} & \textbf{103.92} & \textbf{77.95} & \textbf{85.81} \\
\bottomrule
\end{tabular}
\end{adjustbox}
\end{table}

\subsection{Results on a Domain-Realistic Medical Dataset}
\label{exp:ham10000}

To assess external validity beyond CIFAR-style benchmarks, we evaluate \textsl{DELTA} on HAM10000~\citep{tschandl2018ham10000}, a real medical dataset of dermatoscopic images with seven lesion types, corrupted with 20\% symmetric label noise.
As shown in Table~\ref{tab:ham10000}, \textsl{DELTA} retains its advantage in this real medical domain.
Notably, HAM10000 is heavily class-imbalanced (the largest class alone accounts for 67\% of the data): \textsl{Vanilla}'s accuracy stays at the majority-class level, whereas \textsl{DELTA} clearly exceeds it.

\begin{table}[!h]
\centering
\AppendixTableSetup
\caption{Results on HAM10000 with 20\% symmetric noisy labels.}
\label{tab:ham10000}
\begin{adjustbox}{max width=\linewidth,center}
\begin{tabular}{l|ccccccc|c}
\toprule
Method & FID$\downarrow$ & IS$\uparrow$ & Dens.$\uparrow$ & Cov.$\uparrow$
       & CW-FID$\downarrow$ & CW-Dens.$\uparrow$ & CW-Cov.$\uparrow$ & Test Acc.\ (\%)$\uparrow$ \\
\midrule
\textsl{Vanilla} & 48.16 & 3.25 & 73.94 & 44.51 & 91.38 & 62.75 & 40.16 & 69.87 \\
\textsl{DELTA} & \textbf{39.72} & \textbf{3.52} & \textbf{83.61} & \textbf{52.38} & \textbf{65.44} & \textbf{76.92} & \textbf{47.15} & \textbf{79.34} \\
\bottomrule
\end{tabular}
\end{adjustbox}
\end{table}

\subsection{Ablation Results of Median-Centered Timestep Sampling}
\label{exp:timestep_ablation}

A retrained variant without median-centered sampling is intractable, as posterior estimation would revert to full-timestep computation, which is prohibitively expensive in both time and memory (Table~\ref{tab:efficiency}).
We therefore ablate it at the inference stage, comparing it against the uniform timestep sampling of \citet{chen2024diffusion} on CIFAR-10 under matched step budgets and input noise levels $\sigma_t$.
Table~\ref{tab:timestep_ablation} reports the numbers underlying Figure~\ref{fig:timestep_selection}(b): the median-centered scheme consistently outperforms uniform sampling across all step budgets and noise levels, with the margin growing markedly at higher noise levels (e.g., 60.30\% vs.\ 18.93\% at $\sigma_t = 1.0$ with two steps).

\begin{table}[!h]
\centering
\AppendixTableSetup
\setlength{\tabcolsep}{3pt}%
\begin{minipage}[t]{0.6\linewidth}
\centering
\caption{Classification accuracy (\%) of median-centered versus uniform timestep sampling on CIFAR-10.}
\label{tab:timestep_ablation}
\begin{adjustbox}{max width=\linewidth,center}
\begin{tabular}{c|l|ccccc}
\toprule
\multirow{2}{*}{Noise level} & \multirow{2}{*}{Method} & \multicolumn{5}{c}{Sampling steps} \\
\cmidrule(lr){3-7}
& & 2 & 4 & 8 & 16 & 32 \\
\midrule
\multirow{2}{*}{$\sigma_t \approx 0$} & Median-centered (ours) & \textbf{74.57} & \textbf{87.07} & \textbf{90.70} & \textbf{93.50} & \textbf{95.20} \\
& Uniform~\citep{chen2024diffusion} & 69.90 & 85.40 & 90.57 & 91.97 & 93.33 \\
\midrule
\multirow{2}{*}{$\sigma_t = 0.5$} & Median-centered (ours) & \textbf{75.53} & \textbf{82.57} & \textbf{86.53} & \textbf{86.87} & \textbf{87.10} \\
& Uniform~\citep{chen2024diffusion} & 42.83 & 61.27 & 68.70 & 70.67 & 71.70 \\
\midrule
\multirow{2}{*}{$\sigma_t = 1.0$} & Median-centered (ours) & \textbf{60.30} & \textbf{67.70} & \textbf{71.43} & \textbf{72.80} & \textbf{74.27} \\
& Uniform~\citep{chen2024diffusion} & 18.93 & 29.57 & 35.60 & 38.63 & 39.83 \\
\bottomrule
\end{tabular}
\end{adjustbox}
\end{minipage}
\hfill
\begin{minipage}[t]{0.36\linewidth}
\centering
\caption{Accuracy (\%) under different interval widths $\Delta$ ($\sigma_t = 0.5$; CIFAR-10, Sym-40\%).}
\label{tab:delta_sensitivity}
\begin{adjustbox}{max width=\linewidth,center}
\begin{tabular}{c|ccccc}
\toprule
\multirow{2}{*}{$\Delta$} & \multicolumn{5}{c}{Sampling steps} \\
\cmidrule(lr){2-6}
& 2 & 4 & 8 & 16 & 32 \\
\midrule
4 & 74.86 & 81.92 & 85.71 & 86.03 & 86.24 \\
5 & 75.31 & 82.14 & 86.22 & 86.55 & 86.79 \\
6.4 & \textbf{75.53} & \textbf{82.57} & \textbf{86.53} & \textbf{86.87} & \textbf{87.10} \\
7 & 75.42 & 82.39 & 86.44 & 86.71 & 87.02 \\
8 & 75.18 & 82.08 & 86.12 & 86.49 & 86.71 \\
\bottomrule
\end{tabular}
\end{adjustbox}
\end{minipage}
\end{table}

\subsection{Sensitivity Analysis of the Sampling Interval Width}
\label{exp:delta_sensitivity}

We further examine the sensitivity of the median-centered timestep sampling to the interval width $\Delta$.
Fixing $\sigma_t = 0.5$, we sweep $\Delta \in \{4, 5, 6.4, 7, 8\}$ around the default $\Delta = 6.4$ (Appendix~\ref{exp:model_setup}) and report the classification accuracy across sampling steps on CIFAR-10 with 40\% symmetric label noise.
As shown in Table~\ref{tab:delta_sensitivity}, the accuracy varies by less than 0.9 percentage points at every step budget, and even the least favorable width still outperforms uniform sampling by a wide margin (Table~\ref{tab:timestep_ablation}), confirming that the efficiency gains are robust to the choice of $\Delta$.

\clearpage
\section{Weakly Supervised Dataset Condensation}
\label{sec:condensation}
This section develops the weakly supervised dataset condensation application discussed in the main paper.
The goal is to distill large weakly annotated datasets into compact synthetic counterparts for efficient downstream training.
In this setting, a trained \textsl{DELTA} model acts as a generative compressor that synthesizes class-representative samples from weakly annotated data.
This differs from conventional condensation methods, which usually assume clean labels, and from recent noise-robust extensions such as \textsl{RDC}~\citep{kim2025robust}, which are tailored to a single weak form.

\subsection{Setup}
\label{exp:cond_setup}
During condensation, we use the trained LCDM to synthesize images according to the specified images per class (IPC).
For evaluation, we compare against random selection (\textsl{RandSel}), hard-label-based methods (\textsl{DC}~\citep{zhao2021dataset}, \textsl{DM}~\citep{zhao2023dataset}, \textsl{DSA}~\citep{zhao2021datasetDSA}), soft-label-based methods (\textsl{RDED}~\citep{sun2024diversity}, \textsl{SRe2L}~\citep{yin2024squeeze}), and the noise-robust baseline \textsl{RDC}~\citep{kim2025robust}.
Following standard protocols~\citep{yin2024squeeze}, we adopt ResNet-18 during condensation and evaluate the condensed datasets on a held-out test set using ResNet-34.

\subsection{Representative Results}
\label{exp:cond_main}
Tables~\ref{tab:dd_results2},~\ref{tab:dd_results_ssl}, and~\ref{tab:pl_results} report the classification accuracy on condensed datasets across the three weak-annotation scenarios at multiple IPC settings.
In Table~\ref{tab:dd_results_ssl}, ``-'' indicates that the number of available labeled samples per class is already below the target IPC, so no additional condensation experiment is conducted for that setting.
Standard condensation methods degrade markedly under heavy label noise.
In contrast, \textsl{DELTA} attains competitive or superior accuracy compared to the noise-robust baseline \textsl{RDC}, and consistently outperforms \textsl{RDED} and \textsl{SRe2L} in ambiguous-label and semi-supervised scenarios.
Qualitative comparisons in Figure~\ref{fig:condensed_pll_cifar10} indicate that our method generates samples with clearer class semantics.
Comprehensive IPC analyses and additional visualizations are provided in the following two subsections.

\subsection{Comprehensive IPC Analyses}
\label{exp:cond_full}
We evaluate \textsl{DELTA} on the weakly supervised dataset condensation task across three weak-annotation scenarios with extended IPC settings: \textbf{noisy labels} (Table~\ref{tab:dd_results2}), \textbf{semi-supervised data} (Table~\ref{tab:dd_results_ssl}), and \textbf{ambiguous labels} (Table~\ref{tab:pl_results}).
First, under {label noise}, $\textsl{DELTA}$ almost achieves the best performance across most datasets and noise types, effectively suppressing noise accumulation in the condensed set.
Second, in the {semi-supervised} setting, $\textsl{DELTA}$ exhibits strong data efficiency.
Finally, regarding {ambiguous labels}, our method demonstrates superior disambiguation capabilities. Unlike decoupled methods (e.g., \textsl{RDED}, \textsl{SRe2L}) that struggle with candidate sets, $\textsl{DELTA}$ achieves remarkable accuracy gains.
Across scenarios, performance trends are also stable as IPC increases: \textsl{DELTA} remains competitive in low-IPC regimes where information is severely limited, while preserving gains at higher IPC where optimization instability and label corruption can otherwise accumulate.
The improvements are particularly consistent on more challenging datasets and corruption settings, suggesting that the learned generator captures class-level semantics rather than memorizing noisy supervision.
Compared with hard-label condensation baselines, this behavior indicates stronger robustness to supervision mismatch; compared with soft-label decoupling strategies, it reflects better end-to-end alignment between semantic generation and class assignment.
These results collectively highlight that our generative strategy can effectively recover clean distributions and generate compact, reliable synthetic datasets under diverse weak-annotation conditions.

\begin{table}[!t]
\centering
\AppendixTableSetup
\caption{Test accuracy (\%) on condensed datasets from the noisy-label datasets.}
\label{tab:dd_results2}
\begin{adjustbox}{max width=\linewidth,center}
\begin{tabular}{c|c|c|cccccccc}
\toprule
Dataset & Type & IPC & \textsl{RandSel} & \textsl{DC} & \textsl{DSA} & \textsl{DM} & \textsl{RDED} & \textsl{SRe2L} & \textsl{RDC} & \textsl{DELTA} \\
\midrule
\multirow{6}{*}{\shortstack{Fashion-\\MNIST}} & \multirow{3}{*}{\scriptsize{{Sym-40\%}}}
& {10} & 34.42\tiny{$\pm$0.69} & 22.85\tiny{$\pm$1.69} & 42.07\tiny{$\pm$2.49} & 57.06\tiny{$\pm$1.52} & 18.57\tiny{$\pm$1.06} & 15.80\tiny{$\pm$0.38} & 68.88\tiny{$\pm$1.16} & \textbf{70.18\tiny{$\pm$0.37}} \\
& & {50} & 52.36\tiny{$\pm$0.60} & 35.64\tiny{$\pm$2.26} & 55.22\tiny{$\pm$1.51} & 68.23\tiny{$\pm$0.47} & 23.19\tiny{$\pm$0.74} & 19.51\tiny{$\pm$0.96} & 72.39\tiny{$\pm$1.35} & \textbf{80.73\tiny{$\pm$0.07}} \\
& & {100} & 55.14\tiny{$\pm$0.06} & 30.46\tiny{$\pm$1.74} & 41.30\tiny{$\pm$0.85} & 73.21\tiny{$\pm$0.69} & 25.43\tiny{$\pm$0.21} & 19.66\tiny{$\pm$1.91} & 81.91\tiny{$\pm$0.54} & \textbf{84.26\tiny{$\pm$0.02}} \\
\cmidrule{2-11}
& \multirow{3}{*}{\scriptsize{{Asym-40\%}}}
& {10} & 48.28\tiny{$\pm$0.34} & 53.17\tiny{$\pm$1.59} & 57.15\tiny{$\pm$2.37} & 63.27\tiny{$\pm$1.60} & 18.42\tiny{$\pm$1.62} & 16.45\tiny{$\pm$1.96} & \textbf{65.17\tiny{$\pm$3.80}} & {65.02\tiny{$\pm$1.85}} \\
& & {50} & 69.44\tiny{$\pm$0.17} & 49.21\tiny{$\pm$0.69} & 77.20\tiny{$\pm$0.34} & 76.39\tiny{$\pm$0.57} & 22.31\tiny{$\pm$0.67} & 27.07\tiny{$\pm$0.35} & 73.11\tiny{$\pm$1.48} & \textbf{79.65\tiny{$\pm$0.63}} \\
& & {100} & 70.80\tiny{$\pm$0.91} & 36.95\tiny{$\pm$0.57} & 80.24\tiny{$\pm$0.54} & 78.43\tiny{$\pm$0.63} & 24.03\tiny{$\pm$0.97} & 26.52\tiny{$\pm$1.46} & 81.59\tiny{$\pm$1.29} & \textbf{83.22\tiny{$\pm$0.33}} \\
\midrule
\multirow{6}{*}{{CIFAR-10}} & \multirow{3}{*}{\scriptsize{{Sym-40\%}}}
&  {10} & 16.30\tiny{$\pm$0.96} & 18.11\tiny{$\pm$1.02} & 18.06\tiny{$\pm$1.72} & 23.71\tiny{$\pm$0.40} & 19.85\tiny{$\pm$0.88} & 13.12\tiny{$\pm$1.04} & 25.88\tiny{$\pm$1.16} & \textbf{27.83\tiny{$\pm$0.98}} \\
& & {50} & 26.59\tiny{$\pm$0.70} & 20.63\tiny{$\pm$0.22} & 28.76\tiny{$\pm$0.57} & 29.50\tiny{$\pm$0.56} & 34.64\tiny{$\pm$0.58} & 14.23\tiny{$\pm$1.67} & 42.39\tiny{$\pm$1.17} & \textbf{46.47\tiny{$\pm$0.41}} \\
& & {100} & 31.19\tiny{$\pm$0.74} & 19.91\tiny{$\pm$0.54} & 29.45\tiny{$\pm$0.34} & 32.26\tiny{$\pm$0.75} & 44.03\tiny{$\pm$0.21} & 14.21\tiny{$\pm$0.93} & 51.81\tiny{$\pm$0.93} & \textbf{56.53\tiny{$\pm$0.03}} \\
\cmidrule{2-11}
& \multirow{3}{*}{\scriptsize{{Asym-40\%}}}
& {10} & 24.89\tiny{$\pm$1.65} & 18.51\tiny{$\pm$1.35} & 22.23\tiny{$\pm$1.80} & 26.53\tiny{$\pm$0.07} & 23.48\tiny{$\pm$0.65} & 14.64\tiny{$\pm$1.03} & \textbf{27.39\tiny{$\pm$2.92}} & {24.94\tiny{$\pm$0.49}} \\
& & {50} & 40.95\tiny{$\pm$0.59} & 25.97\tiny{$\pm$0.97} & 40.81\tiny{$\pm$0.29} & 43.09\tiny{$\pm$0.76} & 39.12\tiny{$\pm$0.13} & 16.03\tiny{$\pm$0.21} & 44.41\tiny{$\pm$1.26} & \textbf{47.77\tiny{$\pm$0.78}} \\
& & {100} & 47.49\tiny{$\pm$0.64} & 27.76\tiny{$\pm$0.72} & 42.96\tiny{$\pm$0.84} & 51.61\tiny{$\pm$0.60} & 44.45\tiny{$\pm$0.19} & 17.55\tiny{$\pm$0.91} & 53.81\tiny{$\pm$0.93} & \textbf{55.89\tiny{$\pm$0.39}} \\
\midrule
\multirow{6}{*}{{ImageNette}} & \multirow{3}{*}{\scriptsize{{Sym-40\%}}}
& {10} & 23.09\tiny{$\pm$0.19} & 15.89\tiny{$\pm$0.73} & 27.70\tiny{$\pm$1.25} & 28.83\tiny{$\pm$0.73} & 21.15\tiny{$\pm$1.05} & 25.03\tiny{$\pm$1.17} & 33.23\tiny{$\pm$1.07} & \textbf{34.36\tiny{$\pm$1.05}} \\
& & {50} & 33.83\tiny{$\pm$0.28} & 24.62\tiny{$\pm$0.73} & 32.07\tiny{$\pm$1.01} & 42.66\tiny{$\pm$1.27} & 35.87\tiny{$\pm$0.39} & 35.37\tiny{$\pm$0.82} & 42.88\tiny{$\pm$0.78} & \textbf{44.93\tiny{$\pm$0.28}} \\
& & {100} & 40.04\tiny{$\pm$0.71} & 22.81\tiny{$\pm$1.22} & 36.05\tiny{$\pm$1.76} & 43.25\tiny{$\pm$2.13} & 35.87\tiny{$\pm$0.39} & 41.74\tiny{$\pm$1.37} & 49.76\tiny{$\pm$1.56} & \textbf{56.23\tiny{$\pm$0.84}} \\
\cmidrule{2-11}
& \multirow{3}{*}{\scriptsize{{Asym-40\%}}} & {10} & 26.54\tiny{$\pm$0.88} & 19.26\tiny{$\pm$0.98} & 30.62\tiny{$\pm$2.09} & 33.40\tiny{$\pm$0.48} & 26.23\tiny{$\pm$0.06} & 25.74\tiny{$\pm$2.21} & 35.46\tiny{$\pm$1.54} & \textbf{37.09\tiny{$\pm$0.29}} \\
& & {50} & 47.91\tiny{$\pm$0.61} & 31.68\tiny{$\pm$2.15} & 43.41\tiny{$\pm$1.24} & 50.97\tiny{$\pm$1.61} & 32.75\tiny{$\pm$0.43} & 35.29\tiny{$\pm$0.14} & 53.37\tiny{$\pm$1.40} & \textbf{55.20\tiny{$\pm$0.46}} \\
& & {100} & 59.10\tiny{$\pm$1.41} & 29.19\tiny{$\pm$0.21} & 53.79\tiny{$\pm$0.84} & 60.70\tiny{$\pm$1.88} & 35.48\tiny{$\pm$0.22} & 42.37\tiny{$\pm$0.34} & 67.56\tiny{$\pm$0.52} & \textbf{68.97\tiny{$\pm$0.12}} \\
\bottomrule
\end{tabular}
\end{adjustbox}
\end{table}

\begin{table}[!t]
\centering
\AppendixTableSetup
\caption{Test accuracy (\%) on condensed datasets from the semi-supervised annotations.}
\label{tab:dd_results_ssl}
\begin{adjustbox}{max width=\linewidth,center}
\begin{tabular}{c|c|cccccccc}
\toprule
Dataset & Type & IPC & \textsl{RandSel} & \textsl{DC} & \textsl{DSA} & \textsl{DM} & \textsl{RDED} & \textsl{SRe2L} & \textsl{DELTA} \\
\midrule
\multirow{7}{*}{Fashion-MNIST} & \multirow{3}{*}{\scriptsize{{1\% Labels}}}
& {10} & 64.71\tiny{$\pm$1.30} & 66.32\tiny{$\pm$1.97} & 67.08\tiny{$\pm$1.90} & 68.32\tiny{$\pm$1.70} & 66.63\tiny{$\pm$1.52} & 63.12\tiny{$\pm$2.47} & \textbf{69.20\tiny{$\pm$0.35}} \\
& & {50} & 83.33\tiny{$\pm$0.37} & 83.02\tiny{$\pm$2.96} & 84.46\tiny{$\pm$0.64} & 83.60\tiny{$\pm$0.22} & 87.68\tiny{$\pm$3.30} & 85.82\tiny{$\pm$2.65} & \textbf{88.99\tiny{$\pm$0.39}} \\
& & {100} & - & - & - & - & - & - & \textbf{89.29\tiny{$\pm$0.23}} \\
\cmidrule{2-10}
& \multirow{3}{*}{\scriptsize{{10\% Labels}}}
& {10} & 63.22\tiny{$\pm$0.17} & 64.42\tiny{$\pm$3.60} & 64.27\tiny{$\pm$1.86} & 68.84\tiny{$\pm$3.30} & 67.92\tiny{$\pm$1.68} & 69.96\tiny{$\pm$2.49} & \textbf{70.25\tiny{$\pm$0.53}} \\
& & {50} & 80.92\tiny{$\pm$0.06} & 81.11\tiny{$\pm$0.63} & 81.31\tiny{$\pm$0.56} & 83.58\tiny{$\pm$0.31} & 83.51\tiny{$\pm$1.78} & 83.37\tiny{$\pm$0.26} & \textbf{83.85\tiny{$\pm$1.79}} \\
& & {100} & 83.85\tiny{$\pm$0.24} & 83.60\tiny{$\pm$0.76} & 83.18\tiny{$\pm$0.20} & 86.29\tiny{$\pm$0.38} & \textbf{89.74\tiny{$\pm$0.39}} & 87.03\tiny{$\pm$2.07} & 84.54\tiny{$\pm$0.22} \\
\midrule
\multirow{7}{*}{{CIFAR-10}} & \multirow{3}{*}{\scriptsize{{1\% Labels}}}
& {10} & 24.51\tiny{$\pm$0.34} & 23.26\tiny{$\pm$0.64} & 24.90\tiny{$\pm$1.22} & 28.73\tiny{$\pm$0.37} & 27.01\tiny{$\pm$9.92} & 31.87\tiny{$\pm$1.54} & \textbf{33.43\tiny{$\pm$0.62}} \\
& & {50} & - & - & - & - & - & - & \textbf{44.74\tiny{$\pm$0.67}} \\
& & {100} & - & - & - & - & - & - & \textbf{52.83\tiny{$\pm$1.09}} \\
\cmidrule{2-10}
& \multirow{3}{*}{\scriptsize{{10\% Labels}}}
& {10} & 23.17\tiny{$\pm$1.63} & 21.84\tiny{$\pm$1.37} & 24.90\tiny{$\pm$0.23} & 29.71\tiny{$\pm$1.69} & 32.38\tiny{$\pm$0.71} & 31.61\tiny{$\pm$0.65} & \textbf{33.53\tiny{$\pm$0.58}} \\
& & {50} & 41.51\tiny{$\pm$0.57} & 41.31\tiny{$\pm$0.93} & 44.20\tiny{$\pm$0.64} & 43.71\tiny{$\pm$1.58} &43.11\tiny{$\pm$1.18} & 42.44\tiny{$\pm$0.62} & \textbf{45.15\tiny{$\pm$0.16}} \\
& & {100} & 54.06\tiny{$\pm$0.67} & 55.58\tiny{$\pm$0.89} & 57.06\tiny{$\pm$1.04} & 57.48\tiny{$\pm$1.48} & 57.11\tiny{$\pm$0.60} & 56.36\tiny{$\pm$0.88} & \textbf{58.99\tiny{$\pm$0.95}} \\
\bottomrule
\end{tabular}
\end{adjustbox}
\end{table}

\begin{table}[!t]
\centering
\AppendixTableSetup
\caption{Test accuracy (\%) on condensed datasets from the ambiguous-label annotations.}
\label{tab:pl_results}
\begin{adjustbox}{max width=\linewidth,center}
\begin{tabular}{c|c|ccc|c|ccc}
\toprule
\multirow{2}{*}{Dataset} & \multicolumn{4}{c|}{\textbf{Random}} & \multicolumn{4}{c}{\textbf{Class-Dept.}} \\
\cmidrule{2-9}
& IPC & \textsl{RDED} & \textsl{SRe2L} & \textsl{DELTA}
& IPC & \textsl{RDED} & \textsl{SRe2L} & \textsl{DELTA} \\
\midrule
\multirow{3}{*}{Fashion-MNIST}
& 10  & 13.17\tiny{$\pm$4.66} &  9.93\tiny{$\pm$1.15} & \textbf{72.02\tiny{$\pm$0.77}}
& 10  & 13.53\tiny{$\pm$1.12} & 10.83\tiny{$\pm$0.35} & \textbf{70.46\tiny{$\pm$0.26}} \\
& 50  & 14.47\tiny{$\pm$1.06} & 10.30\tiny{$\pm$0.40} & \textbf{83.98\tiny{$\pm$0.12}}
& 50  & 15.05\tiny{$\pm$0.26} & 11.33\tiny{$\pm$0.76} & \textbf{79.67\tiny{$\pm$0.13}} \\
& 100 & 15.00\tiny{$\pm$0.18} & 10.48\tiny{$\pm$0.74} & \textbf{87.30\tiny{$\pm$0.31}}
& 100 & 16.91\tiny{$\pm$2.75} & 11.05\tiny{$\pm$0.77} & \textbf{81.42\tiny{$\pm$0.25}} \\
\bottomrule
\end{tabular}
\end{adjustbox}
\end{table}

\subsection{Visualization of Weakly Supervised Condensed Datasets}
\label{exp:visual}
\begin{figure}[!h]
\centering
\begin{minipage}[b]{0.16\textwidth}
\begin{tabular}{@{}c@{}c@{}}
    {\scriptsize
    \renewcommand{\arraystretch}{0.45}
    \begin{tabular}[b]{@{}c}
        {Airplane}\\ {Automobile}\\ {Bird}\\ {Cat}\\ {Deer}\\ {Dog}\\ {Frog}\\ {Horse}\\ {Ship}\\ {Truck}
    \end{tabular}} &
    \ \ \adjincludegraphics[valign=b,width=0.91\linewidth]{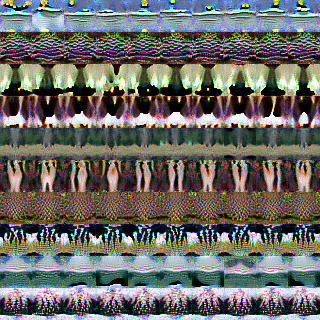}\\
\end{tabular}\\
\centering    
\hspace*{30pt}(f1) SRe2L\!\!\!\!\!\!\!\!\!\!\!\!\!\!\!\!\!\!\!\!\!\!\!\!\!\!\!
\end{minipage}
\hspace{50pt}   
\begin{minipage}[b]{0.16\textwidth}
    \centering
    \includegraphics[width=0.895\linewidth]{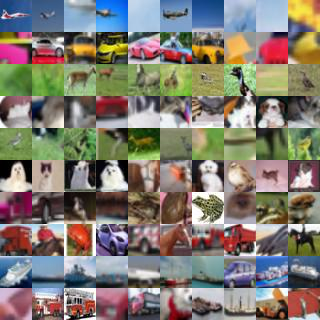}\\
    (f2) RDED
\end{minipage}
\hspace{8pt} 
\begin{minipage}[b]{0.16\textwidth}
    \centering
    \includegraphics[width=0.895\linewidth]{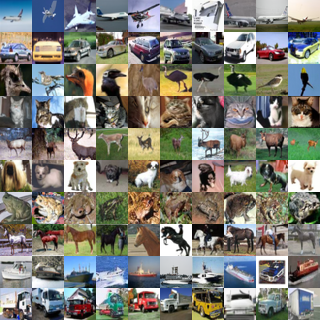}\\
    (f3) \textbf{DELTA}
\end{minipage}
\hspace{180pt} 
\caption{Visualization of condensed images (IPC=10) synthesized by different methods under various weak annotations. 
Rows (a) and (b) show results under {noisy labels} (Sym-40\%) for Fashion-MNIST and CIFAR-10, respectively. 
Rows (c) and (d) show results under {semi-supervised} data (1\% Labels) for Fashion-MNIST and CIFAR-10. 
Rows (e) and (f) display results under {ambiguous labels} (Random) for Fashion-MNIST and CIFAR-10. }
\label{fig:condensed_pll_cifar10}
\end{figure}

We visualize synthesized images on Fashion-MNIST and CIFAR-10 in Figure~\ref{fig:condensed_pll_cifar10}.
Ideally, a robust condensation method should filter out annotation corruption and generate clean, class-representative samples.
Standard optimization-based methods such as \textsl{RDED} and \textsl{SRe2L}, as well as related hard-label condensation baselines such as \textsl{DC}, \textsl{DSA}, and \textsl{DM}, tend to overfit corrupted annotations, producing semantic mismatches between synthesized images and assigned class labels.
In contrast, \textsl{DELTA} produces samples with clearer semantics and better class alignment.
These qualitative comparisons support the quantitative findings above and show that the generative framework can yield high-quality condensed datasets from imperfect annotations.

\clearpage

\begin{figure}[!h]
\centering
\begin{minipage}[b]{0.16\textwidth}
\begin{tabular}{@{}c@{}c@{}}
    {\scriptsize
    \renewcommand{\arraystretch}{0.45}
    \begin{tabular}[b]{@{}c}
        {T-shirt}\\ {Trouser}\\ {Pullover}\\ {Dress}\\ {Coat}\\ {Sandal}\\ {Shirt}\\ {Sneaker}\\ {Bag}\\ {Ankle boot}
    \end{tabular}} &
    \ \ \adjincludegraphics[valign=b,width=0.91\linewidth]{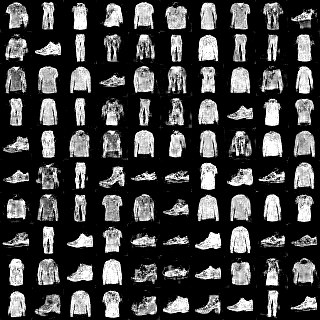}\\
\end{tabular}\\
\centering    
\hspace*{30pt}(a1) DC\!\!\!\!\!\!\!\!\!\!\!\!\!\!\!\!\!\!\!\!\!\!\!\!\!\!\!
\end{minipage}
\hspace{50pt}   
\begin{minipage}[b]{0.16\textwidth}
    \centering
    \includegraphics[width=0.895\linewidth]{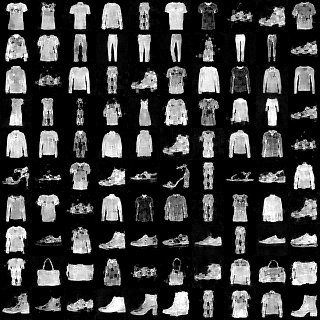}\\
    (a2) DM
\end{minipage}
\hfill
\begin{minipage}[b]{0.16\textwidth}
    \centering
    \includegraphics[width=0.895\linewidth]{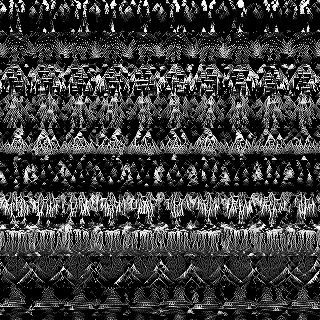}\\
    (a3) SRe2L
\end{minipage}
\hfill
\begin{minipage}[b]{0.16\textwidth}
    \centering
    \includegraphics[width=0.895\linewidth]{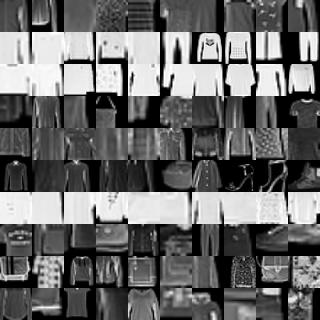}\\
    (a4) RDED
\end{minipage}
\hfill
\begin{minipage}[b]{0.16\textwidth}
    \centering
    \includegraphics[width=0.895\linewidth]{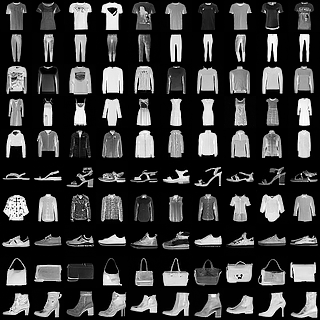}\\
    (a5) \textbf{DELTA}
\end{minipage}
\vspace{-10pt}
\end{figure}

\begin{figure}[!h]
\centering
\begin{minipage}[b]{0.16\textwidth}
\begin{tabular}{@{}c@{}c@{}}
        {\scriptsize
        \renewcommand{\arraystretch}{0.45}
        \begin{tabular}[b]{@{}c}
            {Airplane}\\ {Automobile}\\ {Bird}\\ {Cat}\\ {Deer}\\ {Dog}\\ {Frog}\\ {Horse}\\ {Ship}\\ {Truck}
        \end{tabular}} &
        \ \ \adjincludegraphics[valign=b,width=0.91\linewidth]{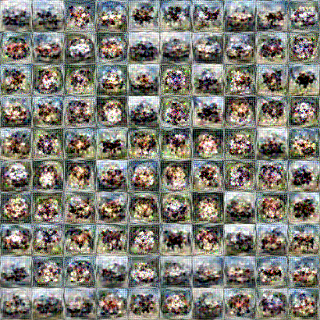}\\
\end{tabular}\\
\centering    
\hspace*{30pt}(b1) DC\!\!\!\!\!\!\!\!\!\!\!\!\!\!\!\!\!\!\!\!\!\!\!\!\!\!\!
\end{minipage}
\hspace{50pt}   
\begin{minipage}[b]{0.16\textwidth}
    \centering
    \includegraphics[width=0.895\linewidth]{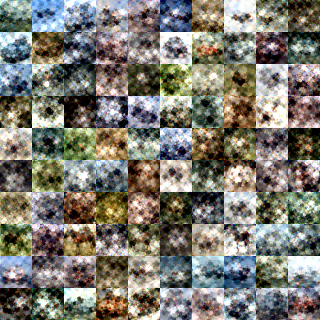}\\
    (b2) DM
\end{minipage}
\hfill
\begin{minipage}[b]{0.16\textwidth}
    \centering
    \includegraphics[width=0.895\linewidth]{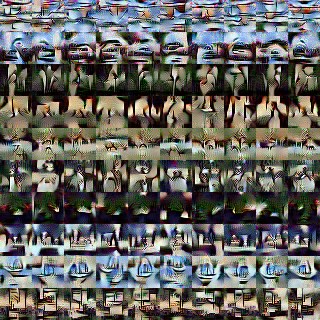}\\
    (b3) SRe2L
\end{minipage}
\hfill
\begin{minipage}[b]{0.16\textwidth}
    \centering
    \includegraphics[width=0.895\linewidth]{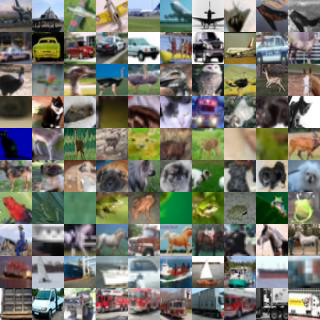}\\
    (b4) RDED
\end{minipage}
\hfill
\begin{minipage}[b]{0.16\textwidth}
    \centering
    \includegraphics[width=0.895\linewidth]{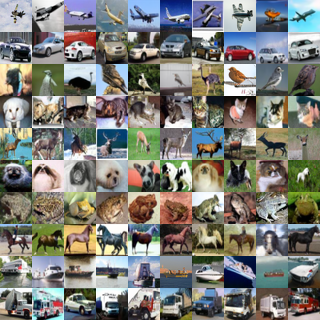}\\
    (b5) \textbf{DELTA}
\end{minipage}
\vspace{-10pt}
\end{figure}

\begin{figure}[!h]
\centering
\begin{minipage}[b]{0.16\textwidth}
\begin{tabular}{@{}c@{}c@{}}
    {\scriptsize
    \renewcommand{\arraystretch}{0.45}
    \begin{tabular}[b]{@{}c}
        {T-shirt}\\ {Trouser}\\ {Pullover}\\ {Dress}\\ {Coat}\\ {Sandal}\\ {Shirt}\\ {Sneaker}\\ {Bag}\\ {Ankle boot}
    \end{tabular}} &
    \ \ \adjincludegraphics[valign=b,width=0.91\linewidth]{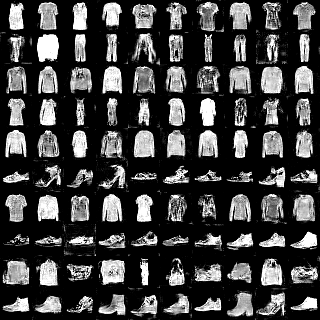}\\
\end{tabular}\\
\centering    
\hspace*{30pt}(c1) DC\!\!\!\!\!\!\!\!\!\!\!\!\!\!\!\!\!\!\!\!\!\!\!\!\!\!\!
\end{minipage}
\hspace{50pt}   
\begin{minipage}[b]{0.16\textwidth}
    \centering
    \includegraphics[width=0.895\linewidth]{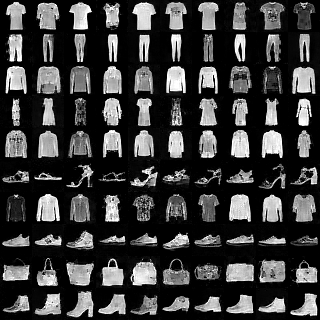}\\
    (c2) DM
\end{minipage}
\hfill
\begin{minipage}[b]{0.16\textwidth}
    \centering
    \includegraphics[width=0.895\linewidth]{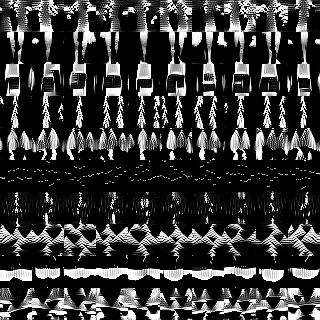}\\
    (c3) SRe2L
\end{minipage}
\hfill
\begin{minipage}[b]{0.16\textwidth}
    \centering
    \includegraphics[width=0.895\linewidth]{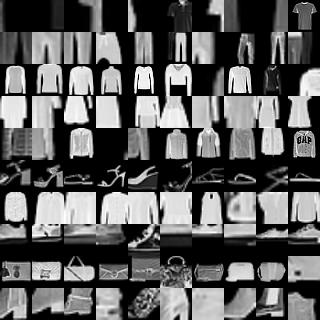}\\
    (c4) RDED
\end{minipage}
\hfill
\begin{minipage}[b]{0.16\textwidth}
    \centering
    \includegraphics[width=0.895\linewidth]{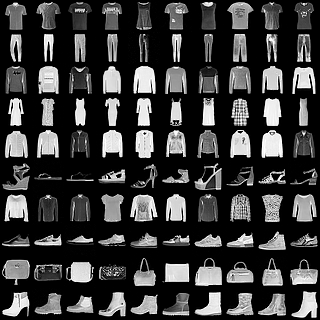}\\
    (c5) \textbf{DELTA}
\end{minipage}
\vspace{-10pt}
\end{figure}

\begin{figure}[!h]
\centering
\begin{minipage}[b]{0.16\textwidth}
\begin{tabular}{@{}c@{}c@{}}
        {\scriptsize
        \renewcommand{\arraystretch}{0.45}
        \begin{tabular}[b]{@{}c}
            {Airplane}\\ {Automobile}\\ {Bird}\\ {Cat}\\ {Deer}\\ {Dog}\\ {Frog}\\ {Horse}\\ {Ship}\\ {Truck}
        \end{tabular}} &
        \ \ \adjincludegraphics[valign=b,width=0.91\linewidth]{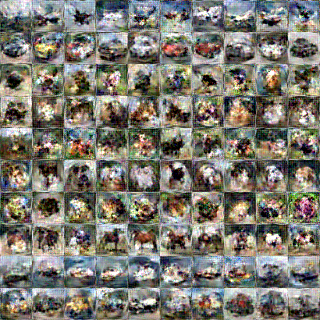}\\
\end{tabular}\\
\centering    
\hspace*{30pt}(d1) DC\!\!\!\!\!\!\!\!\!\!\!\!\!\!\!\!\!\!\!\!\!\!\!\!\!\!\!
\end{minipage}
\hspace{50pt}   
\begin{minipage}[b]{0.16\textwidth}
    \centering
    \includegraphics[width=0.895\linewidth]{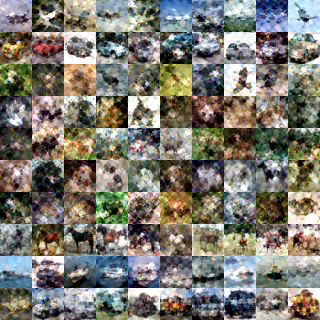}\\
    (d2) DM
\end{minipage}
\hfill
\begin{minipage}[b]{0.16\textwidth}
    \centering
    \includegraphics[width=0.895\linewidth]{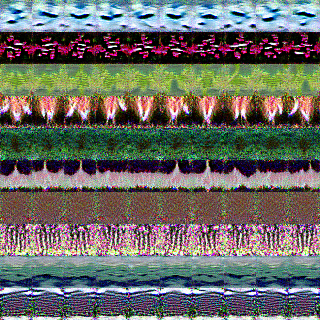}\\
    (d3) SRe2L
\end{minipage}
\hfill
\begin{minipage}[b]{0.16\textwidth}
    \centering
    \includegraphics[width=0.895\linewidth]{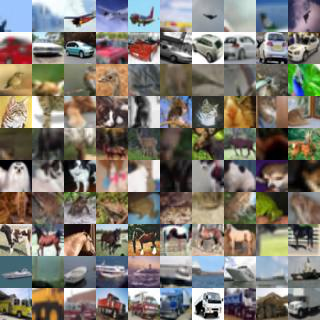}\\
    (d4) RDED
\end{minipage}
\hfill
\begin{minipage}[b]{0.16\textwidth}
    \centering
    \includegraphics[width=0.895\linewidth]{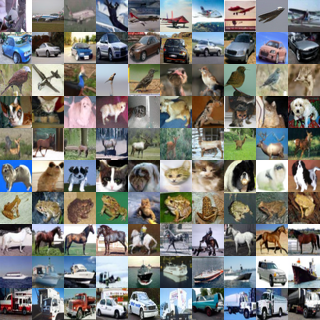}\\
    (d5) \textbf{DELTA}
\end{minipage}
\vspace{-10pt}
\end{figure}

\begin{figure}[!h]
\centering
\begin{minipage}[b]{0.16\textwidth}
\begin{tabular}{@{}c@{}c@{}}
    {\scriptsize
    \renewcommand{\arraystretch}{0.45}
    \begin{tabular}[b]{@{}c}
        {T-shirt}\\ {Trouser}\\ {Pullover}\\ {Dress}\\ {Coat}\\ {Sandal}\\ {Shirt}\\ {Sneaker}\\ {Bag}\\ {Ankle boot}
    \end{tabular}} &
    \ \ \adjincludegraphics[valign=b,width=0.91\linewidth]{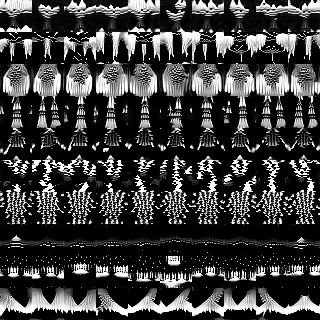}\\
\end{tabular}\\
\centering    
\hspace*{30pt}(e1) SRe2L\!\!\!\!\!\!\!\!\!\!\!\!\!\!\!\!\!\!\!\!\!\!\!\!\!\!\!
\end{minipage}
\hspace{50pt}   
\begin{minipage}[b]{0.16\textwidth}
    \centering
    \includegraphics[width=0.895\linewidth]{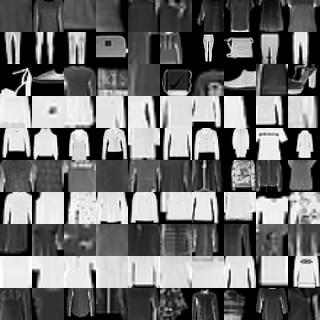}\\
    (e2) RDED
\end{minipage}
\hspace{8pt} 
\begin{minipage}[b]{0.16\textwidth}
    \centering
    \includegraphics[width=0.895\linewidth]{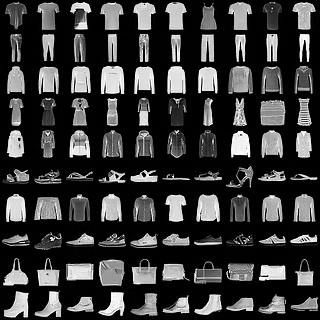}\\
    (e3) \textbf{DELTA}
\end{minipage}
\hspace{180pt} 
\vspace{-10pt}
\end{figure}

\clearpage
\section*{NeurIPS Paper Checklist}

\begin{enumerate}

\item {\bf Claims}
    \item[] Question: Do the main claims made in the abstract and introduction accurately reflect the paper's contributions and scope?
    \item[] Answer: \answerYes{} %
    \item[] Justification: The abstract and Section 1 claim (i) a unified framework handling all three forms of weak annotations with less prior knowledge, (ii) a median-centered timestep sampling strategy, and (iii) consistent improvements over specialized baselines. All three are substantiated in Sections 4 and 5.
    \item[] Guidelines:
    \begin{itemize}
        \item The answer \answerNA{} means that the abstract and introduction do not include the claims made in the paper.
        \item The abstract and/or introduction should clearly state the claims made, including the contributions made in the paper and important assumptions and limitations. A \answerNo{} or \answerNA{} answer to this question will not be perceived well by the reviewers. 
        \item The claims made should match theoretical and experimental results, and reflect how much the results can be expected to generalize to other settings. 
        \item It is fine to include aspirational goals as motivation as long as it is clear that these goals are not attained by the paper. 
    \end{itemize}

\item {\bf Limitations}
    \item[] Question: Does the paper discuss the limitations of the work performed by the authors?
    \item[] Answer: \answerYes{} %
    \item[] Justification: Section 6 discusses two limitations.
    \item[] Guidelines:
    \begin{itemize}
        \item The answer \answerNA{} means that the paper has no limitation while the answer \answerNo{} means that the paper has limitations, but those are not discussed in the paper. 
        \item The authors are encouraged to create a separate ``Limitations'' section in their paper.
        \item The paper should point out any strong assumptions and how robust the results are to violations of these assumptions (e.g., independence assumptions, noiseless settings, model well-specification, asymptotic approximations only holding locally). The authors should reflect on how these assumptions might be violated in practice and what the implications would be.
        \item The authors should reflect on the scope of the claims made, e.g., if the approach was only tested on a few datasets or with a few runs. In general, empirical results often depend on implicit assumptions, which should be articulated.
        \item The authors should reflect on the factors that influence the performance of the approach. For example, a facial recognition algorithm may perform poorly when image resolution is low or images are taken in low lighting. Or a speech-to-text system might not be used reliably to provide closed captions for online lectures because it fails to handle technical jargon.
        \item The authors should discuss the computational efficiency of the proposed algorithms and how they scale with dataset size.
        \item If applicable, the authors should discuss possible limitations of their approach to address problems of privacy and fairness.
        \item While the authors might fear that complete honesty about limitations might be used by reviewers as grounds for rejection, a worse outcome might be that reviewers discover limitations that aren't acknowledged in the paper. The authors should use their best judgment and recognize that individual actions in favor of transparency play an important role in developing norms that preserve the integrity of the community. Reviewers will be specifically instructed to not penalize honesty concerning limitations.
    \end{itemize}

\item {\bf Theory assumptions and proofs}
    \item[] Question: For each theoretical result, does the paper provide the full set of assumptions and a complete (and correct) proof?
    \item[] Answer: \answerYes{} %
    \item[] Justification: All theoretical results (Theorem 4.1, Proposition 4.2, Theorem 4.3, Theorem C.1) state their assumptions explicitly, and full proofs are provided in Appendices C.3, C.4, C.6, and C.7, respectively.
    \item[] Guidelines:
    \begin{itemize}
        \item The answer \answerNA{} means that the paper does not include theoretical results. 
        \item All the theorems, formulas, and proofs in the paper should be numbered and cross-referenced.
        \item All assumptions should be clearly stated or referenced in the statement of any theorems.
        \item The proofs can either appear in the main paper or the supplemental material, but if they appear in the supplemental material, the authors are encouraged to provide a short proof sketch to provide intuition. 
        \item Inversely, any informal proof provided in the core of the paper should be complemented by formal proofs provided in appendix or supplemental material.
        \item Theorems and Lemmas that the proof relies upon should be properly referenced. 
    \end{itemize}

    \item {\bf Experimental result reproducibility}
    \item[] Question: Does the paper fully disclose all the information needed to reproduce the main experimental results of the paper to the extent that it affects the main claims and/or conclusions of the paper (regardless of whether the code and data are provided or not)?
    \item[] Answer: \answerYes{} %
    \item[] Justification: Appendix B fully specifies the weak-annotation construction (B.1), network architecture, optimizer, learning rate, EMA decay, batch size, and total iterations (B.2), and evaluation metrics (B.3).
    \item[] Guidelines:
    \begin{itemize}
        \item The answer \answerNA{} means that the paper does not include experiments.
        \item If the paper includes experiments, a \answerNo{} answer to this question will not be perceived well by the reviewers: Making the paper reproducible is important, regardless of whether the code and data are provided or not.
        \item If the contribution is a dataset and\slash or model, the authors should describe the steps taken to make their results reproducible or verifiable. 
        \item Depending on the contribution, reproducibility can be accomplished in various ways. For example, if the contribution is a novel architecture, describing the architecture fully might suffice, or if the contribution is a specific model and empirical evaluation, it may be necessary to either make it possible for others to replicate the model with the same dataset, or provide access to the model. In general. releasing code and data is often one good way to accomplish this, but reproducibility can also be provided via detailed instructions for how to replicate the results, access to a hosted model (e.g., in the case of a large language model), releasing of a model checkpoint, or other means that are appropriate to the research performed.
        \item While NeurIPS does not require releasing code, the conference does require all submissions to provide some reasonable avenue for reproducibility, which may depend on the nature of the contribution. For example
        \begin{enumerate}
            \item If the contribution is primarily a new algorithm, the paper should make it clear how to reproduce that algorithm.
            \item If the contribution is primarily a new model architecture, the paper should describe the architecture clearly and fully.
            \item If the contribution is a new model (e.g., a large language model), then there should either be a way to access this model for reproducing the results or a way to reproduce the model (e.g., with an open-source dataset or instructions for how to construct the dataset).
            \item We recognize that reproducibility may be tricky in some cases, in which case authors are welcome to describe the particular way they provide for reproducibility. In the case of closed-source models, it may be that access to the model is limited in some way (e.g., to registered users), but it should be possible for other researchers to have some path to reproducing or verifying the results.
        \end{enumerate}
    \end{itemize}

\item {\bf Open access to data and code}
    \item[] Question: Does the paper provide open access to the data and code, with sufficient instructions to faithfully reproduce the main experimental results, as described in supplemental material?
    \item[] Answer: \answerYes{} %
    \item[] Justification: All datasets used are publicly available; anonymized source code with reproduction scripts is included in the supplementary material.
    \item[] Guidelines:
    \begin{itemize}
        \item The answer \answerNA{} means that paper does not include experiments requiring code.
        \item Please see the NeurIPS code and data submission guidelines (\url{https://neurips.cc/public/guides/CodeSubmissionPolicy}) for more details.
        \item While we encourage the release of code and data, we understand that this might not be possible, so \answerNo{} is an acceptable answer. Papers cannot be rejected simply for not including code, unless this is central to the contribution (e.g., for a new open-source benchmark).
        \item The instructions should contain the exact command and environment needed to run to reproduce the results. See the NeurIPS code and data submission guidelines (\url{https://neurips.cc/public/guides/CodeSubmissionPolicy}) for more details.
        \item The authors should provide instructions on data access and preparation, including how to access the raw data, preprocessed data, intermediate data, and generated data, etc.
        \item The authors should provide scripts to reproduce all experimental results for the new proposed method and baselines. If only a subset of experiments are reproducible, they should state which ones are omitted from the script and why.
        \item At submission time, to preserve anonymity, the authors should release anonymized versions (if applicable).
        \item Providing as much information as possible in supplemental material (appended to the paper) is recommended, but including URLs to data and code is permitted.
    \end{itemize}

\item {\bf Experimental setting/details}
    \item[] Question: Does the paper specify all the training and test details (e.g., data splits, hyperparameters, how they were chosen, type of optimizer) necessary to understand the results?
    \item[] Answer: \answerYes{} %
    \item[] Justification: Section 5 and Appendix B specify data splits, weak-annotation protocols, optimizer choice and hyperparameters, model architecture, and training schedule.
    \item[] Guidelines:
    \begin{itemize}
        \item The answer \answerNA{} means that the paper does not include experiments.
        \item The experimental setting should be presented in the core of the paper to a level of detail that is necessary to appreciate the results and make sense of them.
        \item The full details can be provided either with the code, in appendix, or as supplemental material.
    \end{itemize}

\item {\bf Experiment statistical significance}
    \item[] Question: Does the paper report error bars suitably and correctly defined or other appropriate information about the statistical significance of the experiments?
    \item[] Answer: \answerYes{} %
    \item[] Justification: Classification tables (Tables 5, 7, 8, 9) report mean $\pm$ standard deviation over multiple runs. Generation metrics in Tables 1–3 follow the standard FID/IS evaluation protocol on a fixed number of samples, consistent with prior work.
    \item[] Guidelines:
    \begin{itemize}
        \item The answer \answerNA{} means that the paper does not include experiments.
        \item The authors should answer \answerYes{} if the results are accompanied by error bars, confidence intervals, or statistical significance tests, at least for the experiments that support the main claims of the paper.
        \item The factors of variability that the error bars are capturing should be clearly stated (for example, train/test split, initialization, random drawing of some parameter, or overall run with given experimental conditions).
        \item The method for calculating the error bars should be explained (closed form formula, call to a library function, bootstrap, etc.)
        \item The assumptions made should be given (e.g., Normally distributed errors).
        \item It should be clear whether the error bar is the standard deviation or the standard error of the mean.
        \item It is OK to report 1-sigma error bars, but one should state it. The authors should preferably report a 2-sigma error bar than state that they have a 96\% CI, if the hypothesis of Normality of errors is not verified.
        \item For asymmetric distributions, the authors should be careful not to show in tables or figures symmetric error bars that would yield results that are out of range (e.g., negative error rates).
        \item If error bars are reported in tables or plots, the authors should explain in the text how they were calculated and reference the corresponding figures or tables in the text.
    \end{itemize}

\item {\bf Experiments compute resources}
    \item[] Question: For each experiment, does the paper provide sufficient information on the computer resources (type of compute workers, memory, time of execution) needed to reproduce the experiments?
    \item[] Answer: \answerYes{} %
    \item[] Justification: Appendix B states that all experiments were run on NVIDIA Tesla A100 GPUs with CUDA 12.4.
    \item[] Guidelines:
    \begin{itemize}
        \item The answer \answerNA{} means that the paper does not include experiments.
        \item The paper should indicate the type of compute workers CPU or GPU, internal cluster, or cloud provider, including relevant memory and storage.
        \item The paper should provide the amount of compute required for each of the individual experimental runs as well as estimate the total compute. 
        \item The paper should disclose whether the full research project required more compute than the experiments reported in the paper (e.g., preliminary or failed experiments that didn't make it into the paper). 
    \end{itemize}
    
\item {\bf Code of ethics}
    \item[] Question: Does the research conducted in the paper conform, in every respect, with the NeurIPS Code of Ethics \url{https://neurips.cc/public/EthicsGuidelines}?
    \item[] Answer: \answerYes{} %
    \item[] Justification: The research conforms with the NeurIPS Code of Ethics. It uses only publicly available benchmark datasets and involves no human subjects or sensitive data.
    \item[] Guidelines:
    \begin{itemize}
        \item The answer \answerNA{} means that the authors have not reviewed the NeurIPS Code of Ethics.
        \item If the authors answer \answerNo, they should explain the special circumstances that require a deviation from the Code of Ethics.
        \item The authors should make sure to preserve anonymity (e.g., if there is a special consideration due to laws or regulations in their jurisdiction).
    \end{itemize}

\item {\bf Broader impacts}
    \item[] Question: Does the paper discuss both potential positive societal impacts and negative societal impacts of the work performed?
    \item[] Answer: \answerYes{} %
    \item[] Justification: Our work improves the data-efficiency of generative models under imperfect supervision, which is broadly beneficial in domains such as medical imaging and remote sensing where clean labels are costly.
    \item[] Guidelines:
    \begin{itemize}
        \item The answer \answerNA{} means that there is no societal impact of the work performed.
        \item If the authors answer \answerNA{} or \answerNo, they should explain why their work has no societal impact or why the paper does not address societal impact.
        \item Examples of negative societal impacts include potential malicious or unintended uses (e.g., disinformation, generating fake profiles, surveillance), fairness considerations (e.g., deployment of technologies that could make decisions that unfairly impact specific groups), privacy considerations, and security considerations.
        \item The conference expects that many papers will be foundational research and not tied to particular applications, let alone deployments. However, if there is a direct path to any negative applications, the authors should point it out. For example, it is legitimate to point out that an improvement in the quality of generative models could be used to generate Deepfakes for disinformation. On the other hand, it is not needed to point out that a generic algorithm for optimizing neural networks could enable people to train models that generate Deepfakes faster.
        \item The authors should consider possible harms that could arise when the technology is being used as intended and functioning correctly, harms that could arise when the technology is being used as intended but gives incorrect results, and harms following from (intentional or unintentional) misuse of the technology.
        \item If there are negative societal impacts, the authors could also discuss possible mitigation strategies (e.g., gated release of models, providing defenses in addition to attacks, mechanisms for monitoring misuse, mechanisms to monitor how a system learns from feedback over time, improving the efficiency and accessibility of ML).
    \end{itemize}
    
\item {\bf Safeguards}
    \item[] Question: Does the paper describe safeguards that have been put in place for responsible release of data or models that have a high risk for misuse (e.g., pre-trained language models, image generators, or scraped datasets)?
    \item[] Answer: \answerNA{} %
    \item[] Justification: Our models do not pose elevated risks of misuse beyond those of comparable academic diffusion models.
    \item[] Guidelines:
    \begin{itemize}
        \item The answer \answerNA{} means that the paper poses no such risks.
        \item Released models that have a high risk for misuse or dual-use should be released with necessary safeguards to allow for controlled use of the model, for example by requiring that users adhere to usage guidelines or restrictions to access the model or implementing safety filters. 
        \item Datasets that have been scraped from the Internet could pose safety risks. The authors should describe how they avoided releasing unsafe images.
        \item We recognize that providing effective safeguards is challenging, and many papers do not require this, but we encourage authors to take this into account and make a best faith effort.
    \end{itemize}

\item {\bf Licenses for existing assets}
    \item[] Question: Are the creators or original owners of assets (e.g., code, data, models), used in the paper, properly credited and are the license and terms of use explicitly mentioned and properly respected?
    \item[] Answer: \answerYes{} %
    \item[] Justification: All datasets and baseline methods are properly cited.
    \item[] Guidelines:
    \begin{itemize}
        \item The answer \answerNA{} means that the paper does not use existing assets.
        \item The authors should cite the original paper that produced the code package or dataset.
        \item The authors should state which version of the asset is used and, if possible, include a URL.
        \item The name of the license (e.g., CC-BY 4.0) should be included for each asset.
        \item For scraped data from a particular source (e.g., website), the copyright and terms of service of that source should be provided.
        \item If assets are released, the license, copyright information, and terms of use in the package should be provided. For popular datasets, \url{paperswithcode.com/datasets} has curated licenses for some datasets. Their licensing guide can help determine the license of a dataset.
        \item For existing datasets that are re-packaged, both the original license and the license of the derived asset (if it has changed) should be provided.
        \item If this information is not available online, the authors are encouraged to reach out to the asset's creators.
    \end{itemize}

\item {\bf New assets}
    \item[] Question: Are new assets introduced in the paper well documented and is the documentation provided alongside the assets?
    \item[] Answer: \answerNA{} %
    \item[] Justification: The paper does not release new datasets or pretrained models.
    \item[] Guidelines:
    \begin{itemize}
        \item The answer \answerNA{} means that the paper does not release new assets.
        \item Researchers should communicate the details of the dataset\slash code\slash model as part of their submissions via structured templates. This includes details about training, license, limitations, etc. 
        \item The paper should discuss whether and how consent was obtained from people whose asset is used.
        \item At submission time, remember to anonymize your assets (if applicable). You can either create an anonymized URL or include an anonymized zip file.
    \end{itemize}

\item {\bf Crowdsourcing and research with human subjects}
    \item[] Question: For crowdsourcing experiments and research with human subjects, does the paper include the full text of instructions given to participants and screenshots, if applicable, as well as details about compensation (if any)? 
    \item[] Answer: \answerNA{} %
    \item[] Justification: The paper does not involve crowdsourcing or human subjects.
    \item[] Guidelines:
    \begin{itemize}
        \item The answer \answerNA{} means that the paper does not involve crowdsourcing nor research with human subjects.
        \item Including this information in the supplemental material is fine, but if the main contribution of the paper involves human subjects, then as much detail as possible should be included in the main paper. 
        \item According to the NeurIPS Code of Ethics, workers involved in data collection, curation, or other labor should be paid at least the minimum wage in the country of the data collector. 
    \end{itemize}

\item {\bf Institutional review board (IRB) approvals or equivalent for research with human subjects}
    \item[] Question: Does the paper describe potential risks incurred by study participants, whether such risks were disclosed to the subjects, and whether Institutional Review Board (IRB) approvals (or an equivalent approval/review based on the requirements of your country or institution) were obtained?
    \item[] Answer: \answerNA{} %
    \item[] Justification: The paper does not involve research with human subjects.
    \item[] Guidelines:
    \begin{itemize}
        \item The answer \answerNA{} means that the paper does not involve crowdsourcing nor research with human subjects.
        \item Depending on the country in which research is conducted, IRB approval (or equivalent) may be required for any human subjects research. If you obtained IRB approval, you should clearly state this in the paper. 
        \item We recognize that the procedures for this may vary significantly between institutions and locations, and we expect authors to adhere to the NeurIPS Code of Ethics and the guidelines for their institution. 
        \item For initial submissions, do not include any information that would break anonymity (if applicable), such as the institution conducting the review.
    \end{itemize}

\item {\bf Declaration of LLM usage}
    \item[] Question: Does the paper describe the usage of LLMs if it is an important, original, or non-standard component of the core methods in this research? Note that if the LLM is used only for writing, editing, or formatting purposes and does \emph{not} impact the core methodology, scientific rigor, or originality of the research, declaration is not required.
    \item[] Answer: \answerNA{} %
    \item[] Justification: LLMs were not used as a component of the core methodology.
    \item[] Guidelines:
    \begin{itemize}
        \item The answer \answerNA{} means that the core method development in this research does not involve LLMs as any important, original, or non-standard components.
        \item Please refer to our LLM policy in the NeurIPS handbook for what should or should not be described.
    \end{itemize}

\end{enumerate}

\end{document}